\documentclass[letterpaper]{article} 
\usepackage[submission]{aaai2026}  
\usepackage{times}  
\usepackage{helvet}  
\usepackage{courier}  
\usepackage[hyphens]{url}  
\usepackage{graphicx} 
\usepackage{natbib}  
\usepackage{caption} 
\usepackage{algorithm}
\usepackage{algorithmic}

\usepackage{newfloat}
\usepackage{listings}
\DeclareCaptionStyle{ruled}{labelfont=normalfont,labelsep=colon,strut=off} 
\floatstyle{ruled}
\newfloat{listing}{tb}{lst}{}
\floatname{listing}{Listing}
\usepackage{subfigure}
\usepackage{multirow}
\usepackage{amsmath}
\usepackage{booktabs}
\usepackage{amsfonts}

\title{Personalized One-shot Federated Graph Learning for Heterogeneous Clients}
\author{
    Written by AAAI Press Staff\textsuperscript{\rm 1}\thanks{With help from the AAAI Publications Committee.}\\
    AAAI Style Contributions by Pater Patel Schneider,
    Sunil Issar,\\
    J. Scott Penberthy,
    George Ferguson,
    Hans Guesgen,
    Francisco Cruz\equalcontrib,
    Marc Pujol-Gonzalez\equalcontrib
}
\affiliations{
    \textsuperscript{\rm 1}Association for the Advancement of Artificial Intelligence\\

    1101 Pennsylvania Ave, NW Suite 300\\
    Washington, DC 20004 USA\\
    proceedings-questions@aaai.org
}

\begin{document}

\maketitle

\begin{abstract}
Federated Graph Learning (FGL) has emerged as a promising paradigm for breaking data silos among distributed private graphs. In practical scenarios involving heterogeneous distributed graph data, personalized Federated Graph Learning (pFGL) aims to enhance model utility by training personalized models tailored to client needs. However, existing pFGL methods often require numerous communication rounds under heterogeneous graphs, leading to significant communication overhead and security concerns. While One-shot Federated Learning (OFL) enables collaboration in a single round, existing OFL methods are designed for image-centric tasks and are ineffective for graph data, leaving a critical gap in the field. Additionally, personalized models derived from existing methods suffer from bias, failing to effectively generalize to the minority. To address these challenges, we propose the first \textbf{O}ne-shot \textbf{p}ersonalized \textbf{F}ederated \textbf{G}raph \textbf{L}earning method (\textbf{O-pFGL}) for node classification, compatible with Secure Aggregation protocols for privacy preservation. Specifically, for effective graph learning in one communication round, our method estimates and aggregates class-wise feature distribution statistics to construct a global surrogate graph on the server, facilitating the training of a global graph model. To mitigate bias, we introduce a two-stage personalized training approach that adaptively balances local personal information and global insights from the surrogate graph, improving both personalization and generalization. Extensive experiments on 14 diverse real-world graph datasets demonstrate that our method significantly outperforms state-of-the-art baselines across various settings.
\end{abstract}


\section{Introduction}

Graphs are widely employed to model complex relationships between entities across a variety of domains~\cite{hyun2023anti, bang2023biomedical}. While various algorithms and models~\cite{gasteiger2018predict, wu2020comprehensive} have been proposed, most of them assume that graph data from different sources are managed centrally. However, collecting and managing such graph data is often costly, impractical, and poses significant privacy risks in many sensitive scenarios~\cite{fu2022federated}.

\begin{figure*}[tbp]
\centering
\includegraphics[width=0.95\textwidth]{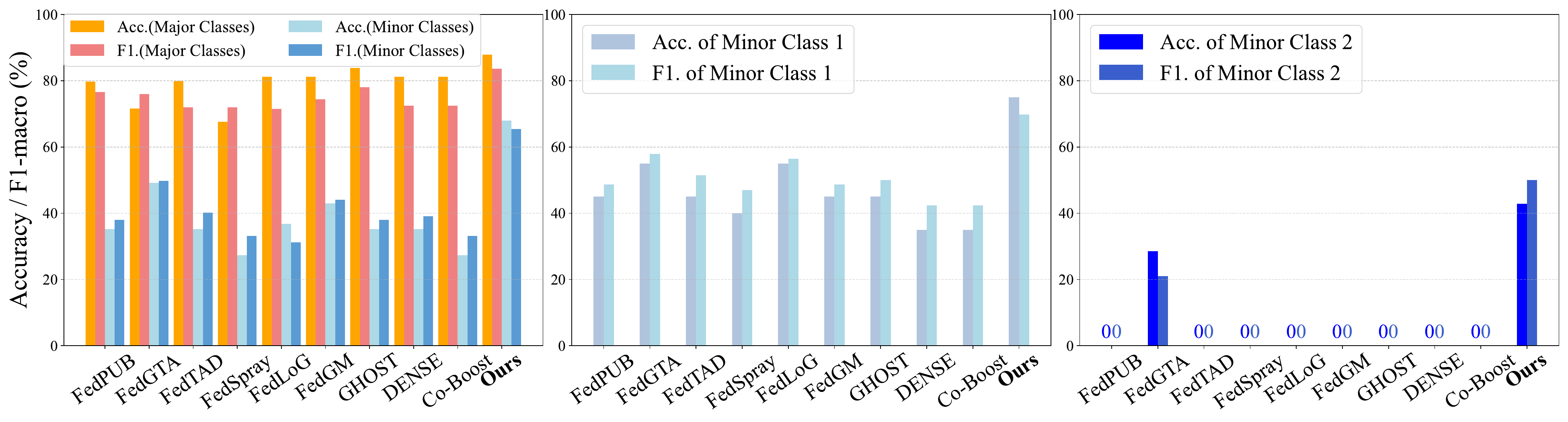}
\caption{Empirical analysis of a client on CiteSeer dataset to illustrate \textbf{L3}. \textbf{Left}: Existing methods result in biased models that neglect the minority. \textbf{Middle}: Performance of existing methods and our method on class 1 with 50 samples. \textbf{Right}: Performance of existing methods and our method on class 2 with 14 samples. Our method improves on both the majority and the minority.}
\label{fig:citeseer_major_minor}
\end{figure*}

To address the tension between the need for vast datasets and the growing demand for privacy protection, Federated Graph Learning (FGL) has emerged as a promising solution, which enables collaborative training on distributed graph data without requiring centralized management~\cite{pan2023lumos, gu2023dynamic}. While FGL shows its potential, it faces two major limitations: (1) Typical FGL struggles to effectively personalize models for joint improvement across heterogeneous clients' graph data~\cite{baek2023personalized}. This heterogeneity arises from variations of both node and structural properties, with adjacent nodes influencing each other, further complicating personalization. (2) FGL requires iterative communication between clients and the server, raising efficiency concerns due to communication overhead~\cite{kairouz2021advances, dai2022dispfl} and security concerns (e.g., man-in-the-middle attacks~\cite{wang2020man}). These limitations can be mitigated by integrating personalized federated graph learning (pFGL) and one-shot federated learning (OFL), which offer novel solutions to improve communication efficiency and model personalization.

To address these challenges, we explore the integration of pFGL and OFL. The pFGL methods allow for personalized model training by focusing on individual client graph data. OFL methods limit communication to a single round, thereby reducing the communication overhead and circumventing security risks such as man-in-the-middle attacks~\cite{guha2019one, zhang2022dense}. However, both existing pFGL and OFL methods face the following challenges and limitations:

\textbf{L1: Ineffectiveness for graph learning in one communication round under heterogeneity.} Existing pFGL methods~\cite{baek2023personalized, li2024fedgta} rely on iterative communication, making them ineffective within one communication round, especially under heterogeneous data. A recent work, GHOST~\cite{qianghost}, explores training a generalized model in one-shot communication. But it's not robust under severe non-IIDness and does not address personalization issues, as confirmed in our experiments. Furthermore, such parameter-aggregation-based methods inherently fail to support heterogeneous client models. Meanwhile, existing OFL methods are incompatible with graph data, as they primarily focus on image-centric tasks. Ensemble-based OFL methods~\cite{zhang2022dense, dai2024enhancing} can generate images from ensemble models but struggle with graph data due to the inter-dependency between nodes. Distillation-based OFL faces problems distilling interrelated graphs. Generative-based OFL methods~\cite{heinbaugh2023data, yang2024feddeo, yang2024exploring} rely on pre-trained generative models (e.g., Stable Diffusion~\cite{rombach2022high}), which are impractical for graph generation.

\textbf{L2: Incompatibility with Secure Aggregation.} Secure Aggregation is a widely used technique~\cite{bonawitz2017practical} designed to safeguard client privacy during uploading and aggregation in federated learning. However, most existing methods require clients to upload raw model parameters or auxiliary information independently, which complicates the process (e.g., similarity calculation or model ensemble) and hinders the implementation of privacy-preserving uploading and aggregation. The individual information could be revealed in the communication~\cite{zhu2024evaluating}.

\textbf{L3: Imbalance between personalization and local generalization.} Local graph data often exhibits a significant imbalance in both quantity and topology. Personalized models tend to be biased toward major classes (e.g., with larger quantities and high homophily to form compact communities), while neglecting minor classes with fewer samples and often surrounded by nodes from major classes. This imbalance leads to a degradation in model performance, particularly in underrepresented classes. A detailed empirical analysis is provided in Figure~\ref{fig:citeseer_major_minor}, which shows that existing methods result in heavy bias toward majority classes (class 3 and class 5, with a total of 180 samples) while failing to adequately represent minority classes (class 1 and class 2, with only 64 samples), highlighting the limitations of current methods in handling imbalanced graph data.

To overcome these limitations, we propose a \textbf{O}ne-shot \textbf{p}ersonalized \textbf{F}ederated \textbf{G}raph \textbf{L}earning method (\textbf{O-pFGL}). Our method supports model heterogeneity and Secure Aggregation protocols for privacy preservation. Specifically, for \textbf{L1} and \textbf{L2}, clients upload estimated feature distribution statistics, augmented by our proposed homophily-guided reliable node expansion strategy, instead of raw model parameters. The server securely aggregates these statistics to generate a global surrogate graph, which is subsequently utilized to train personalized models on clients, eliminating the need for direct model parameter exchanges and further protecting the client's model intellectual property. To address \textbf{L3}, we propose a two-stage personalized training approach. In the first stage, clients train a global generalized model based on the surrogate graph, which serves as both a fixed teacher capturing global knowledge and a starting point for subsequent fine-tuning. In the second stage, clients fine-tune the global model on local graph data with the node-adaptive distillation, which leverages the global knowledge from the teacher model to counteract biases toward the majority, ensuring effective learning for the minority and achieving a robust balance between personalization and generalization.

Our method significantly outperforms state-of-the-art through comprehensive experiments across various settings, including different graph scales, homophilic and heterophilic, learning paradigms, and model architectures. Experimental results demonstrate our method's superiority in achieving a personalization-generalization balanced model within one-shot communication, considering both data and model privacy. To sum up, our contributions are:

\textbf{New Problem:} One-shot Personalized FGL. We are the first to tackle the problem of personalization in one-shot federated graph learning under dual challenges of data and model heterogeneity, and ensuring both privacy.

\textbf{Novel Solution:} Our solution firstly aggregates client statistics rather than model parameters to generate a global surrogate graph that captures global distribution. Leveraging this, we then introduce two-stage personalized training with node-adaptive distillation to strike an effective balance between personalization and generalization.

\textbf{SOTA Performance:} We conduct extensive experiments on 14 diverse real-world graph datasets and demonstrate that our method consistently and significantly outperforms existing baselines across various settings.

\section{Related Works}
\subsection{Federated Graph Learning}
From the graph level~\cite{xie2021federated, tan2023federated}, each client possesses multiple independent graphs and targets at graph-level tasks. From the subgraph level, each client possesses a subgraph of an implicit global graph. FedPUB~\cite{baek2023personalized} and FedGTA~\cite{li2024fedgta} personalize models by similarity of functional embedding and mixed moments of neighbor features. AdaFGL~\cite{li2024adafgl} studies the structure non-IID problem. FedTAD~\cite{zhu2024fedtad} performs data-free distillation on the server. FedGM~\cite{zhang2025rethinking} adopts gradient matching to synthesize graphs and aggregate the gradients. GHOST~\cite{qianghost} trains aligned proxy models, and the server integrates the knowledge from uploaded proxy models. To complete missing connections between subgraphs, FedSage~\cite{zhang2021subgraph} and FedDEP~\cite{zhang2024deep} additionally train a neighborhood generator. FedStruct~\cite{aliakbari2024decoupled} decouples the structure learning and node representation learning. To better represent the minority in local graph data, FedSpray~\cite{fu2024federated} learns class-wise structure proxies to mitigate biased neighborhoods. FedLoG~\cite{kim2025subgraph} synthesizes global data from clients' synthetic nodes and generalizes the local training. Detailed description and analysis are in the Appendix~\ref{sec:appendix_related_works}.

\subsection{One-shot Federated Learning}
One-shot federated learning largely reduces communication costs and circumvents potential man-in-the-middle attacks. Mainstream OFL methods can be classified into: (1) Ensemble-based methods~\cite{guha2019one, zhang2022dense, dai2024enhancing}, (2) Distillation-based methods~\cite{zhou2020distilled, song2023federated}, (3) Generative-based methods~\cite{heinbaugh2023data, yang2024exploring, yang2024feddeo}. However, existing OFL methods primarily focus on image data and are either incompatible or ineffective for graph learning. Detailed description and analysis are in the Appendix~\ref{sec:appendix_related_works}.

\begin{figure*}[tbp]
\centerline{\includegraphics[width=0.82\textwidth]{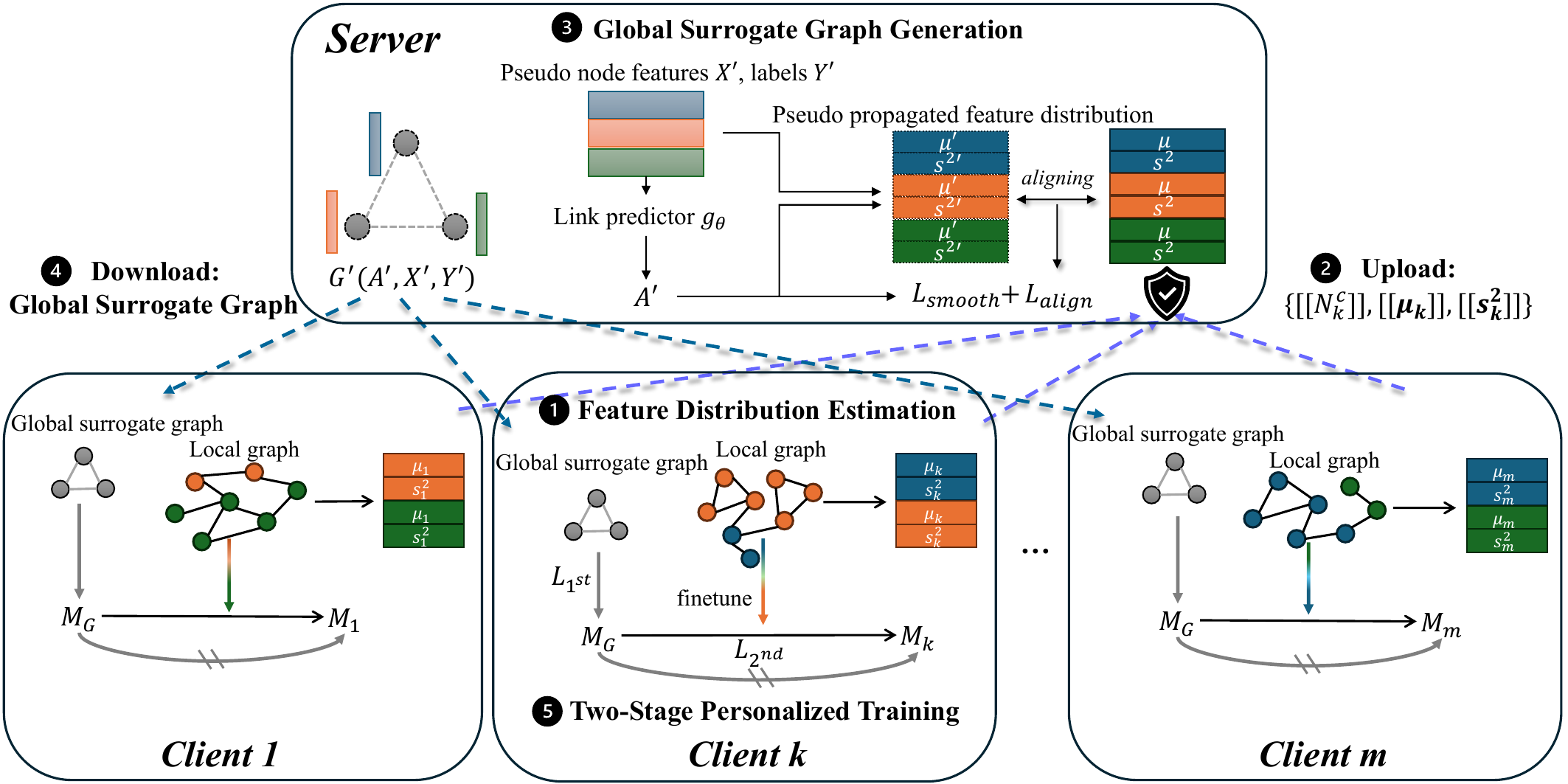}}
\caption{The overall pipeline of our proposed method, which comprises 5 steps: (1) Each client estimates the class-wise feature distribution and computes corresponding statistics on its local graph data; (2) Each client uploads its class-wise statistics to the server (compatible with Secure Aggregation); (3) The server aggregates uploaded statistics to recover the global distribution. Based on the recovered distribution, the server generates a small-size global surrogate graph; (4) The server distributes the generated global surrogate graph to clients; (5) Each client performs two-stage personalized training using both local graph data and the downloaded global surrogate graph, deriving a model with improved personalization and generalization.}
\label{fig:overall}
\end{figure*}

\section{Preliminaries}

\subsection{Notations and Definitions} 
Consider a FGL setting with $m$ clients, where $k$-th client possesses the local undirected graph $G_k(\mathcal{V}_k, \mathcal{E}_k)$ with $|\mathcal{E}_k|$ edges and $|\mathcal{V}_k| = N_k$ nodes. Node features matrix is $\mathbf{X}_k=\{\mathbf{x}_1, ..., \mathbf{x}_{N_k}\}\in \mathbb{R}^{|\mathcal{V}_k| \times f}$, where $f$ is the feature dimension and $\mathbf{x}_i$ is the feature of node $v_i$. $\mathbf{A}_k$ is the adjacency matrix. The labels of nodes are denoted by $\mathbf{Y}_k = \{y_1,..., y_{n_k}\} \in \mathbb{R}^{|\mathcal{V}_k|\times C}$, where $C$ is the number of distinct classes. 

In the semi-supervised node classification task~\cite{kipf2016semi}, nodes on the $k$-th client are partitioned into labeled nodes $\mathcal{V}_{k, L}$ (i.e., training set) and unlabeled nodes $\mathcal{V}_{k, U}$ (i.e., test set). The corresponding labels are $\mathbf{Y}_{k, L}$ and $\mathbf{Y}_{k, U}$, respectively. The neighborhood of node $v_i$ in $G_k(\mathcal{V}_k, \mathcal{E}_k)$ is $\mathcal{N}_{v_i} = \{u | \exists e_{u, v_i} \in \mathcal{E}_k \}$, where edge $e_{u, v_i}$ connects node $u$ and $v_i$. $d(v_i)$ denotes the degree of node $v_i$. We define the node homophily for node $v_i$ as:
\begin{equation}
    h_{node}(v_i) = \frac{\left|\{u | (u \in \mathcal{N}_{v_i} \cap \mathcal{V}_L) \text{and} (y_{v_i}=y_u) \}\right|}{|\mathcal{N}_{v_i} \cap \mathcal{V}_L|},
\end{equation}
and the class homophily for class $c$ is:
\begin{equation}
\label{eq:H(c)}
    H(c) = \sum_{u \in \mathcal{N}^c} h_{node}(u),
\end{equation}
where $\mathcal{N}^c = \{v_i \in \mathcal{V}_L | y_i = c\}$ is the set of labeled nodes belonging to class $c$. $H(c)$ considers both the quantity and topology of nodes. A higher $H(c)$ indicates nodes of class $c$ are numerous and exhibit strong homophily, making them more predictable for a vanilla graph model.

\subsection{Problem Formulation}
We formulate the one-shot personalized federated graph learning for the node classification over $m$ clients with non-overlapped distributed subgraphs as:
\begin{equation}
    \min \sum_{k=1}^{m} L(M_k(\mathbf{A}_k, \mathbf{X}_k), \mathbf{Y}_{k, U}).
\label{eq:overall_obj_func}
\end{equation}
This objective aims to minimize the aggregated generalization loss $L$ on their unlabeled node set with personalized models $M_k$ across all clients within a single upload-download communication round.

\section{Proposed Method}

The overall pipeline of our proposed method is illustrated in Figure~\ref{fig:overall}. The details are explained in the following sections.

\subsection{Feature Distribution Estimation}
\label{sec:feat_dist_esti}
To estimate the class-wise feature distribution locally, each client propagates node features on local graph data. Specifically, on client $k$, the propagated features are:
\begin{equation}
    \mathbf{X}^{prop}_k = \|_{i=0}^h \tilde{\mathbf{A}}_k^i \mathbf{X}_k,
\label{eq:client_feat_prop}
\end{equation}
where $h$ is propagation depth and $\|$ represents concatenation. $\tilde{\mathbf{A}}_k$ is augmented normalized adjacency matrix. Using propagated features, we estimate the unbiased sample mean and variance of features for labeled nodes in each class:
\begin{equation}
\begin{aligned}
    \boldsymbol{\mu}_k^c &= \frac{1}{|\mathcal{V}_{k,L}^c|} \sum_{v_i \in \mathcal{V}_{k,L}^c} \mathbf{x}^{prop}_i, \\
    \boldsymbol{s^2}_k^c &= \frac{1}{|\mathcal{V}_{k,L}^c|-1} \sum_{v_i \in \mathcal{V}_{k,L}^c}(\mathbf{x}^{prop}_i - \boldsymbol{\mu}_k^c)^2,
\end{aligned}
\label{eq:client_esimate_mean_var}
\end{equation}
where $\mathcal{V}_{k,L}^c$ denotes the labeled nodes of class $c$ on client $k$. The statistics are estimated only for classes with sufficient labeled nodes, determined by a client-specified threshold.

Our estimation is compatible with both homophilic and heterophilic graphs. For heterophilic graphs, the concatenation in Eq~\ref{eq:client_feat_prop} ensures the ego-information and high-order information are not conflated. This aligns with the core principles of MixHop~\cite{abu2019mixhop}, of which effectiveness has been widely proven on heterophilic graphs. For homophilic graphs, we further propose a plug-and-play Homophily-guided Reliable node Expansion (HRE) strategy to enhance the accurate estimation when the labeled nodes are scarce. In HRE, we first apply Label Propagation~\cite{iscen2019label} to obtain the soft labels $\tilde{\mathbf{y}} \in \mathbb{R}^C$ for unlabeled nodes. We denote $c_i' = \arg\max_{0\leq i < C} \tilde{\mathbf{y}}(i)$, then identify reliable soft labels based on the following criteria: (C1) High Node Degree: $d(v_i) \geq d_{th}$; (C2) High Confidence: $\tilde{\mathbf{y}}_i(c_i') \geq f_{th}$; (C3) High Class Homophily: $c_i' \in topK(H)$, indicating the node is likely to share the same label with its neighbors. A node with more neighbors could reduce the variance in predictions. With these criteria, the reliable nodes $\mathcal{V}_{k, r}$ set with inferred true labels is:
\begin{equation}
\begin{aligned}
    \mathcal{V}_{k, r} =& \{v_i | (v_i \in \mathcal{V}_{k, U}) \wedge (d(v_i) \geq d_{th}) \\
    &\wedge (\tilde{\mathbf{y}}_i(c') \geq f_{th}) \wedge (c' \in topK(H))\},
\end{aligned}
\end{equation}
where $topK(H)$ is the set of classes with top $K$ largest values of $H(c)$. $K, f_{th}$ and $d_{th}$ are client-defined hyper-parameters. The reliable nodes $\mathcal{V}_{k, r}$, along with their inferred labels, can expand the original labeled nodes set $\mathcal{V}_{k, L}$. The estimation in Eq.~\ref{eq:client_esimate_mean_var} could be augmented as:
\begin{equation}
\begin{aligned}
    \boldsymbol{\mu}_k^c &= \frac{1}{|\mathcal{V}_{k,L}^c \cup \mathcal{V}_{k, r}^c|} \sum_{v_i \in \mathcal{V}_{k,L}^c \cup \mathcal{V}_{k, r}^c } \mathbf{x}^{prop}_i, \\
    \boldsymbol{s^2}_k^c &= \frac{1}{|\mathcal{V}_{k,L}^c \cup \mathcal{V}_{k, r}^c|-1} \sum_{v_i \in \mathcal{V}_{k,L}^c \cup \mathcal{V}_{k, r}^c }(\mathbf{x}^{prop}_i - \boldsymbol{\mu}_k^c)^2,
\label{eq:final_client_esimate_mean_var}
\end{aligned}
\end{equation}
where $\mathcal{V}_{k, r}^c$ is the subset of reliable nodes in $\mathcal{V}_{k, r}$ belonging to class $c$. This augmentation ensures a more accurate estimation of class-wise feature distributions by incorporating reliable nodes into the computation.

\subsection{Global Surrogate Graph Generation}

After clients estimate the class-wise feature distribution statistics $\{\boldsymbol{\mu}_k^c, \boldsymbol{s^2}_k^c\}$, they upload them along with sample quantity of each class $\{N_k^c\}$ to the server. The server aggregates them to recover the global class-wise feature distribution. Specifically, the global mean and global variance of class $c$ are computed unbiasedly as:
\begin{equation}
\small
\begin{aligned}
    N^c &= \sum_{k=1}^{m} N_k^c, \quad \boldsymbol{\mu}^c = \frac{1}{N^c}\sum_{k=1}^{m} N_k^c\boldsymbol{\mu}_k^c, \\
    \boldsymbol{s^2}^c &= \frac{1}{N^c-m}(\sum_{k=1}^{m}(N_k^c - 1)\boldsymbol{s^2}_k^c + \sum_{k=1}^{m}N_k^c(\boldsymbol{\mu}_k^c - \boldsymbol{\mu}^c)^2).
\end{aligned}
\label{eq:server_mean_var}
\end{equation}
Notably, the recovered global statistics remain unbiased regardless of the degree of data non-IIDness.

With recovered global distribution, we generate a small-sized global surrogate graph $G' = \{\mathbf{A}', \mathbf{X}', \mathbf{Y}'\}$ by aligning its feature distribution with global distribution. The surrogate graph serves as a representative of the latent overall graph data across all clients. The generation process is illustrated at the top of Figure~\ref{fig:overall}. We first pre-set the number of nodes in the surrogate graph and initialize the learnable node features $\mathbf{X}'$ with Gaussian noise. The adjacency matrix $\mathbf{A}'$ is derived using a learnable link predictor $g_\theta$:
\begin{equation}
\small
\label{eq:adj_design}
    \mathbf{A}'_{i, j} = g_\theta(\mathbf{X}', \delta) = \mathbb{I}(\delta \leq \sigma(\frac{g_\theta(\mathbf{x}_i \| \mathbf{x}_j)+ g_\theta(\mathbf{x}_j \| \mathbf{x}_i)}{2})),
\end{equation}
where $\sigma$ is Sigmoid function, $\delta$ is a hyper-parameter to control sparsity, and $\|$ is concatenation. We implement $g_{\theta}$ as an MLP. We also discuss other designs in the Appendix~\ref{sec:adj_design}.

Following~\cite{xiao2024simple}, to optimize $\mathbf{X}'$ and $g_\theta$, we first propagate node features $\mathbf{X}'$ as in Eq.~\ref{eq:client_feat_prop}. Then we calculate the class-wise sample mean $\boldsymbol{\mu}'^c$ and sample variance $\boldsymbol{s^2}'^c$ of the propagated node features $\mathbf{X}'^{prop}$ following Eq.~\ref{eq:client_esimate_mean_var}. Then we calculate the alignment loss:
\begin{equation}
    L_{align} = \sum_{c=0}^{C-1} \lambda_c ((\boldsymbol{\mu}'^c - \boldsymbol{\mu}^c)^2 + (\boldsymbol{s^2}'^c - \boldsymbol{s^2}^c)^2),
\end{equation}
where $\lambda_c$ is the proportion of nodes belonging to class $c$ relative to the total number of nodes. To ensure the smoothness of the surrogate graph, a smoothness loss is applied:
\begin{equation}
    L_{smt} = \frac{1}{\sum_{i, j}\mathbf{A}'_{i,j}} \sum_{i, j}\mathbf{A}'_{i,j}\exp{(-\frac{\|\mathbf{x}_i - \mathbf{x}_j\|^2}{2})}.
\end{equation}
We optimize $\mathbf{X}'$ and $g_\theta$ by the overall optimization objective: $\min_{\mathbf{X}', \theta}\ (L_{align} + \alpha L_{smt})$. 
Once the optimization finishes, the server distributes $G' = \{\mathbf{A}', \mathbf{X}', \mathbf{Y}'\}$ to clients for further personalized training.

\textbf{Privacy-preservation with Secure Aggregation.} We demonstrate compatibility with Secure Aggregation protocols in uploading and aggregating $\{\boldsymbol{\mu}_k^c, \boldsymbol{s^2}_k^c, N_k^c\}$. Secure Aggregation is a crucial technique in federated learning that enables the server to compute the sum of large, user-held data vectors securely, without accessing individual client contributions. This property aligns with the aggregation process in Eq.~\ref{eq:server_mean_var}, as it can be expressed as a series of weighted average operations. The calculations of $N^c$ and $\boldsymbol{\mu}^c$ are inherently weighted average operations. The calculation of $\boldsymbol{s^2}^c$ can also be decomposed into weighted average operations:
\begin{equation}
\begin{aligned}
    \boldsymbol{s^2}^c
     &= \frac{\sum_{k}^{m}(N_k^c - 1)\boldsymbol{s^2}_k^c}{N^c - m} +  \\
     &\frac{\sum_k^{m}N_k^c (\boldsymbol{\mu}_k^c)^2 - 2 \boldsymbol{\mu}^c \sum_k^{m}\boldsymbol{\mu}_k^c + N^c \sum_k^{m}(\boldsymbol{\mu}^c)^2}{N^c - m}.
\end{aligned}
\end{equation}
Breaking this down further, the calculation involves the weighted averages of $(N_k^c-1)\boldsymbol{s^2}_k^c, N_k^c (\boldsymbol{\mu}_k^c)^2$ and $\boldsymbol{\mu}_k^c$. Thus, our entire aggregation process is compatible with off-the-shelf Secure Aggregation protocols and \textbf{would not reveal individual client data distribution}. We detail the implementation with pseudo code in Appendix~\ref{sec:secure_agg_implementation}.

\subsection{Personalization with Node Adaptive Distillation}
\label{sec:apt}

With the downloaded global surrogate graph, each client trains its personalized model locally. However, the imbalanced local graph data often results in biased models that perform well on major classes but neglect minor classes. To achieve better personalization (accuracy$\uparrow$) and generalization (F1-macro$\uparrow$), we propose a two-stage personalization training approach with node-adaptive distillation, which leverages both global information in the global surrogate graph and personal information in the local graph.

\paragraph{Stage 1}Each client trains a generalized model $M_G$ using the global surrogate graph, ensuring $M_G$ to capture global, balanced knowledge. The training objective is to minimize:
\begin{equation}
    L_{1^{st}} = L_{ce}(M_G(\mathbf{A}', \mathbf{X}'), \mathbf{Y}').
\end{equation}
A copy of $M_G$ is detached as a fixed teacher model for the next stage. $L_{ce}$ denotes the cross-entropy loss. Note that if the server knows the model architecture, an alternative is to conduct this stage on the server. However, we can adapt to more rigorous scenarios where models' intellectual property needs to be protected. In these cases, it's necessary to conduct this stage on each client.

\paragraph{Stage 2} Client $k$ fine-tunes $M_G$ with local graph data to derive a personalized model $M_k$. However, fine-tuning on imbalanced local graph data often leads to over-fitting on major classes while forgetting minor classes. As illustrated in Figure~\ref{fig:citeseer_gap_H_major}, the fine-tuned model shows improvement in major classes but suffers degradation in minor classes.

To mitigate this, we propose to utilize the global knowledge of $M_G$ and personalized knowledge in local graph data to train a better $M_k$ with the node adaptive distillation. For nodes of major classes, the model directly learns through supervised fine-tuning. For nodes of minor classes, $M_k$ additionally distills the predictive ability from $M_G$. The training objective for the second stage is to minimize:
\begin{equation}
    L_{2^{nd}} = L_{ft} + L_{dist},
\end{equation}
where $L_{ft}$ is cross-entropy loss for fine-tuning on local data, and $L_{dist}$ is the weighted sum of point-wise KL divergence loss to distill global knowledge from $M_G$ into $M_k$:
\begin{equation}
    L_{dist} = \sum_{\mathbf{x}_i \in \mathbf{X}_k} \gamma_i L_{kl}(M_k(\mathbf{A}_k, \mathbf{x}_i), M_G(\mathbf{A}_k, \mathbf{x}_i)),
\label{eq:L_dist}
\end{equation}
where $\gamma_i$ is a node-specific weighting factor to control the balance between local fine-tuning and global knowledge distillation. To determine $\gamma_i$ for node $v_i$, we consider both the node property and class property. We first define the class-aware distillation weight vector $\mathbf{w_{dist}}\in \mathbb{R}^C$ where:
\begin{equation}
    \mathbf{w_{dist}}[c] = \frac{1}{1+\log(H(c) + 1)},
\label{eq:w_dist}
\end{equation}
$H(c)$ is defined in Eq~\ref{eq:H(c)}. Then we use node $v_i$'s soft label $\tilde{\mathbf{y}}_i$ by Label Propagation and $\mathbf{w}_{dist}$ to calculate $\gamma_i$:
\begin{equation}
    \gamma_i = \beta \tilde{\mathbf{y}}_i \cdot \mathbf{w}_{dist},
\end{equation}
where $\beta$ is a scaling hyper-parameter. Intuitively explained, if a node is predicted to class $c$ with a small $w_{dist}(c)$ (e.g., node $b$ in Figure~\ref{fig:client_ada_per} in Appendix~\ref{sec:illustration_gamma}), it likely belongs to major classes with high homophily. In this case, $\gamma_i$ is small, and $M_i$ primarily learns through fine-tuning; If a node is predicted to class $c$ with a large $w_{dist}(c)$ or has a blended prediction (e.g. node $a$ in Figure~\ref{fig:client_ada_per} in Appendix~\ref{sec:illustration_gamma}), it likely belongs to minor classes or has low homophily. These nodes are often neglected during fine-tuning. Thus $\gamma_i$ should be large to incorporate the corresponding global knowledge from $M_G$.

\begin{figure}[tbp]
    \centering
    \includegraphics[width=\linewidth]{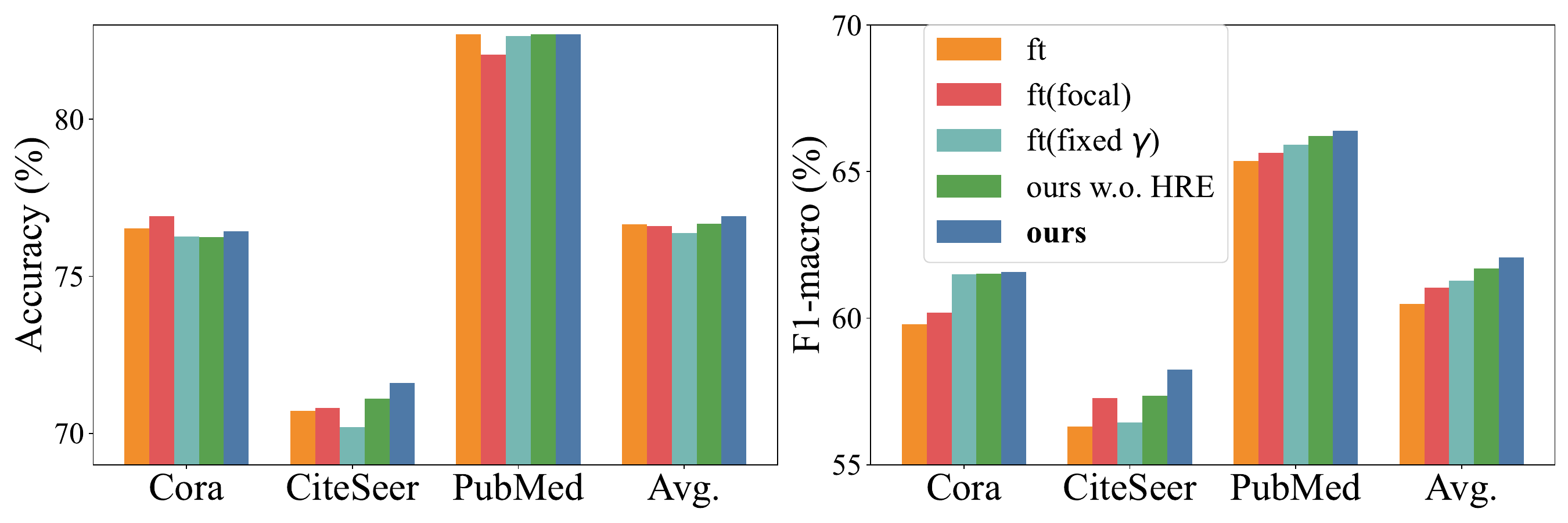}
    \caption{Ablation study under Louvain partition.}
    \label{fig:louvain_ablation}
\end{figure}

\begin{table*}[tbp]
\setlength{\tabcolsep}{1mm}
\small
\centering
\begin{tabular}{c|cc|cc|cc|cc|cc}
\toprule[1pt]
\multirow{2}{*}{Method} & \multicolumn{2}{c|}{\textbf{Cora}} & \multicolumn{2}{c|}{\textbf{CiteSeer}} & \multicolumn{2}{c|}{\textbf{PubMed}} & \multicolumn{2}{c|}{\textbf{ogbn-arxiv}} & \multicolumn{2}{c}{\textbf{ogbn-products}} \\ \cline{2-11} 
 & Acc.           & F1            & Acc.           & F1            & Acc.           & F1            & Acc.           & F1            & Acc.           & F1            \\ \midrule
Standalone & 67.17{\scriptsize$\pm$0.38} & 41.79{\scriptsize$\pm$0.52} & 59.82{\scriptsize$\pm$0.12} & 42.70{\scriptsize$\pm$0.34} & 81.74{\scriptsize$\pm$0.06} & 64.61{\scriptsize$\pm$0.18} & 66.46{\scriptsize$\pm$0.24} & 16.59{\scriptsize$\pm$0.57} & 85.25{\scriptsize$\pm$0.02} & 22.35{\scriptsize$\pm$0.19} \\
FedAvg & 69.68{\scriptsize$\pm$0.48} & 45.10{\scriptsize$\pm$0.75} & 63.92{\scriptsize$\pm$0.10} & 45.97{\scriptsize$\pm$0.43} & 78.24{\scriptsize$\pm$0.08} & 42.24{\scriptsize$\pm$0.13} & 67.13{\scriptsize$\pm$0.14} & 18.46{\scriptsize$\pm$0.23} & 85.34{\scriptsize$\pm$0.04} & 22.80{\scriptsize$\pm$0.04} \\ \midrule
FedPUB & 68.50{\scriptsize$\pm$0.38} & 43.35{\scriptsize$\pm$0.27} & 61.80{\scriptsize$\pm$0.37} & 43.96{\scriptsize$\pm$1.05} & 81.63{\scriptsize$\pm$0.54} & 63.39{\scriptsize$\pm$0.69} & 65.34{\scriptsize$\pm$0.28} & 12.50{\scriptsize$\pm$0.20} & 83.91{\scriptsize$\pm$0.31} & 17.29{\scriptsize$\pm$0.52} \\ 
FedGTA & 42.61{\scriptsize$\pm$0.76} & 23.47{\scriptsize$\pm$1.06} & 67.12{\scriptsize$\pm$1.58} & 52.68{\scriptsize$\pm$1.81} & 75.99{\scriptsize$\pm$1.21} & 57.48{\scriptsize$\pm$1.44} & 58.30{\scriptsize$\pm$0.30} & 12.22{\scriptsize$\pm$0.26} & 75.87{\scriptsize$\pm$0.04} & 16.58{\scriptsize$\pm$0.15} \\ 
FedTAD & 70.10{\scriptsize$\pm$0.22} & 45.57{\scriptsize$\pm$0.27} & 63.93{\scriptsize$\pm$0.27} & 44.59{\scriptsize$\pm$0.41} & 78.21{\scriptsize$\pm$0.31} & 42.45{\scriptsize$\pm$0.80} & 68.77{\scriptsize$\pm$0.12} & 27.33{\scriptsize$\pm$0.50} & \multicolumn{2}{c}{OOM} \\ 
FedSpray & 63.14{\scriptsize$\pm$1.73} & 40.82{\scriptsize$\pm$1.54} & 54.89{\scriptsize$\pm$4.24} & 39.56{\scriptsize$\pm$3.58} & 77.89{\scriptsize$\pm$1.67} & 64.23{\scriptsize$\pm$2.03} & 68.68{\scriptsize$\pm$0.05} & 19.90{\scriptsize$\pm$0.16} & 84.06{\scriptsize$\pm$0.11} & 20.80{\scriptsize$\pm$0.05} \\ 
FedLoG & 66.24{\scriptsize$\pm$0.60} & 38.68{\scriptsize$\pm$0.54} & 61.19{\scriptsize$\pm$0.33} & 39.56{\scriptsize$\pm$3.58} & 79.40{\scriptsize$\pm$0.64} & 57.33{\scriptsize$\pm$0.47} & \multicolumn{2}{c|}{OOM} & \multicolumn{2}{c}{OOM} \\ 
FedGM & 66.61{\scriptsize$\pm$0.11} & 37.47{\scriptsize$\pm$0.36} & 61.70{\scriptsize$\pm$0.24} & 42.38{\scriptsize$\pm$0.45} & 71.83{\scriptsize$\pm$0.27} & 29.96{\scriptsize$\pm$0.31} & 45.72{\scriptsize$\pm$0.55} & 1.67{\scriptsize$\pm$0.05} & \multicolumn{2}{c}{OOM} \\
GHOST & 69.16{\scriptsize$\pm$0.11} & 43.45{\scriptsize$\pm$0.27} & 62.78{\scriptsize$\pm$0.23} & 43.72{\scriptsize$\pm$0.37} & 81.93{\scriptsize$\pm$0.21} & 65.36{\scriptsize$\pm$0.21} & 67.69{\scriptsize$\pm$0.03} & 19.04{\scriptsize$\pm$0.13} & \multicolumn{2}{c}{OOM} \\
\midrule
DENSE & 69.89{\scriptsize$\pm$0.30} & 45.35{\scriptsize$\pm$0.55} & 64.09{\scriptsize$\pm$0.48} & 46.00{\scriptsize$\pm$0.39} & 78.16{\scriptsize$\pm$0.06} & 41.96{\scriptsize$\pm$0.33} & 68.80{\scriptsize$\pm$0.04} & 27.60{\scriptsize$\pm$0.36} & 86.12{\scriptsize$\pm$0.03} & 26.58{\scriptsize$\pm$0.06} \\ 
Co-Boost & 69.95{\scriptsize$\pm$0.40} & 45.43{\scriptsize$\pm$0.52} & 64.21{\scriptsize$\pm$0.27} & 46.00{\scriptsize$\pm$0.11} & 78.13{\scriptsize$\pm$0.05} & 41.84{\scriptsize$\pm$0.22} & 66.27{\scriptsize$\pm$0.08} & 16.37{\scriptsize$\pm$0.14} & 85.24{\scriptsize$\pm$0.00} & 22.31{\scriptsize$\pm$0.11} \\ \midrule
O-pFGL& \textbf{76.43}{\scriptsize$\pm$1.24} & \textbf{61.58}{\scriptsize$\pm$2.16} & \textbf{71.61}{\scriptsize$\pm$0.40} & \textbf{58.24}{\scriptsize$\pm$0.96} & \textbf{82.71}{\scriptsize$\pm$0.03} & \textbf{66.40}{\scriptsize$\pm$0.34} & \textbf{70.93}{\scriptsize$\pm$0.59} & \textbf{36.66}{\scriptsize$\pm$0.22} & \textbf{86.63}{\scriptsize$\pm$0.02} & \textbf{27.21}{\scriptsize$\pm$0.15} \\ 
\bottomrule[1pt]
\end{tabular}%
\caption{Performance under Louvain partition with 10 clients. OOM represents out-of-memory.}
\label{tab:transductive_louvain_10_clients}
\end{table*}

\subsection{Privacy Discussion}
Our method provides privacy for both client models and data. Model Privacy: By not sharing model parameters, our method inherently mitigates model-based attacks (e.g., model inversion and membership inference attacks~\cite{zhu2024evaluating}). This design crucially allows clients to use their proprietary models, fostering collaboration without exposing intellectual property. Data Privacy: Client data is protected twofold. First, the shared statistics are computed via a non-injective operation (Eq.~\ref{eq:final_client_esimate_mean_var}), making it computationally infeasible to reverse-engineer specific data samples. Second, our framework is fully compatible with Secure Aggregation protocols, which prevent the server from accessing any individual client's statistics.

\begin{table}[tbp]
\setlength{\tabcolsep}{1mm}
\small
\centering
\begin{tabular}{c|cc|cc}
\toprule
\multirow{2}{*}{Method} & \multicolumn{2}{c|}{\textbf{Actor}}           & \multicolumn{2}{c}{\textbf{genius}}   \\ \cline{2-5}  \rule{0pt}{2ex}
  & Acc.           & F1            & Acc.           & F1       \\ \midrule
  
Standalone & 58.77{\scriptsize$\pm$0.28} & 20.64{\scriptsize$\pm$0.22} & 96.27{\scriptsize$\pm$1.38} & 53.30{\scriptsize$\pm$1.71}\\ 

FedAvg & 60.01{\scriptsize$\pm$0.21} & 21.17{\scriptsize$\pm$0.12} & 94.79{\scriptsize$\pm$1.95} & 53.30{\scriptsize$\pm$1.71} \\ \midrule

FedPUB & 63.43{\scriptsize$\pm$0.06} & 21.11{\scriptsize$\pm$0.17} & 95.58{\scriptsize$\pm$1.55} & 51.61{\scriptsize$\pm$.65} \\

FedGTA & 62.58{\scriptsize$\pm$0.07} & 16.05{\scriptsize$\pm$0.06} & 92.23{\scriptsize$\pm$1.67} & 47.67{\scriptsize$\pm$0.26} \\

FedTAD & 60.32{\scriptsize$\pm$0.12} & 21.30{\scriptsize$\pm$0.11} & \multicolumn{2}{c}{OOM}  \\

FedSpray & 65.81{\scriptsize$\pm$0.46} & 20.13{\scriptsize$\pm$0.18} & 95.34{\scriptsize$\pm$1.71} & 53.33{\scriptsize$\pm$0.36} \\

FedLoG & 61.49{\scriptsize$\pm$0.82} & 20.53{\scriptsize$\pm$0.38} & \multicolumn{2}{c}{OOM}  \\ 

FedGM & 58.99{\scriptsize$\pm$0.09} & 20.94{\scriptsize$\pm$0.19} & 96.88{\scriptsize$\pm$0.27} & 53.92{\scriptsize$\pm$0.28} \\ 

GHOST & 58.94{\scriptsize$\pm$0.09} & 20.70{\scriptsize$\pm$0.07} & \multicolumn{2}{c}{OOM} \\ 
\midrule

DENSE & 59.88{\scriptsize$\pm$0.11} & 21.05{\scriptsize$\pm$0.02} & 96.81{\scriptsize$\pm$0.13} & 53.97{\scriptsize$\pm$0.10} \\

Co-Boost & 60.02{\scriptsize$\pm$0.53} & 21.11{\scriptsize$\pm$0.22} & 96.73{\scriptsize$\pm$0.11} & 53.80{\scriptsize$\pm$0.12} \\ \midrule

O-pFGL & \textbf{67.54}{\scriptsize$\pm$0.32} & \textbf{22.63}{\scriptsize$\pm$0.26} & \textbf{97.05}{\scriptsize$\pm$0.17} & \textbf{54.22}{\scriptsize$\pm$0.17} \\ 

\bottomrule
\end{tabular}
\caption{Performance on heterophilic graph datasets.}
\label{tab:heterophilic}
\end{table}

\section{Experiments}

\subsection{Experimental Setup}

\paragraph{Datasets.}
We conduct experiments on 14 datasets, including Cora, CiteSeer, PubMed~\cite{yang2016revisiting}, ogbn-arxiv, ogbn-products~\cite{hu2020open}, Computers, Photo, CS, Physics datasets~\cite{shchur2018pitfalls}, Flickr~\cite{zeng2019graphsaint}, Reddit~\cite{hamilton2017inductive}, Reddit2~\cite{zeng2019graphsaint}, Actor, and genius~\cite{luan2024heterophilic}. Table~\ref{tab:data} summarizes these datasets. For Actor and genius datasets, we employ Label non-IID partition~\cite{qianghost} with $Dir(0.2)$. For the others, we employ Louvain/Metis-based Label Imbalance Split~\cite{li2024openfgl} to simulate distributed subgraphs.

\paragraph{Baselines.}
\label{sec:exp_baselines}
Our baselines include 7 FGL methods\footnote{We do not include FedSage~\cite{zhang2021subgraph}, AdaFGL~\cite{li2024adafgl}, and FedStruct~\cite{aliakbari2024decoupled} since they need extra communication rounds before federated optimization. They cannot train models within one-shot communication.}: (1) FedPUB~\cite{baek2023personalized}, (2) FedGTA~\cite{li2024fedgta}, (3) FedTAD~\cite{zhu2024fedtad}, (4) FedSpray~\cite{fu2024federated}, (5) FedLoG~\cite{kim2025subgraph}, (6) FedGM~\cite{zhang2025rethinking}, (7) GHOST~\cite{qianghost} and 2 OFL methods: (8) DENSE~\cite{zhang2022dense}, (9) Co-Boost~\cite{dai2024enhancing}. We also include (10) Standalone, which locally trains models, and (11) FedAvg with fine-tuning. Detailed descriptions of these baselines are in the Appendix~\ref{sec:appendix_baselines}.

\paragraph{Configurations.}\label{sec:hyper-parameter} We conduct experiments over a single communication round. We employ the two-layer GCN with a hidden dimension of 64. We employ both \textbf{accuracy}(\%) and \textbf{F1-macro}(\%) metrics. F1-macro provides a more robust measure of the model's generalization under imbalanced data. We report the mean and standard deviation across three runs. In our O-pFGL, the global surrogate graph is set small, with its size not exceeding 300 nodes across all datasets. Detailed configurations are in the Appendix~\ref{sec:app_hyper-parameter}.

\begin{figure}[tbp]
    \centering
    \includegraphics[width=0.8\linewidth]{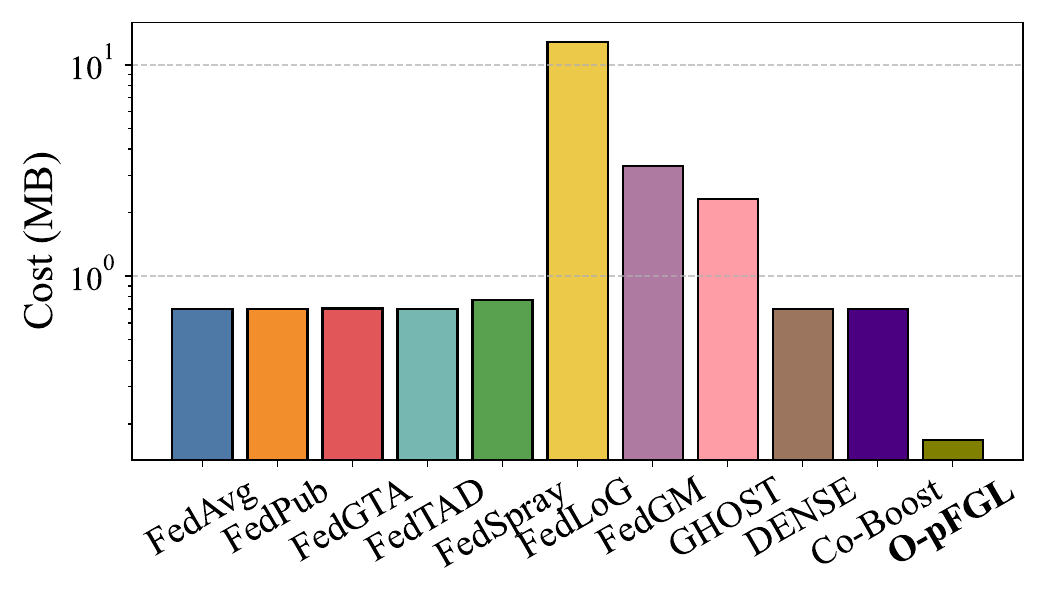}
    \caption{Communication costs.}
    \label{fig:cora_comm}
\end{figure}

\begin{figure*}[tbp]
\centerline{\includegraphics[width=0.99\textwidth]{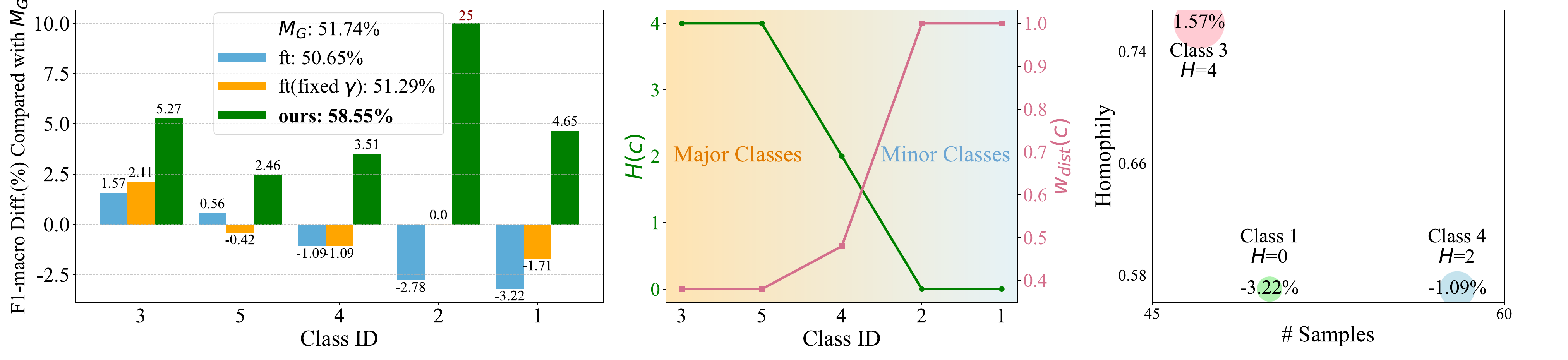}}
\caption{\textbf{Left}: Comparison of different personalized training methods. $M_G$ is the generalized model trained in Stage 1. \textbf{Middle}: The value of $H$ and $w_{dist}$ for each class. \textbf{Right}: The value of $H$ is determined by both samples quantity and homophily.}
\label{fig:citeseer_gap_H_major}
\end{figure*}

\subsection{Experimental Results}

\subsubsection{Performance Comparison}
We conduct experiments on various datasets with different properties to compare the performance. \textbf{(1)} Our method consistently outperforms on both accuracy and F1-macro, demonstrating the superiority of our method in deriving models with better personalization and generalization. As shown in Table~\ref{tab:transductive_louvain_10_clients}, most baselines are ineffective in one-shot communication under data heterogeneity. FedSpray and FedLoG focus on underrepresented classes, but they require numerous communications for optimization, offering little advantage in one-shot communication. FedGM and GHOST fail to handle severe heterogeneity, as biased gradients and skewed local data lead to inferior model performance. And they do not address the personalization issues. FedTAD, FedLOG, FedGM, and GHOST are not scalable and face out-of-memory problems on large-scale datasets. \textbf{(2)} The superiority of our method extends to heterophilic graph datasets, which is shown in Table~\ref{tab:heterophilic}. \textbf{(3)} Furthermore, our one-shot method achieves better average performance than multi-round federated methods that run for 100 rounds, which is shown in Appendix~\ref{sec:multi_rounds}. Full results for more clients, Metis partition, more datasets, and other prevalent GNN models (GraphSage, GAT, and SGC) are available in Appendix~\ref{sec:appendix_perf_comp}, Appendix~\ref{sec:more_datasets}, and Appendix~\ref{sec:appendix_other_gnn}.

\subsubsection{Model Heterogeneity}
We conduct experiments with heterogeneous client models. The results are available in Appendix~\ref{sec:appendix_model_hete}. Our method outperforms other methods.

\subsubsection{Inductive Performance}
We conduct experiments on inductive datasets, and the results are available in Appendix~\ref{sec:appendix_inductive}. Our method is scalable and outperforms.

\subsubsection{Ablation Study}

We compare our method with 4 variants: (1) ft: only fine-tuning $M_G$ to obtain $M_i$ on local graph data, (2) ft(focal): fine-tuning with focal loss~\cite{ross2017focal} to handle data imbalance, (3) ft(fixed $\gamma$): in the second stage, performing distillation with a fixed $\gamma_i$ for all nodes; (4) ours w.o. HRE: performing our method without HRE.

The results are shown in Figure~\ref{fig:louvain_ablation}. \textbf{(1)} The focal loss is ineffective on non-IID graph data since it's hard to determine proper weights. \textbf{(2)} Fixed $\gamma$ cannot balance improvements in accuracy and F1-macro as it cannot distinguish majority from minority. For major classes, the global knowledge of $M_G$ may have a negative impact, as shown in the left of Figure~\ref{fig:citeseer_gap_H_major}. This motivates our node adaptive distillation, which is more flexible in determining when to introduce global knowledge in fine-tuning. \textbf{(3)} HRE further improves performance via more precise estimations on homophilic graphs. Furthermore, we justify the rationale of considering homophily in determining the major classes by $H$ in Eq.~\ref{eq:H(c)} in the right of Figure~\ref{fig:citeseer_gap_H_major}. More analysis is in Appendix~\ref{sec:appendix_full_ablation_study}.

\subsubsection{Hyper-parameters Analysis}
We study our method's sensitivity to the hyper-parameters in Appendix~\ref{sec:appendix_hyperparamter_sensitive}. Our method is robust under a proper range of hyper-parameters values. We also provide guidance on selecting proper values.

\subsubsection{Communication Costs}
We analyze communication costs both theoretically and empirically. Our method involves uploading class-wise statistics ($O(Chf)$) and downloading the generated global surrogate graph ($O(n'f + n'^2)$). Here $n'$ is a small number detailed in Appendix~\ref{sec:app_hyper-parameter}. Both are independent of the model size. As empirically demonstrated on the Cora dataset in Figure~\ref{fig:cora_comm}, our method achieves the lowest communication cost. Furthermore, we introduce a server-efficient variant in Appendix~\ref{sec:appendix_sever_efficient_variant} to minimize server-side overhead.

\subsubsection{Complexity Analysis}
Additional complexity of our method comes from feature distribution estimation and surrogate graph generation. The former is extremely lightweight, requiring only a single forward pass, and can be offloaded to CPUs to ensure scalability. The surrogate graph generation is also scalable as the optimization is confined to a small, fixed-size graph (e.g., under 300 nodes in our experiments). This process is equivalent to forward and backward on this small graph. In our experiments, our optimization for the surrogate graph takes less than 2 minutes on the RTX 3090 GPU across all datasets. For comparison on the ogbn-arxiv dataset, FedTAD requires 6 minutes to calculate its diffusion matrix, GHOST needs 12 minutes to compute its Topology Consistency Criterion, and FedGM takes over 20 minutes for local graph synthesis. A formal theoretical complexity analysis is provided in Appendix~\ref{sec:appendix_complexity}.

\section{Conclusion}
In this paper, we introduce O-pFGL, the first one-shot personalized federated graph learning method for heterogeneous clients. It effectively balances personalization and generalization while supporting model heterogeneity and Secure Aggregation. In O-pFGL, clients first estimate class-wise feature statistics, which the server aggregates to generate a global surrogate graph without compromising raw data. Second, we introduce a novel two-stage personalized training with node-adaptive distillation, which effectively balances local knowledge with global insights. Extensive experiments demonstrate the state-of-the-art performance of our method across diverse and challenging settings.

\bibliography{aaai2026}

\appendix

\setcounter{secnumdepth}{2}


\section{Illustration for Determining {$\gamma_i$} in Node Adaptive Distillation}
\label{sec:illustration_gamma}
\begin{figure}[htbp]
\centerline{\includegraphics[width=0.98\columnwidth]{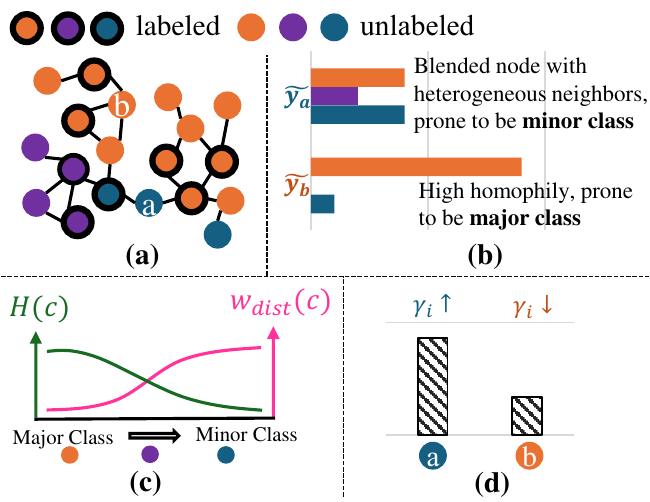}}
\caption{The process to determine $\gamma_i$ for each node in node adaptive distillation in the second stage.}
\label{fig:client_ada_per}
\end{figure}

We illustrate how to determine $\gamma_i$ for each node in each client's local graph. We pre-compute the vector $\mathbf{w}_{dist} \in \mathbb R^C$ by Eq.~\ref{eq:w_dist} and soft labels by Label Propagation in Section~\ref{sec:feat_dist_esti}. Considering the soft label of each node, if a node is predicted to class $c$ with a small $w_{dist}(c)$ (e.g. \textcolor{orange}{node $b$} in Figure~\ref{fig:client_ada_per}), it likely belongs to major classes with high homophily. In this case, $\gamma_i$ is small, and $M_i$ primarily learns through supervised fine-tuning; If a node is predicted to class $c$ with a large $w_{dist}(c)$ or has a blended prediction (e.g. \textcolor{blue}{node $a$} in Figure~\ref{fig:client_ada_per}), it likely belongs to minor classes or has low homophily. These nodes are often neglected during supervised fine-tuning. Thus $\gamma_i$ should be large to incorporate the corresponding global knowledge from $M_G$.

\section{Compatibility with Classical Secure Aggregation Implementation}
\label{sec:secure_agg_implementation}
The uploading and aggregation process of our method is compatible with the classical Secure Aggregation implementation since it can be decomposed into a series of weighted average operations. We show the pseudo-code of a classical implementation in Algorithm~\ref{alg:secure_agg_simple}.

\begin{algorithm}[H]
\caption{Implementation of Secure Aggregation for Our Federated Aggregation}
\label{alg:secure_agg_simple}
\textbf{Input}: A set of $m$ clients $\{Client_1, \dots, Client_m\}$. Each client $k$ has local data to compute $N_k^c, \boldsymbol{\mu}_k^c, \boldsymbol{s^2}_k^c$. \\
\textbf{Output}: $N^c, \boldsymbol{\mu}^c, \boldsymbol{s^2}^c$.
\begin{algorithmic}[1]
\STATE \textit{// Assume a key exchange protocol (e.g., Diffie-Hellman~\cite{li2010research}) has been run. Each pair of clients $(k, j)$ now shares a common secret key $s_{k,j}$.}

\STATE \textit{// --- \textbf{Masking on each Client $k$ (in parallel)} ---}
\STATE \textit{// Construct the private data vector $\boldsymbol{U}_k$}
\STATE $A_k^c \leftarrow N_k^c$; $\boldsymbol{B}_k^c \leftarrow N_k^c \boldsymbol{\mu}_k^c$; $\boldsymbol{C}_k^c \leftarrow (N_k^c - 1)\boldsymbol{s^2}_k^c$; $\boldsymbol{D}_k^c \leftarrow N_k^c(\boldsymbol{\mu}_k^c)^2$
\STATE $\boldsymbol{U}_k \leftarrow \text{concatenate}(A_k^c, \boldsymbol{B}_k^c, \boldsymbol{C}_k^c, \boldsymbol{D}_k^c)$
\STATE \textit{// Generate masks}
\STATE $\boldsymbol{b}_k \leftarrow \text{PRNG}(\text{random\_seed()})$ \COMMENT{Store personal mask $\boldsymbol{b}_k$}
\STATE $\boldsymbol{p}_k \leftarrow \boldsymbol{0}$
\FOR{$j=1$ to $m$}
    \IF{$j \neq k$}
        \STATE $\boldsymbol{r}_{k,j} \leftarrow \text{PRNG}(s_{k,j})$ \COMMENT{$s_{k,j}$ is the pre-shared secret with client $j$}
        \IF{$k < j$}
            \STATE $\boldsymbol{p}_k \leftarrow \boldsymbol{p}_k + \boldsymbol{r}_{k,j}$
        \ELSE
            \STATE $\boldsymbol{p}_k \leftarrow \boldsymbol{p}_k - \boldsymbol{r}_{k,j}$
        \ENDIF
    \ENDIF
\ENDFOR
\STATE \textit{// Create and send the final masked vector}
\STATE $\boldsymbol{y}_k \leftarrow \boldsymbol{U}_k + \boldsymbol{b}_k + \boldsymbol{p}_k$
\STATE Each client $k$ sends its masked vector $\boldsymbol{y}_k$ to the Server.

\STATE \textit{// --- \textbf{Aggregation on the Server} ---}
\STATE The Server collects all masked vectors $\{\boldsymbol{y}_1, \dots, \boldsymbol{y}_m\}$ and sums them: $\boldsymbol{Y} \leftarrow \sum_{k=1}^{m} \boldsymbol{y}_k$.
\STATE The Server collects all personal masks $\{\boldsymbol{b}_1, \dots, \boldsymbol{b}_m\}$ from the clients.
\STATE The Server sums the personal masks: $\boldsymbol{B}_{sum} \leftarrow \sum_{k=1}^{m} \boldsymbol{b}_k$.
\STATE The Server computes the true sum: $\boldsymbol{U}_{sum} \leftarrow \boldsymbol{Y} - \boldsymbol{B}_{sum}$.

\STATE \textit{// Final Statistics Computation by Server}
\STATE $(A_{sum}^c, \boldsymbol{B}_{sum}^c, \boldsymbol{C}_{sum}^c, \boldsymbol{D}_{sum}^c) \leftarrow \text{parse}(\boldsymbol{U}_{sum})$
\STATE $N^c \leftarrow A_{sum}^c$
\STATE $\boldsymbol{\mu}^c \leftarrow \boldsymbol{B}_{sum}^c / N^c$
\STATE $\Sigma_{within} \leftarrow \boldsymbol{C}_{sum}^c$
\STATE $\Sigma_{between} \leftarrow \boldsymbol{D}_{sum}^c - (\boldsymbol{B}_{sum}^c)^2 / N^c$
\STATE $\boldsymbol{s^2}^c \leftarrow \frac{1}{N^c - m} (\Sigma_{within} + \Sigma_{between})$
\STATE \textbf{return} $N^c, \boldsymbol{\mu}^c, \boldsymbol{s^2}^c$
\end{algorithmic}
\end{algorithm}

\section{Experimental Setup}
\label{sec:appendix_exp_setup}

\begin{table*}[htbp]
\setlength{\tabcolsep}{1mm}
\small
\centering
\begin{tabular}{llrrrrrcc}
\toprule
\textbf{Dataset} & \textbf{Task} & \textbf{Nodes} & \textbf{Edges} & \textbf{Features} & \textbf{Classes}  &
\textbf{Train/Val/Test} & \textbf{Description} & \textbf{Property} \\
\midrule  
Cora &  Transductive & 2,708 & 10,556 & 1433 & 7 & 20\%/40\%/40\% & Citation Network & Homophilic \\
CiteSeer &  Transductive  & 3,327 & 4,732 & 3703 & 6 & 20\%/40\%/40\% & Citation Network & Homophilic \\
PubMed &  Transductive  & 19,717 & 44,338 & 500 & 3 & 20\%/40\%/40\% & Citation Network & Homophilic \\
Ogbn-arxiv & Transductive  & 169,343 & 1,166,243 & 128 & 40  & 60\%/20\%/20\%   & Citation Network & Homophilic \\ 
Ogbn-products & Transductive  & 2,449,029 & 61,859,140	& 100 & 47  & 10\%/5\%/85\% & Co-purchase Network & Homophilic \\ 
Computers & Transductive  & 13,381 & 245,778 & 767 & 10  & 20\%/40\%/40\% & Co-purchase Network & Homophilic \\ 
Photo & Transductive  & 7,487 & 119,043	& 745 & 8  & 20\%/40\%/40\% & Co-purchase Network & Homophilic \\ 
CS & Transductive  & 18,333 & 81,894 & 6,805 & 15  & 20\%/40\%/40\% & Co-author Network & Homophilic \\ 
Physics & Transductive  & 34,493 & 247,962 & 8,415 & 5  & 20\%/40\%/40\% & Co-author Network & Homophilic \\ 
Actor & Transductive  & 7,600 & 30,019 & 932 & 5  & 50\%/25\%/25\% &  Actor Network & Heterophlic \\
genius & Transductive  & 421,961 & 984,979 & 12 & 2  & 50\%/25\%/25\% &  Actor Network & Heterophlic \\
\midrule
Flickr &  Inductive & 89,250 & 899,756 & 500 & 7 & 60\%/20\%/20\% & Image Network & Heterophilic \\
Reddit & Inductive & 232,965 & 114,615,892 & 602 & 41  & 80\%/10\%/10\% & Social Network & Homophilic \\ 
Reddit2 & Inductive & 232,965 & 23,213,838 & 602 & 41  & 65\%/10\%/25\%  & Social Network & Homophilic \\
\bottomrule
\end{tabular}
\caption{The statistics and description of the datasets.}
\label{tab:data}
\end{table*}

\paragraph{Dataset Description.}
We conduct experiments on 14 diverse real-world graph datasets, and the summary is listed in Table~\ref{tab:data}. Our datasets cover a wide range from 1) small-scale to large-scale, 2) transductive to inductive, 3) homophilic to heterophilic, and 4) academic domain to industry domain.

\paragraph{Hyper-parameters and Model Configurations.}
\label{sec:app_hyper-parameter} 
In our one-shot personalized federated graph learning setting, we limit the communication round of the 7 pFGL methods and the FedAvg method to a single round. For 2 OFL methods (DENSE and Co-Boost), the generators are trained to generate node features, and the topology structure is constructed using the $K$-Nearest Neighbors strategy as outlined in~\cite{zhu2024fedtad}. Note that we perform additional fine-tuning processes on FedAvg, FedTAD, FedGM, GHOST, DENSE, and Co-Boost to ensure better performance.

For each client, we employ a two-layer GCN~\cite{kipf2016semi} with the hidden layer dimension set to 64 by default. For baseline methods, the number of local training epochs is tuned from 1 to 100, and hyper-parameters are set as recommended in their respective papers or fine-tuned for optimal overall performance. For our method, we keep the size of the generated global surrogate graph as small as possible. On small-scale Cora, CiteSeer, and PubMed datasets, we set the number of nodes in each class in the global surrogate graph to 1. For other datasets, we set the number of nodes in each class to $p$ percent of the total number of nodes of the corresponding class in the training set. We set $p$ as 0.25\%, 0.04\%, 1\%, 1\%, 1\%, 0.5\%, 0.2\%, 0.05\%, 0.05\%, 2\%, 0.1\% on ogbn-arxiv, ogbn-products, Computers, Photo, CS, Physics, Flickr, Reddit, Reddit2, Actor, and genius datasets. We use the Adam optimizer and tune the learning rate. We tune $f_{th}$ within [0.95, 1] and tune $d_{th}$ according to the scale of the graphs. We tune the $\beta$ within [0.1, 1] to control the range of $\gamma_i$. For other hyper-parameters in global surrogate graph generation, we adopt the setting in~\cite{xiao2024simple}.

Under model heterogeneity, we adopt 10 different GNN models categorized into two types: coupled GNNs and decoupled GNNs. For coupled GNNs, we adopt 6 models, which are: 2-layer GCN with 64, 128, 256, and 512 hidden dimensions, and 3-layer GCN with 64 and 512 hidden dimensions. For decoupled GNNs, we adopt APPNP~\cite{gasteiger2018predict}, SSGC~\cite{zhu2021simple}, SGC ($K$=2) and SGC ($K$=4)~\cite{wu2019simplifying}.

For metrics, we employ both accuracy(\%) and F1-macro(\%) metrics to comprehensively evaluate the performance. F1-macro averages the F1 scores across all classes, providing a more robust measurement of the model’s generalization, particularly under imbalanced data. For each experiment, we report the mean and standard deviation across three runs.

\paragraph{Experimental Environments.} The experimental machine is equipped Intel(R) Xeon(R) CPU E5-2680 v4@2.40GHz, and 4$\times$NVIDIA GeForce RTX 3090. The operating system is Ubuntu 20.04, and the version of PyTorch is 2.1.2. We use PyG 2.5.2 in the implementation.

\begin{table*}[htbp]
\setlength{\tabcolsep}{1mm}
\small
\centering
\begin{tabular}{c|cc|cc|cc|cc|cc}
\toprule[1pt]
\multirow{2}{*}{Method} & \multicolumn{2}{c|}{\textbf{Cora}} & \multicolumn{2}{c|}{\textbf{CiteSeer}} & \multicolumn{2}{c|}{\textbf{PubMed}} & \multicolumn{2}{c|}{\textbf{ogbn-arxiv}} & \multicolumn{2}{c}{\textbf{ogbn-products}} \\ \cline{2-11}
 & Acc.           & F1            & Acc.           & F1            & Acc.           & F1            & Acc.           & F1            & Acc.           & F1            \\ \midrule
Standalone & 61.57{\scriptsize$\pm$0.84} & 30.35{\scriptsize$\pm$0.59} & 57.86{\scriptsize$\pm$0.62} & 36.44{\scriptsize$\pm$0.52} & 81.84{\scriptsize$\pm$0.04} & 57.95{\scriptsize$\pm$0.62} & 66.56{\scriptsize$\pm$0.16} & 13.53{\scriptsize$\pm$0.34} & 86.44{\scriptsize$\pm$0.01} & 17.34{\scriptsize$\pm$0.05} \\ 
FedAvg & 63.21{\scriptsize$\pm$0.45} & 31.77{\scriptsize$\pm$0.64} & 61.51{\scriptsize$\pm$0.27} & 40.60{\scriptsize$\pm$0.40} & 78.59{\scriptsize$\pm$0.14} & 41.60{\scriptsize$\pm$0.28} & 67.17{\scriptsize$\pm$0.12} & 15.33{\scriptsize$\pm$0.44} & 86.48{\scriptsize$\pm$0.01} & 17.57{\scriptsize$\pm$0.12} \\ \midrule
FedPUB & 62.52{\scriptsize$\pm$0.27} & 30.96{\scriptsize$\pm$0.25} & 59.40{\scriptsize$\pm$0.29} & 37.59{\scriptsize$\pm$0.19} & 81.41{\scriptsize$\pm$0.14} & 56.81{\scriptsize$\pm$1.94} & 65.83{\scriptsize$\pm$0.33} & 10.32{\scriptsize$\pm$0.53} & 83.47{\scriptsize$\pm$1.18} & 11.23{\scriptsize$\pm$0.74} \\ 
FedGTA & 37.03{\scriptsize$\pm$1.34} & 14.74{\scriptsize$\pm$0.76} & 62.02{\scriptsize$\pm$0.83} & 42.78{\scriptsize$\pm$0.92} & 64.01{\scriptsize$\pm$0.98} & 36.47{\scriptsize$\pm$1.33} & 57.59{\scriptsize$\pm$1.39} & 7.72{\scriptsize$\pm$0.67} & 79.04{\scriptsize$\pm$0.00} & 12.15{\scriptsize$\pm$0.10} \\ 
FedTAD & 63.16{\scriptsize$\pm$0.44} & 31.72{\scriptsize$\pm$0.52} & 62.32{\scriptsize$\pm$0.46} & 41.04{\scriptsize$\pm$0.55} & 78.64{\scriptsize$\pm$0.11} & 41.66{\scriptsize$\pm$0.16} & 68.90{\scriptsize$\pm$0.14} & 23.31{\scriptsize$\pm$0.22} & \multicolumn{2}{c}{OOM} \\ 
FedSpray & 60.15{\scriptsize$\pm$1.05} & 29.30{\scriptsize$\pm$0.61} & 55.43{\scriptsize$\pm$0.79} & 36.77{\scriptsize$\pm$0.63} & 80.28{\scriptsize$\pm$0.30} & 49.29{\scriptsize$\pm$0.73} & 68.88{\scriptsize$\pm$0.14} & 17.56{\scriptsize$\pm$0.27} & 85.43{\scriptsize$\pm$0.09} & 16.08{\scriptsize$\pm$0.10} \\ 
FedLoG & 60.92{\scriptsize$\pm$0.28} & 28.68{\scriptsize$\pm$0.12} & 58.29{\scriptsize$\pm$0.15} & 33.99{\scriptsize$\pm$0.34} & 79.73{\scriptsize$\pm$0.23} & 56.26{\scriptsize$\pm$0.52} & \multicolumn{2}{c|}{OOM} & \multicolumn{2}{c}{OOM} \\
FedGM & 60.68{\scriptsize$\pm$0.43} & 28.52{\scriptsize$\pm$0.26} & 58.24{\scriptsize$\pm$0.29} & 35.08{\scriptsize$\pm$0.27} & 72.61{\scriptsize$\pm$0.12} & 31.29{\scriptsize$\pm$0.22} & 48.92{\scriptsize$\pm$0.39} & 1.82{\scriptsize$\pm$0.05} & \multicolumn{2}{c}{OOM} \\
GHOST & 62.66{\scriptsize$\pm$0.20} & 31.22{\scriptsize$\pm$0.35} & 62.78{\scriptsize$\pm$0.23} & 43.72{\scriptsize$\pm$0.37} & 81.79{\scriptsize$\pm$0.05} & 58.78{\scriptsize$\pm$0.23} & 67.81{\scriptsize$\pm$0.29} & 16.75{\scriptsize$\pm$0.79} & \multicolumn{2}{c}{OOM} \\
\midrule
DENSE & 63.24{\scriptsize$\pm$0.32} & 31.74{\scriptsize$\pm$0.44} & 61.71{\scriptsize$\pm$0.21} & 40.69{\scriptsize$\pm$0.44} & 78.59{\scriptsize$\pm$0.15} & 41.63{\scriptsize$\pm$0.31} & 68.81{\scriptsize$\pm$0.04} & 22.97{\scriptsize$\pm$0.18} & 86.95{\scriptsize$\pm$0.01} & 21.14{\scriptsize$\pm$0.00} \\ 
Co-Boost & 63.21{\scriptsize$\pm$0.41} & 31.70{\scriptsize$\pm$0.49} & 61.89{\scriptsize$\pm$0.42} & 40.83{\scriptsize$\pm$0.54} & 78.67{\scriptsize$\pm$0.10} & 41.76{\scriptsize$\pm$0.27} & 66.77{\scriptsize$\pm$0.11} & 13.90{\scriptsize$\pm$0.14} & 86.43{\scriptsize$\pm$0.01} & 17.35{\scriptsize$\pm$0.14} \\ \midrule
O-pFGL & \textbf{69.84}{\scriptsize$\pm$1.76} & \textbf{49.37}{\scriptsize$\pm$1.98} & \textbf{68.69}{\scriptsize$\pm$0.04} & \textbf{50.26}{\scriptsize$\pm$0.80} & \textbf{82.04}{\scriptsize$\pm$0.20} & \textbf{58.79}{\scriptsize$\pm$0.68} & \textbf{70.61}{\scriptsize$\pm$0.05} & \textbf{32.28}{\scriptsize$\pm$0.68} & \textbf{87.03}{\scriptsize$\pm$0.00} & \textbf{21.87}{\scriptsize$\pm$0.05} \\ 
\bottomrule[1pt]
\end{tabular}%
\caption{Performance of methods on datasets under Louvain partition with 20 clients.}
\label{tab:transductive_louvain_20_clients}
\end{table*}

\section{More Experimental Results and Detailed Analysis}

\subsection{Performance Comparison}
\label{sec:appendix_perf_comp}
We first conduct experiments in the transductive learning setting and report the accuracy and F1-macro metrics. The datasets range from small-scale graphs to large-scale graphs with millions of nodes. We also vary the number of clients to evaluate the stability of methods. With 10 clients, the experimental results under Louvain and Metis partitions are shown in Table~\ref{tab:transductive_louvain_10_clients} and Table~\ref{tab:transductive_metis_10_clients}, respectively. With 20 clients, the experimental results under Louvain and Metis partitions are shown in Table~\ref{tab:transductive_louvain_20_clients} and Table~\ref{tab:transductive_metis_20_clients}, respectively. Our method consistently outperforms others in terms of accuracy and F1-macro. Under the non-IID scenarios, pFGL methods show marginal or even negative performance gain over Standalone, highlighting the ineffectiveness of current pFGL methods in personalizing models on non-IID graph data in a single communication round. The state-of-the-art OFL methods (DENSE and Co-Boost) also have difficulty generating high-quality pseudo-data by the ensemble model when each model is trained on non-IID local graph data. FedTAD encounters OOM when handling million-scale graphs due to its graph diffusion operation. FedLoG encounters OOM issues due to its node-wise $k$-hop subgraph construction. In contrast, our method remains scalable. FedGTA performs badly because its aggregation is based on the unreliable local models' predictions under non-IID data in a single round. FedSpray focuses on minor classes, but it requires numerous communications for optimization, offering little advantage over others in one-shot communication. FedLoG focuses on local generalization, but it cannot work well under severe non-IID scenarios within one-shot communication. FedGM synthesized graphs with gradient matching locally. But with biased local graph data, the gradients could also be skewed, which limits the effectiveness under heterogeneous client graph data. GHOST integrates aligned proxy models. However, the proxy models are aligned with skewed client graph data. Also, it does not address the personalized issues under non-IID graph data, leading to inferior performance under severe data non-IID scenarios. Moreover, FedGM and GHOST are not scalable. On large-scale datasets, the gradient matching consumes more GPU memory. GHOST needs to reconstruct a pseudo-graph with the same size as the local graph by feature matrix multiplication, which is of high complexity.

\begin{table*}[htbp]
\setlength{\tabcolsep}{1mm}
\small
\centering
\begin{tabular}{c|cc|cc|cc|cc|cc}
\toprule[1pt]
\multirow{2}{*}{Method} & \multicolumn{2}{c|}{\textbf{Cora}} & \multicolumn{2}{c|}{\textbf{CiteSeer}} & \multicolumn{2}{c|}{\textbf{PubMed}} & \multicolumn{2}{c|}{\textbf{ogbn-arxiv}} & \multicolumn{2}{c}{\textbf{ogbn-products}} \\ \cline{2-11}
 & Acc.           & F1            & Acc.           & F1            & Acc.           & F1            & Acc.           & F1            & Acc.           & F1            \\ \midrule
Standalone & 75.15{\scriptsize$\pm$0.08} & 31.00{\scriptsize$\pm$0.09} & 64.72{\scriptsize$\pm$0.37} & 37.36{\scriptsize$\pm$0.38} & 83.24{\scriptsize$\pm$0.09} & 56.61{\scriptsize$\pm$0.77} & 66.39{\scriptsize$\pm$0.26} & 24.05{\scriptsize$\pm$0.70} & 85.57{\scriptsize$\pm$0.02} & 23.54{\scriptsize$\pm$0.10} \\ 
FedAvg & 76.18{\scriptsize$\pm$0.14} & 31.42{\scriptsize$\pm$0.08} & 66.51{\scriptsize$\pm$0.34} & 35.89{\scriptsize$\pm$0.54} & 79.26{\scriptsize$\pm$0.09} & 33.75{\scriptsize$\pm$0.18} & 67.04{\scriptsize$\pm$0.13} & 26.06{\scriptsize$\pm$0.66} & 85.63{\scriptsize$\pm$0.01} & 24.16{\scriptsize$\pm$0.27} \\ \midrule
FedPUB & 75.31{\scriptsize$\pm$0.20} & 32.00{\scriptsize$\pm$0.53} & 64.94{\scriptsize$\pm$0.42} & 37.50{\scriptsize$\pm$0.57} & 82.69{\scriptsize$\pm$0.10} & 56.44{\scriptsize$\pm$1.21} & 64.86{\scriptsize$\pm$0.62} & 18.29{\scriptsize$\pm$1.14} & 84.43{\scriptsize$\pm$0.56} & 17.24{\scriptsize$\pm$0.64} \\ 
FedGTA & 53.85{\scriptsize$\pm$0.29} & 11.74{\scriptsize$\pm$0.66} & 64.03{\scriptsize$\pm$0.28} & 31.33{\scriptsize$\pm$0.54} & 78.57{\scriptsize$\pm$0.58} & 46.47{\scriptsize$\pm$0.94} & 55.90{\scriptsize$\pm$1.23} & 12.72{\scriptsize$\pm$1.53} & 74.36{\scriptsize$\pm$0.34} & 16.67{\scriptsize$\pm$0.19} \\ 
FedTAD & 76.36{\scriptsize$\pm$0.26} & 31.99{\scriptsize$\pm$0.22} & 66.76{\scriptsize$\pm$0.26} & 36.06{\scriptsize$\pm$0.42} & 79.41{\scriptsize$\pm$0.07} & 34.19{\scriptsize$\pm$0.13} & 68.96{\scriptsize$\pm$0.28} & 34.91{\scriptsize$\pm$0.20} & \multicolumn{2}{c}{OOM} \\ 
FedSpray & 74.31{\scriptsize$\pm$0.34} & 31.95{\scriptsize$\pm$0.69} & 62.49{\scriptsize$\pm$0.85} & 36.04{\scriptsize$\pm$0.54} & 81.02{\scriptsize$\pm$0.56} & 60.97{\scriptsize$\pm$1.27} & 68.21{\scriptsize$\pm$0.18} & 26.57{\scriptsize$\pm$0.17} & 86.45{\scriptsize$\pm$0.05} & 22.02{\scriptsize$\pm$0.06} \\ 
FedLoG & 75.08{\scriptsize$\pm$0.21} & 23.75{\scriptsize$\pm$0.35} & 64.92{\scriptsize$\pm$0.36} & 33.72{\scriptsize$\pm$0.58} & 75.52{\scriptsize$\pm$6.73} & 48.26{\scriptsize$\pm$4.12} & \multicolumn{2}{c|}{OOM} & \multicolumn{2}{c}{OOM} \\ 
FedGM & 75.41{\scriptsize$\pm$0.11} & 23.43{\scriptsize$\pm$1.18} & 65.81{\scriptsize$\pm$0.37} & 36.67{\scriptsize$\pm$0.42} & 76.41{\scriptsize$\pm$0.37} & 29.84{\scriptsize$\pm$0.43} & 45.48{\scriptsize$\pm$1.04} & 1.91{\scriptsize$\pm$0.27} & \multicolumn{2}{c}{OOM} \\ 
GHOST & 76.63{\scriptsize$\pm$0.07} & 31.68{\scriptsize$\pm$0.05} & 64.92{\scriptsize$\pm$0.07} & 37.49{\scriptsize$\pm$0.12} & 83.56{\scriptsize$\pm$0.04} & 52.32{\scriptsize$\pm$0.96} & 68.22{\scriptsize$\pm$0.22} & 22.43{\scriptsize$\pm$0.48} & \multicolumn{2}{c}{OOM} \\ 
\midrule
DENSE & 76.21{\scriptsize$\pm$0.08} & 31.64{\scriptsize$\pm$0.24} & 65.62{\scriptsize$\pm$0.27} & 37.88{\scriptsize$\pm$0.24} & 79.29{\scriptsize$\pm$0.05} & 33.81{\scriptsize$\pm$0.09} & 68.89{\scriptsize$\pm$0.04} & 34.34{\scriptsize$\pm$0.34} & 86.43{\scriptsize$\pm$0.03} & 28.42{\scriptsize$\pm$0.25} \\ 
Co-Boost & 76.54{\scriptsize$\pm$0.31} & 31.90{\scriptsize$\pm$0.36} & 64.55{\scriptsize$\pm$0.73} & 31.37{\scriptsize$\pm$0.75} & 76.56{\scriptsize$\pm$0.36} & 29.89{\scriptsize$\pm$0.34} & 66.22{\scriptsize$\pm$0.02} & 23.58{\scriptsize$\pm$0.18} & 85.59{\scriptsize$\pm$0.03} & 23.63{\scriptsize$\pm$0.25} \\ \midrule
O-pFGL & \textbf{81.79}{\scriptsize$\pm$0.23} & \textbf{50.85}{\scriptsize$\pm$1.27} & \textbf{72.76}{\scriptsize$\pm$0.24} & \textbf{50.94}{\scriptsize$\pm$0.80} & \textbf{83.99}{\scriptsize$\pm$0.18} & \textbf{61.03}{\scriptsize$\pm$1.21} & \textbf{71.06}{\scriptsize$\pm$0.09} & \textbf{41.35}{\scriptsize$\pm$0.37} & \textbf{86.90}{\scriptsize$\pm$0.01} & \textbf{28.90}{\scriptsize$\pm$0.05} \\ 
\bottomrule[1pt]
\end{tabular}%
\caption{Performance of methods on datasets under Metis partition with 10 clients.}
\label{tab:transductive_metis_10_clients}
\end{table*}

In addition to improving accuracy, our method consistently achieves a significant increase in F1-macro, an often overlooked metric in previous studies. This improvement highlights that our method enhances both personalization and generalization, resulting in more robust models.

We also compare our method with ReNode~\cite{chen2021topology}, a representative graph imbalance learning method, which aims to train robust graph models under imbalanced graph data. The comparison results are shown in Table~\ref{tab:compare_renode}. The results show that ReNode cannot effectively model the imbalanced graph data in our scenarios.

\begin{table}[htbp]
\centering
\resizebox{\columnwidth}{!}{%
\begin{tabular}{c|cc|cc|cc}
\toprule[1pt]
\multirow{2}{*}{Louvain} & \multicolumn{2}{c|}{\textbf{Cora}}           & \multicolumn{2}{c|}{\textbf{CiteSeer}}       & \multicolumn{2}{c}{\textbf{PubMed}} \\ \cline{2-7}  \rule{0pt}{2ex}
  & Acc.           & F1            & Acc.           & F1            & Acc.           & F1             \\ \midrule
ReNode & 68.07{\scriptsize$\pm$0.29} & 41.70{\scriptsize$\pm$0.57} & 61.57{\scriptsize$\pm$0.25} & 41.91{\scriptsize$\pm$0.26} & 74.64{\scriptsize$\pm$0.36} & 35.59{\scriptsize$\pm$0.69} \\ 

O-pFGL & \textbf{76.43}{\scriptsize$\pm$1.24} & \textbf{61.58}{\scriptsize$\pm$2.16} &  \textbf{71.61}{\scriptsize$\pm$0.40} & \textbf{58.24}{\scriptsize$\pm$0.96} &\textbf{82.71}{\scriptsize$\pm$0.03} & \textbf{66.40}{\scriptsize$\pm$0.34} \\ \midrule[1pt]

\multirow{2}{*}{Metis} & \multicolumn{2}{c|}{\textbf{Cora}}           & \multicolumn{2}{c|}{\textbf{CiteSeer}}       & \multicolumn{2}{c}{\textbf{PubMed}} \\ \cline{2-7}  \rule{0pt}{2ex}
  & Acc.           & F1            & Acc.           & F1            & Acc.           & F1             \\ \midrule
  
ReNode & 76.44{\scriptsize$\pm$0.06} & 27.69{\scriptsize$\pm$0.45} & 65.89{\scriptsize$\pm$0.23} & 37.74{\scriptsize$\pm$0.15} & 78.81{\scriptsize$\pm$0.14} & 32.94{\scriptsize$\pm$0.18} \\ 

O-pFGL & \textbf{81.79}{\scriptsize$\pm$0.23} & \textbf{50.85}{\scriptsize$\pm$1.27} &  \textbf{72.76}{\scriptsize$\pm$0.24} & \textbf{50.94}{\scriptsize$\pm$0.80} &\textbf{83.99}{\scriptsize$\pm$0.18} & \textbf{61.03}{\scriptsize$\pm$1.21} \\

\bottomrule[1pt]
\end{tabular}%
}
\caption{Comparison with the graph imbalance learning method.}
\label{tab:compare_renode}
\end{table}

\begin{table*}[htbp]
\setlength{\tabcolsep}{1mm}
\small
\centering
\begin{tabular}{c|cc|cc|cc|cc|cc}
\toprule[1pt]
\multirow{2}{*}{Method} & \multicolumn{2}{c|}{\textbf{Cora}} & \multicolumn{2}{c|}{\textbf{CiteSeer}} & \multicolumn{2}{c|}{\textbf{PubMed}} & \multicolumn{2}{c|}{\textbf{ogbn-arxiv}} & \multicolumn{2}{c}{\textbf{ogbn-products}} \\ \cline{2-11}
 & Acc.           & F1            & Acc.           & F1            & Acc.           & F1            & Acc.           & F1            & Acc.           & F1            \\ \midrule
Standalone & 71.57{\scriptsize$\pm$0.29} & 20.89{\scriptsize$\pm$0.06} & 61.36{\scriptsize$\pm$0.27} & 26.19{\scriptsize$\pm$0.40} & 83.24{\scriptsize$\pm$0.05} & 55.21{\scriptsize$\pm$0.35} & 66.77{\scriptsize$\pm$0.17} & 19.76{\scriptsize$\pm$0.49} & 86.44{\scriptsize$\pm$0.00} & 18.32{\scriptsize$\pm$0.14} \\ 
FedAvg & 73.14{\scriptsize$\pm$0.22} & 18.38{\scriptsize$\pm$0.39} & 62.62{\scriptsize$\pm$0.48} & 28.48{\scriptsize$\pm$0.38} & 79.86{\scriptsize$\pm$0.09} & 34.81{\scriptsize$\pm$0.05} & 67.54{\scriptsize$\pm$0.10} & 21.68{\scriptsize$\pm$0.16} & 86.52{\scriptsize$\pm$0.00} & 19.04{\scriptsize$\pm$0.09} \\ \midrule
FedPUB & 72.32{\scriptsize$\pm$0.32} & 20.16{\scriptsize$\pm$0.82} & 60.73{\scriptsize$\pm$0.14} & 29.40{\scriptsize$\pm$0.16} & 82.98{\scriptsize$\pm$0.21} & 50.51{\scriptsize$\pm$0.26} & 65.41{\scriptsize$\pm$0.34} & 14.45{\scriptsize$\pm$0.59} & 83.94{\scriptsize$\pm$1.18} & 11.66{\scriptsize$\pm$0.60} \\ 
FedGTA & 55.04{\scriptsize$\pm$0.14} & 10.60{\scriptsize$\pm$0.07} & 59.43{\scriptsize$\pm$0.40} & 20.04{\scriptsize$\pm$0.82} & 74.34{\scriptsize$\pm$0.36} & 35.18{\scriptsize$\pm$0.61} & 58.76{\scriptsize$\pm$0.74} & 10.77{\scriptsize$\pm$0.44} & 78.35{\scriptsize$\pm$0.06} & 13.81{\scriptsize$\pm$0.26} \\ 
FedTAD & 73.55{\scriptsize$\pm$0.37} & 18.78{\scriptsize$\pm$0.31} & 62.60{\scriptsize$\pm$0.40} & 28.44{\scriptsize$\pm$0.20} & 77.42{\scriptsize$\pm$0.47} & 31.17{\scriptsize$\pm$0.52} & 69.29{\scriptsize$\pm$0.07} & 29.84{\scriptsize$\pm$0.34} & \multicolumn{2}{c}{OOM} \\ 
FedSpray & 71.44{\scriptsize$\pm$0.04} & 21.49{\scriptsize$\pm$0.15} & 58.98{\scriptsize$\pm$1.87} & 29.35{\scriptsize$\pm$1.52} & 81.05{\scriptsize$\pm$0.33} & 40.51{\scriptsize$\pm$0.56} & 68.86{\scriptsize$\pm$0.44} & 22.49{\scriptsize$\pm$0.44} & 85.28{\scriptsize$\pm$0.05} & 17.31{\scriptsize$\pm$0.04} \\
FedLoG & 73.79{\scriptsize$\pm$0.12} & 18.72{\scriptsize$\pm$0.30} & 61.66{\scriptsize$\pm$0.15} & 26.16{\scriptsize$\pm$0.35} & 80.35{\scriptsize$\pm$0.83} & 49.59{\scriptsize$\pm$0.64} & \multicolumn{2}{c|}{OOM} & \multicolumn{2}{c}{OOM} \\
FedGM & 73.20{\scriptsize$\pm$0.24} & 18.97{\scriptsize$\pm$0.20} & 62.10{\scriptsize$\pm$0.14} & 28.69{\scriptsize$\pm$0.02} & 77.50{\scriptsize$\pm$0.32} & 31.06{\scriptsize$\pm$0.30} & 49.13{\scriptsize$\pm$0.35} & 1.69{\scriptsize$\pm$0.06} & \multicolumn{2}{c}{OOM} \\ 
GHOST & 73.55{\scriptsize$\pm$0.37} & 18.78{\scriptsize$\pm$0.31} & 62.60{\scriptsize$\pm$0.40} & 28.44{\scriptsize$\pm$0.20} & 77.42{\scriptsize$\pm$0.47} & 31.17{\scriptsize$\pm$0.52} & 69.29{\scriptsize$\pm$0.07} & 29.84{\scriptsize$\pm$0.34} & \multicolumn{2}{c}{OOM} \\ 
\midrule
DENSE & 72.62{\scriptsize$\pm$0.17} & 20.19{\scriptsize$\pm$0.13} & 61.36{\scriptsize$\pm$0.19} & 29.66{\scriptsize$\pm$0.20} & 79.94{\scriptsize$\pm$0.04} & 34.87{\scriptsize$\pm$0.07} & 69.21{\scriptsize$\pm$0.15} & 29.31{\scriptsize$\pm$0.54} & 86.97{\scriptsize$\pm$0.00} & 22.94{\scriptsize$\pm$0.12} \\ 
Co-Boost & 73.42{\scriptsize$\pm$0.06} & 18.45{\scriptsize$\pm$0.24} & 61.30{\scriptsize$\pm$0.56} & 25.87{\scriptsize$\pm$0.78} & 79.98{\scriptsize$\pm$0.06} & 35.04{\scriptsize$\pm$0.27} & 66.72{\scriptsize$\pm$0.06} & 19.48{\scriptsize$\pm$0.28} & 86.43{\scriptsize$\pm$0.01} & 18.51{\scriptsize$\pm$0.24} \\ \midrule
O-pFGL & \textbf{78.31}{\scriptsize$\pm$0.88} & \textbf{37.21}{\scriptsize$\pm$1.96} & \textbf{68.88}{\scriptsize$\pm$0.22} & \textbf{41.95}{\scriptsize$\pm$0.25} & \textbf{83.94}{\scriptsize$\pm$0.05} & \textbf{56.61}{\scriptsize$\pm$1.36} & \textbf{70.87}{\scriptsize$\pm$0.18} & \textbf{36.06}{\scriptsize$\pm$0.43} & \textbf{87.15}{\scriptsize$\pm$0.02} & \textbf{23.41}{\scriptsize$\pm$0.26} \\ 
\bottomrule[1pt]
\end{tabular}%
\caption{Performance of methods on datasets under Metis partition with 20 clients.}
\label{tab:transductive_metis_20_clients}
\end{table*}

\subsection{Compared with Multi-Rounds FL Methods}
\label{sec:multi_rounds}
To demonstrate the superiority of our method, we further compare our method with multi-round federated learning methods, including FedAvg, FedPUB, FedGTA, FedTAD, FedSpray, FedLoG, and FedGM. We set the communication round of these 6 methods to 100. The experimental results on 5 multi-scale graph datasets under Louvain and Metis partitions are shown in Table~\ref{tab:multiround_louvain_10_clients} and Table~\ref{tab:multiround_metis_10_clients}, respectively. Our method still outperforms in most cases and remains the best average performance.

\begin{table*}[tbp]
\centering
\resizebox{\textwidth}{!}{%
\begin{tabular}{c|cc|cc|cc|cc|cc|cc}
\toprule[1pt]
\multirow{2}{*}{Method} & \multicolumn{2}{c|}{\textbf{Cora}} & \multicolumn{2}{c|}{\textbf{CiteSeer}} & \multicolumn{2}{c|}{\textbf{PubMed}} & \multicolumn{2}{c|}{\textbf{ogbn-arxiv}} & \multicolumn{2}{c|}{\textbf{ogbn-products}} & \multicolumn{2}{c}{\textbf{Average}}  \\ \cline{2-13}
 & Acc.           & F1            & Acc.           & F1            & Acc.           & F1            & Acc.           & F1            & Acc.           & F1       & Acc.           & F1     \\ \midrule
FedAvg & 75.27{\scriptsize$\pm$0.17} & 61.74{\scriptsize$\pm$0.21} & 67.11{\scriptsize$\pm$0.48} & 51.86{\scriptsize$\pm$0.81} & 85.22{\scriptsize$\pm$0.14} & 71.17{\scriptsize$\pm$0.22} & 68.10{\scriptsize$\pm$0.07} & 30.72{\scriptsize$\pm$0.14} & 86.58{\scriptsize$\pm$0.02} & \textbf{28.28}{\scriptsize$\pm$0.07} & 76.46 & 48.75 \\ \midrule
FedPUB & \textbf{78.32}{\scriptsize$\pm$0.37} & \textbf{64.59}{\scriptsize$\pm$0.80} & 69.00{\scriptsize$\pm$0.74} & 56.43{\scriptsize$\pm$0.68} & 83.12{\scriptsize$\pm$0.08} & 70.18{\scriptsize$\pm$0.36} & 65.34{\scriptsize$\pm$0.28} & 12.50{\scriptsize$\pm$0.20} & 85.18{\scriptsize$\pm$0.14} & 16.31{\scriptsize$\pm$0.46} & 76.19 & 44.00 \\ 
FedGTA & 73.32{\scriptsize$\pm$0.17} & 60.08{\scriptsize$\pm$0.32} & 67.12{\scriptsize$\pm$1.58} & 52.68{\scriptsize$\pm$1.81} & 84.39{\scriptsize$\pm$0.08} & 72.06{\scriptsize$\pm$0.12} & 58.30{\scriptsize$\pm$0.30} & 12.22{\scriptsize$\pm$0.26} & 83.25{\scriptsize$\pm$0.17} & 24.71{\scriptsize$\pm$0.10} & 73.28 & 44.35 \\ 
FedTAD & 74.62{\scriptsize$\pm$0.44} & 60.37{\scriptsize$\pm$1.01} & 67.68{\scriptsize$\pm$0.39} & 53.43{\scriptsize$\pm$0.82} & \textbf{84.70}{\scriptsize$\pm$0.21} & 68.56{\scriptsize$\pm$0.72} & 68.77{\scriptsize$\pm$0.12} & 27.33{\scriptsize$\pm$0.50} & \multicolumn{2}{c|}{OOM} & N/A & N/A \\ 
FedSpray & 70.08{\scriptsize$\pm$0.41} & 56.46{\scriptsize$\pm$1.30} & 68.31{\scriptsize$\pm$0.88} & 55.07{\scriptsize$\pm$0.94} & 84.04{\scriptsize$\pm$0.10} & \textbf{74.06}{\scriptsize$\pm$0.29} & 70.20{\scriptsize$\pm$0.08} & 34.09{\scriptsize$\pm$0.33} & 84.06{\scriptsize$\pm$0.11} & 20.80{\scriptsize$\pm$0.05} & 75.34 & 48.10 \\ 
FedLoG & 69.08{\scriptsize$\pm$1.96} & 53.80{\scriptsize$\pm$1.43} & 61.19{\scriptsize$\pm$0.33} & 39.42{\scriptsize$\pm$0.36} & 79.40{\scriptsize$\pm$0.62} & 57.33{\scriptsize$\pm$0.47} & \multicolumn{2}{c|}{OOM} & \multicolumn{2}{c|}{OOM} & N/A & N/A \\ 
FedGM & 69.39{\scriptsize$\pm$0.56} & 44.04{\scriptsize$\pm$0.55} & 61.70{\scriptsize$\pm$0.24} & 42.38{\scriptsize$\pm$0.45} & 79.30{\scriptsize$\pm$0.29} & 44.69{\scriptsize$\pm$0.39} & 54.65{\scriptsize$\pm$0.69} & 8.65{\scriptsize$\pm$0.60} & \multicolumn{2}{c|}{OOM} & N/A & N/A \\
\midrule
O-pFGL & 76.43{\scriptsize$\pm$1.24} & 61.58{\scriptsize$\pm$2.16} & \textbf{71.61}{\scriptsize$\pm$0.40} & \textbf{58.24}{\scriptsize$\pm$0.96} & 82.71{\scriptsize$\pm$0.03} & 66.40{\scriptsize$\pm$0.34} & \textbf{70.93}{\scriptsize$\pm$0.59} & \textbf{36.66}{\scriptsize$\pm$0.22} & \textbf{86.63}{\scriptsize$\pm$0.02} & 27.21{\scriptsize$\pm$0.15} & \textbf{77.66} & \textbf{50.02} \\ 
\bottomrule[1pt]
\end{tabular}%
}
\caption{Comparison with FL methods with 100 communication rounds under Louvain partition.}
\label{tab:multiround_louvain_10_clients}
\end{table*}

\begin{table*}[tbp]
\centering
\resizebox{\textwidth}{!}{%
\begin{tabular}{c|cc|cc|cc|cc|cc|cc}
\toprule[1pt]
\multirow{2}{*}{Method} & \multicolumn{2}{c|}{\textbf{Cora}} & \multicolumn{2}{c|}{\textbf{CiteSeer}} & \multicolumn{2}{c|}{\textbf{PubMed}} & \multicolumn{2}{c|}{\textbf{ogbn-arxiv}} & \multicolumn{2}{c|}{\textbf{ogbn-products}} & \multicolumn{2}{c}{\textbf{Average}} \\ \cline{2-13}
 & Acc.           & F1            & Acc.           & F1            & Acc.           & F1            & Acc.           & F1            & Acc.           & F1       & Acc.           & F1      \\ \midrule

FedAvg & 77.10{\scriptsize$\pm$0.30} & 45.59{\scriptsize$\pm$1.81} & 67.30{\scriptsize$\pm$0.30} & 42.88{\scriptsize$\pm$0.10} & 83.70{\scriptsize$\pm$0.03} & 66.65{\scriptsize$\pm$0.18} & 68.76{\scriptsize$\pm$0.17} & 36.77{\scriptsize$\pm$0.07} & 86.80{\scriptsize$\pm$0.01} & \textbf{30.02}{\scriptsize$\pm$0.10} & 76.73 & 44.38 \\ \midrule
FedPUB & 76.31{\scriptsize$\pm$0.61} & 37.22{\scriptsize$\pm$1.68} & 66.07{\scriptsize$\pm$1.05} & 40.65{\scriptsize$\pm$0.97} & 84.66{\scriptsize$\pm$1.04} & 68.19{\scriptsize$\pm$1.51} & 64.86{\scriptsize$\pm$0.62} & 18.29{\scriptsize$\pm$1.14} & 85.83{\scriptsize$\pm$0.35} & 19.45{\scriptsize$\pm$0.35} & 75.55 & 36.76 \\ 
FedGTA & 75.67{\scriptsize$\pm$0.52} & 44.91{\scriptsize$\pm$1.54} & 64.03{\scriptsize$\pm$0.28} & 31.33{\scriptsize$\pm$0.54} & \textbf{85.04}{\scriptsize$\pm$0.06} & 65.87{\scriptsize$\pm$0.32} & 55.90{\scriptsize$\pm$1.23} & 12.72{\scriptsize$\pm$1.53} & 74.36{\scriptsize$\pm$0.34} & 16.67{\scriptsize$\pm$0.19} & 71.00 & 34.30 \\ 
FedTAD & 77.33{\scriptsize$\pm$0.44} & 45.49{\scriptsize$\pm$0.51} & 66.76{\scriptsize$\pm$0.26} & 36.06{\scriptsize$\pm$0.42} & 79.41{\scriptsize$\pm$0.07} & 34.19{\scriptsize$\pm$0.13} & 68.96{\scriptsize$\pm$0.28} & 34.91{\scriptsize$\pm$0.20} & \multicolumn{2}{c|}{OOM} & N/A & N/A \\ 
FedSpray & 74.31{\scriptsize$\pm$0.34} & 31.95{\scriptsize$\pm$0.69} & 66.40{\scriptsize$\pm$0.68} & 49.06{\scriptsize$\pm$1.61} & 84.17{\scriptsize$\pm$0.14} & \textbf{70.45}{\scriptsize$\pm$0.34} & 70.12{\scriptsize$\pm$0.15} & 41.26{\scriptsize$\pm$0.26} & 86.45{\scriptsize$\pm$0.05} & 22.02{\scriptsize$\pm$0.06} & 76.29 & 42.95 \\ 
FedLoG & 75.08{\scriptsize$\pm$0.21} & 23.75{\scriptsize$\pm$0.35} & 64.92{\scriptsize$\pm$0.36} & 33.72{\scriptsize$\pm$0.58} & 75.52{\scriptsize$\pm$6.74} & 48.26{\scriptsize$\pm$4.12} & \multicolumn{2}{c|}{OOM} & \multicolumn{2}{c|}{OOM} & N/A & N/A \\ 
FedGM & 76.92{\scriptsize$\pm$0.22} & 31.66{\scriptsize$\pm$0.40} & 65.81{\scriptsize$\pm$0.37} & 36.67{\scriptsize$\pm$0.42} & 79.86{\scriptsize$\pm$0.31} & 36.73{\scriptsize$\pm$0.70} & 50.15{\scriptsize$\pm$0.48} & 8.55{\scriptsize$\pm$0.74} & \multicolumn{2}{c|}{OOM} & N/A & N/A \\ 
\midrule
O-pFGL & \textbf{81.79}{\scriptsize$\pm$0.23} & \textbf{50.85}{\scriptsize$\pm$1.27} & \textbf{72.76}{\scriptsize$\pm$0.24} & \textbf{50.94}{\scriptsize$\pm$0.80} & 83.99{\scriptsize$\pm$0.18} & 61.03{\scriptsize$\pm$1.21} & \textbf{71.06}{\scriptsize$\pm$0.09} & \textbf{41.35}{\scriptsize$\pm$0.37} & \textbf{86.90}{\scriptsize$\pm$0.01} & 28.90{\scriptsize$\pm$0.05} & \textbf{79.30} & \textbf{46.61} \\ 
\bottomrule[1pt]
\end{tabular}%
}
\caption{Comparison with FL methods with 100 communication rounds under Metis partition.}
\label{tab:multiround_metis_10_clients}
\end{table*}

\subsection{Detailed Ablation Study}
\label{sec:appendix_full_ablation_study}
We conduct the ablation study to evaluate the performance gain of our method. We compare our method with 4 variants: (1) ft: only fine-tuning $M_G$ to obtain $M_i$ on local graph data, (2) ft(focal): fine-tuning with focal loss~\cite{ross2017focal} to handle data imbalance, (3) ft(fixed $\gamma$): in the second stage, performing distillation with a fixed $\gamma$ for all nodes; (4) ours w.o. HRE: performing node adaptive distillation along with fine-tuning, without HRE.

\begin{figure}[tbp]
    \centering
    \includegraphics[width=\linewidth]{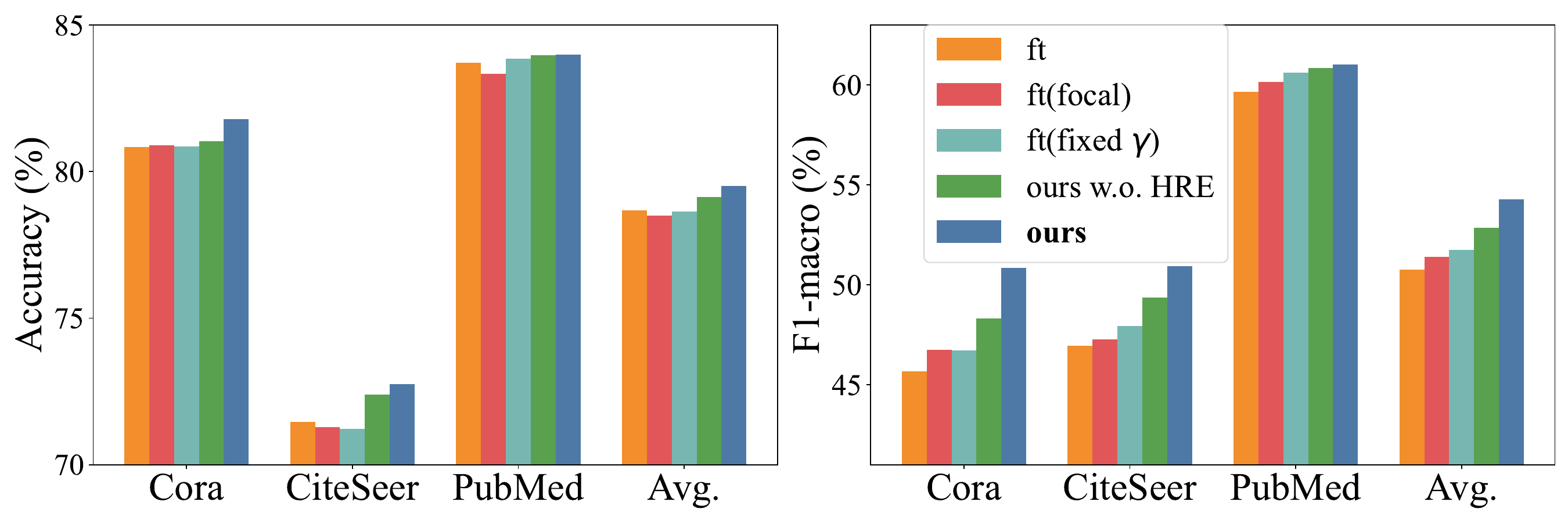}
    \caption{Ablation study under Metis partition.}
    \label{fig:metis_ablation}
\end{figure}

Figure~\ref {fig:louvain_ablation} and Figure~\ref {fig:metis_ablation} show our experimental results of accuracy and F1-macro under Louvain and Metis partition, respectively. We make the following 4 observations. \textbf{(O1)} Comparing ft and ft(focal), the focal loss in ft(focal) cannot effectively address the non-IID graph data. We believe the reasons are that it's hard to determine the proper weights under extreme non-IID scenarios. \textbf{(O2)} Comparing ft and ft(fixed $\gamma$), a fixed $\gamma$ in distillation cannot balance the improvements of accuracy and F1-macro. We believe the reason is that it does not distinguish the majority and minority in the distillation. For nodes in major classes with high homophily, the global knowledge of $M_G$ may have a negative impact, as demonstrated by the left panel in Figure~\ref{fig:citeseer_gap_H_major}. \textbf{(O3)} Comparing ft and ours w.o. HRE, the node adaptive distillation is more flexible in determining when to introduce the global knowledge of $M_G$ in fine-tuning. Thus ours w.o. HRE achieves better accuracy and F1-macro. \textbf{(O4)} Comparing ours w.o. HRE and ours, the HRE strategy further improves performance by generating a higher-quality global surrogate graph through more precise estimations.

We make a further ablation at the client level in Figure~\ref{fig:citeseer_gap_H_major}, evaluating different methods on a client using the CiteSeer dataset with the Louvain partition. As shown in the middle figure, class 3 and class 5 are regarded as major classes due to their larger quantity and relatively high homophily, while class 2 and class 1 are regarded as minor classes with fewer nodes and lower homophily. We make the following 3 observations. \textbf{(O1)} Compared to $M_G$, the ft method shows performance gain on major class 3 (+1.57\%) and class 5 (+0.56\%) but degrades performance on minor class 4 (-1.09\%), class 2 (-2.78\%), and class 1 (-3.22\%). The performance differences after fine-tuning, shown in the left figure, align with the $H$ values in the middle figure, highlighting the limitations of fine-tuning. \textbf{(O2)} The ft(fixed $\gamma$) method cannot handle this limitation effectively. With a fixed $\gamma$, the performance on major class 5 worsens (-0.42\%), and the performance on minor classes still degrades. This phenomenon motivates us to develop node-adaptive distillation. \textbf{(O3)} Our method, with Class-aware Distillation Factor $w_{dist}$ and node adaptive distillation, boosts performance on all classes by more than +2.4\%.

We further justify the rationale for considering homophily in determining the major classes by $H$ values in Eq.~\ref{eq:H(c)}. The right panel of Figure~\ref{fig:citeseer_gap_H_major} shows that while class 1 and class 4 have slightly more nodes, their average homophily is low ($<$ 0.6), leading to less compact communities in the local graph. As a result, they are often neglected during fine-tuning, causing performance drops of -3.22\% and -1.09\%, respectively. In contrast, class 3 exhibits high average homophily ($>$ 0.7), forming compact connections, allowing it to dominate the fine-tuning process with a performance improvement of +1.57\%.

\subsection{Different Adjacency Matrix Implementations}
\label{sec:adj_design}
In our method, we derive the adjacency matrix $A'$ of the generated global surrogate graph by Eq.~\ref{eq:adj_design}. We implement the $g_\theta$ as a simple 3-layer MLP with input dimensions $2\times f$ ($f$ is the node feature dimension) and output dimension is 1 (output $\in [0, 1]$, represents the probability of existence of a link). Here we compare the other two different implementations of $g_{\theta}$: (1) make $A'$ a learnable matrix and directly optimize $A'$, (2) $A'$ is determined by the dot product of (normalized) node features. The experimental results are shown in Table~\ref{tab:diff_adj_design}. The advantages of the other two designs cannot be generalized to most datasets. Our design, in contrast, offers flexibility via the learnable MLP and $\delta$ value, thus showing overall performance advantages. Note that when setting $\delta=1$, our design falls back to derive a global surrogate graph where all nodes are isolated.

\begin{table*}[tbp]
\setlength{\tabcolsep}{1mm}
\small
\centering
\begin{tabular}{lcccccccc}
\toprule[1pt]
 & \multicolumn{2}{c}{\textbf{Cora}} & \multicolumn{2}{c}{\textbf{CiteSeer}} & \multicolumn{2}{c}{\textbf{PubMed}} & \multicolumn{2}{c}{\textbf{ogbn-arxiv}} \\
\cline{2-9}
& Acc. & F1 & Acc. & F1 & Acc. & F1 & Acc. & F1 \\ \midrule
Directly optimize $A'$         & 70.31{\scriptsize$\pm$4.21} & 54.36{\scriptsize$\pm$3.71} & 66.27{\scriptsize$\pm$0.80} & 53.12{\scriptsize$\pm$0.18} & \textbf{82.71}{\scriptsize$\pm$0.16} & \textbf{66.53}{\scriptsize$\pm$0.23} & 70.66{\scriptsize$\pm$0.06} & 35.54{\scriptsize$\pm$0.44} \\
$A' = \langle\mathbf{x}_i, \mathbf{x}_j\rangle$ & 73.01{\scriptsize$\pm$1.76} & 58.78{\scriptsize$\pm$3.01} & 69.57{\scriptsize$\pm$0.40} & \textbf{58.37}{\scriptsize$\pm$0.08} & 82.68{\scriptsize$\pm$0.05} & 66.06{\scriptsize$\pm$0.92} & 70.68{\scriptsize$\pm$0.01} & 35.30{\scriptsize$\pm$0.54} \\
\textbf{Ours}                          & \textbf{76.43}{\scriptsize$\pm$1.24} & \textbf{61.58}{\scriptsize$\pm$2.16} & \textbf{71.61}{\scriptsize$\pm$0.40} & 58.24{\scriptsize$\pm$0.96} & \textbf{82.71}{\scriptsize$\pm$0.03} & 66.40{\scriptsize$\pm$0.34} & \textbf{70.93}{\scriptsize$\pm$0.59} & \textbf{36.66}{\scriptsize$\pm$0.22} \\
\bottomrule[1pt]
\end{tabular}
\caption{Comparison of $g_{\theta}$ implementations}
\label{tab:diff_adj_design}
\end{table*}

\subsection{Performance under Model Heterogeneity}
\label{sec:appendix_model_hete}
We conduct experiments on 10 clients with heterogeneous models. The model configuration is detailed in Section~\ref{sec:app_hyper-parameter}. Under model heterogeneity, most pFGL methods (except FedTAD and FedSpray) do not work since they rely on the model parameter aggregation. Thus, we compare our method with FedTAD, FedSpray, DENSE, and Co-Boost. The results under Louvain and Metis partitions are presented in Table~\ref{tab:hete_louvain} and Table~\ref{tab:hete_metis}. The results demonstrate the superior performance of our method in both accuracy and F1-macro.

\begin{table}[tbp]
\setlength{\tabcolsep}{1mm}
\small
\centering
\resizebox{\columnwidth}{!}{%
\begin{tabular}{c|cc|cc|cc}
\toprule
\multirow{2}{*}{Method} & \multicolumn{2}{c|}{\textbf{Cora}}           & \multicolumn{2}{c|}{\textbf{CiteSeer}}       & \multicolumn{2}{c}{\textbf{ogbn-arxiv}} \\ \cline{2-7}  \rule{0pt}{2ex}
  & Acc.           & F1            & Acc.           & F1            & Acc.           & F1             \\ \midrule
  
Standalone & 68.37{\scriptsize$\pm$0.19} & 43.51{\scriptsize$\pm$0.31} & 62.50{\scriptsize$\pm$0.15} & 43.94{\scriptsize$\pm$0.09} & 66.13{\scriptsize$\pm$0.08} & 19.60{\scriptsize$\pm$0.38} \\ \midrule

FedTAD & 68.18{\scriptsize$\pm$0.22} & 42.91{\scriptsize$\pm$0.34} & 61.81{\scriptsize$\pm$0.18} & 43.78{\scriptsize$\pm$0.14} & 69.08{\scriptsize$\pm$0.08} & 29.29{\scriptsize$\pm$0.44} \\

FedSpray & 65.73{\scriptsize$\pm$1.05} & 42.23{\scriptsize$\pm$0.34} & 53.14{\scriptsize$\pm$1.19} & 39.13{\scriptsize$\pm$1.17} & 68.63{\scriptsize$\pm$0.03} & 22.25{\scriptsize$\pm$0.26} \\ \midrule

DENSE & 68.15{\scriptsize$\pm$0.47} & 43.02{\scriptsize$\pm$0.36} & 62.31{\scriptsize$\pm$0.51} & 44.27{\scriptsize$\pm$0.44} & 69.03{\scriptsize$\pm$0.02} & 29.64{\scriptsize$\pm$0.14} \\

Co-Boost & 68.38{\scriptsize$\pm$0.40} & 43.13{\scriptsize$\pm$0.52} & 62.22{\scriptsize$\pm$0.18} & 44.23{\scriptsize$\pm$0.05} & 69.08{\scriptsize$\pm$0.13} & 29.88{\scriptsize$\pm$0.27} \\ \midrule

O-pFGL& \textbf{75.04}{\scriptsize$\pm$0.50} & \textbf{57.26}{\scriptsize$\pm$1.93} & \textbf{70.87}{\scriptsize$\pm$0.21} & \textbf{56.67}{\scriptsize$\pm$0.20} & \textbf{70.98}{\scriptsize$\pm$0.09} & \textbf{36.41}{\scriptsize$\pm$0.43} \\ 

\bottomrule
\end{tabular}
}
\caption{Performance of methods on 10 clients with heterogeneous models under Louvain partition.}
\label{tab:hete_louvain}
\end{table}

\begin{table}[tbp]
\centering
\resizebox{\columnwidth}{!}{%
\begin{tabular}{c|cc|cc|cc}
\toprule[1pt]
\multirow{2}{*}{Method} & \multicolumn{2}{c|}{\textbf{Cora}}           & \multicolumn{2}{c|}{\textbf{CiteSeer}}       & \multicolumn{2}{c}{\textbf{ogbn-arxiv}} \\ \cline{2-7}  \rule{0pt}{2ex}
  & Acc.           & F1            & Acc.           & F1            & Acc.           & F1            \\ \midrule
  
Standalone & 74.33{\scriptsize$\pm$0.14} & 30.73{\scriptsize$\pm$0.18} & 64.53{\scriptsize$\pm$0.34} & 37.52{\scriptsize$\pm$0.33} & 65.18{\scriptsize$\pm$0.11} & 25.25{\scriptsize$\pm$0.14} \\ \midrule

FedTAD & 74.37{\scriptsize$\pm$0.30} & 31.27{\scriptsize$\pm$0.25} & 64.75{\scriptsize$\pm$0.64} & 37.29{\scriptsize$\pm$0.34} & 68.57{\scriptsize$\pm$0.08} & 34.32{\scriptsize$\pm$0.44} \\

FedSpray & 73.09{\scriptsize$\pm$0.69} & 30.84{\scriptsize$\pm$0.65} & 60.38{\scriptsize$\pm$0.67} & 34.96{\scriptsize$\pm$0.16} & 67.39{\scriptsize$\pm$0.11} & 27.88{\scriptsize$\pm$0.24} \\ \midrule

DENSE & 73.51{\scriptsize$\pm$0.99} & 30.62{\scriptsize$\pm$0.42} & 65.33{\scriptsize$\pm$0.24} & 37.62{\scriptsize$\pm$0.09} & 68.56{\scriptsize$\pm$0.03} & 34.64{\scriptsize$\pm$0.03} \\

Co-Boost & 73.19{\scriptsize$\pm$0.72} & 30.56{\scriptsize$\pm$0.33} & 65.40{\scriptsize$\pm$0.09} & 37.66{\scriptsize$\pm$0.19} & 68.55{\scriptsize$\pm$0.09} & 34.66{\scriptsize$\pm$0.21} \\ \midrule

O-pFGL & \textbf{79.64}{\scriptsize$\pm$0.94} & \textbf{44.86}{\scriptsize$\pm$2.04} & \textbf{71.82}{\scriptsize$\pm$0.34} & \textbf{47.57}{\scriptsize$\pm$0.98} & \textbf{70.31}{\scriptsize$\pm$0.06} & \textbf{40.31}{\scriptsize$\pm$0.11} \\ 

\bottomrule[1pt] 
\end{tabular}%
}
\caption{Performance of methods on 10 clients with heterogeneous models under Metis partition.}
\label{tab:hete_metis}
\end{table}

\begin{table}[tbp]
\centering
\resizebox{\columnwidth}{!}{%
\begin{tabular}{c|cc|cc|cc}
\toprule[1pt]
\multirow{2}{*}{Method} & \multicolumn{2}{c|}{\textbf{Flickr}} & \multicolumn{2}{c|}{\textbf{Reddit}} & \multicolumn{2}{c}{\textbf{Reddit2}} \\ \cline{2-7}
 & Acc.           & F1            & Acc.           & F1            & Acc.           & F1            \\  \midrule
Standalone & 44.97{\scriptsize$\pm$0.70} & 12.84{\scriptsize$\pm$0.06} & 90.35{\scriptsize$\pm$0.02} & 28.79{\scriptsize$\pm$0.07} & 91.92{\scriptsize$\pm$0.01} & 32.53{\scriptsize$\pm$0.20} \\ 
FedAvg & 45.50{\scriptsize$\pm$0.97} & 13.66{\scriptsize$\pm$0.20} & 80.75{\scriptsize$\pm$0.81} & 16.85{\scriptsize$\pm$0.49} & 82.86{\scriptsize$\pm$0.62} & 19.20{\scriptsize$\pm$0.44} \\  \midrule
FedPUB & 42.17{\scriptsize$\pm$0.00} & 8.32{\scriptsize$\pm$0.00} & \textbf{90.74}{\scriptsize$\pm$0.02} & 27.86{\scriptsize$\pm$0.01} & 92.03{\scriptsize$\pm$0.02} & 30.37{\scriptsize$\pm$0.09} \\ 
FedGTA & 29.99{\scriptsize$\pm$1.71} & 14.38{\scriptsize$\pm$1.15} & 90.35{\scriptsize$\pm$0.02} & 28.76{\scriptsize$\pm$0.24} & 91.91{\scriptsize$\pm$0.38} & 32.71{\scriptsize$\pm$0.11} \\ 
FedTAD & 45.44{\scriptsize$\pm$0.47} & 13.85{\scriptsize$\pm$0.50} & \multicolumn{2}{c|}{OOM} & \multicolumn{2}{c}{OOM} \\ 
FedSpray & 43.08{\scriptsize$\pm$0.32} & 13.08{\scriptsize$\pm$1.75} & 89.07{\scriptsize$\pm$0.21} & 27.28{\scriptsize$\pm$0.24} & 91.28{\scriptsize$\pm$0.11} & 31.90{\scriptsize$\pm$0.17} \\  
FedLoG & \multicolumn{2}{c|}{OOM} & \multicolumn{2}{c|}{OOM} & \multicolumn{2}{c}{OOM} \\ 
FedGM & 42.36{\scriptsize$\pm$1.47} & 10.29{\scriptsize$\pm$0.38} & \multicolumn{2}{c|}{OOM} & \multicolumn{2}{c}{OOM} \\ 
GHOST & 42.41{\scriptsize$\pm$0.56} & \textbf{15.38}{\scriptsize$\pm$0.30} & \multicolumn{2}{c|}{OOM} & \multicolumn{2}{c}{OOM} \\ 
\midrule
DENSE & 44.60{\scriptsize$\pm$0.28} & 12.47{\scriptsize$\pm$0.39} & 90.32{\scriptsize$\pm$0.04} & 30.08{\scriptsize$\pm$0.10} & 92.00{\scriptsize$\pm$0.05} & 34.42{\scriptsize$\pm$0.29} \\ 
Co-Boost & 45.79{\scriptsize$\pm$0.54} & 13.52{\scriptsize$\pm$0.52} & 90.35{\scriptsize$\pm$0.03} & 28.62{\scriptsize$\pm$0.05} & 91.94{\scriptsize$\pm$0.02} & 32.68{\scriptsize$\pm$0.04} \\  \midrule
O-pFGL (Ours) & \textbf{46.83}{\scriptsize$\pm$0.44} & 14.31{\scriptsize$\pm$0.04} & \textbf{90.74}{\scriptsize$\pm$0.11} & \textbf{30.09}{\scriptsize$\pm$0.49} & \textbf{92.14}{\scriptsize$\pm$0.05} & \textbf{35.51}{\scriptsize$\pm$0.15} \\ 
\bottomrule[1pt]
\end{tabular}%
}
\caption{Performance on inductive datasets under Louvain partition with 10 clients.}
\label{tab:inductive_louvain_10_clients}
\end{table}

\begin{table}[tbp]
\centering
\resizebox{\columnwidth}{!}{%
\begin{tabular}{c|cc|cc|cc}
\toprule[1pt]
\multirow{2}{*}{Method} & \multicolumn{2}{c|}{\textbf{Flickr}} & \multicolumn{2}{c|}{\textbf{Reddit}} & \multicolumn{2}{c}{\textbf{Reddit2}} \\ \cline{2-7} \rule{0pt}{2ex}
  & Acc.           & F1            & Acc.           & F1            & Acc.           & F1           \\ \midrule
Standalone & 42.10{\scriptsize$\pm$0.14} & 15.39{\scriptsize$\pm$0.25} & 89.90{\scriptsize$\pm$0.06} & 13.08{\scriptsize$\pm$0.21} & 91.09{\scriptsize$\pm$0.02} & 16.52{\scriptsize$\pm$0.04} \\
FedAvg & 28.86{\scriptsize$\pm$1.96} & 13.55{\scriptsize$\pm$0.36} & 83.99{\scriptsize$\pm$0.28} & 7.81{\scriptsize$\pm$0.08} & 85.94{\scriptsize$\pm$0.30} & 8.48{\scriptsize$\pm$0.06} \\ \midrule
FedPUB & 42.06{\scriptsize$\pm$0.00} & 8.31{\scriptsize$\pm$0.00} & 90.34{\scriptsize$\pm$0.08} & 11.95{\scriptsize$\pm$0.27} & 91.23{\scriptsize$\pm$0.02} & 13.96{\scriptsize$\pm$0.11} \\
FedGTA & 40.78{\scriptsize$\pm$4.26} & 14.11{\scriptsize$\pm$1.51} & 89.83{\scriptsize$\pm$0.05} & 12.90{\scriptsize$\pm$0.32} & 91.14{\scriptsize$\pm$0.05} & 16.64{\scriptsize$\pm$0.12} \\
FedTAD & 41.61{\scriptsize$\pm$0.42} & 15.88{\scriptsize$\pm$0.31} & \multicolumn{2}{c|}{OOM} & \multicolumn{2}{c}{OOM} \\
FedSpray & 42.32{\scriptsize$\pm$1.20} & 14.03{\scriptsize$\pm$0.08} & 89.41{\scriptsize$\pm$0.06} & 10.71{\scriptsize$\pm$0.14} & 90.51{\scriptsize$\pm$0.05} & 15.29{\scriptsize$\pm$0.06} \\ 
FedLoG & \multicolumn{2}{c|}{OOM} & \multicolumn{2}{c|}{OOM} & \multicolumn{2}{c}{OOM} \\ 
FedGM & 42.53{\scriptsize$\pm$2.35} & 10.28{\scriptsize$\pm$0.25} & \multicolumn{2}{c|}{OOM} & \multicolumn{2}{c}{OOM} \\ 
GHOST & 39.27{\scriptsize$\pm$0.35} & \textbf{16.93}{\scriptsize$\pm$0.12} & \multicolumn{2}{c|}{OOM} & \multicolumn{2}{c}{OOM} \\ 
\midrule
DENSE & 42.45{\scriptsize$\pm$0.72} & 11.67{\scriptsize$\pm$1.07} & 89.89{\scriptsize$\pm$0.05} & 14.27{\scriptsize$\pm$0.14} & 91.25{\scriptsize$\pm$0.03} & 18.14{\scriptsize$\pm$0.44} \\
Co-Boost & 42.07{\scriptsize$\pm$0.07} & 15.28{\scriptsize$\pm$0.45} & 89.93{\scriptsize$\pm$0.03} & 13.08{\scriptsize$\pm$0.19} & 91.12{\scriptsize$\pm$0.03} & 16.62{\scriptsize$\pm$0.15} \\ \midrule
O-pFGL & \textbf{44.00}{\scriptsize$\pm$0.29} & 15.60{\scriptsize$\pm$0.47} & \textbf{90.38}{\scriptsize$\pm$0.05} & \textbf{15.70}{\scriptsize$\pm$0.47} & \textbf{91.34}{\scriptsize$\pm$0.03} & \textbf{18.34}{\scriptsize$\pm$0.18} \\
\bottomrule[1pt]
\end{tabular}%
}
\caption{Performance on inductive datasets under Louvain partition with 20 clients.}
\label{tab:inductive_louvain_20_clients}
\end{table}

\begin{table}[tbp]
\centering
\resizebox{\columnwidth}{!}{%
\begin{tabular}{c|cc|cc|cc}
\toprule[1pt]
\multirow{2}{*}{Method} & \multicolumn{2}{c|}{\textbf{Flickr}} & \multicolumn{2}{c|}{\textbf{Reddit}} & \multicolumn{2}{c}{\textbf{Reddit2}} \\ \cline{2-7} \rule{0pt}{2ex}
  & Acc.           & F1            & Acc.           & F1            & Acc.           & F1           \\ \midrule
Standalone & 47.47{\scriptsize$\pm$0.12} & 12.89{\scriptsize$\pm$0.47} & 88.80{\scriptsize$\pm$0.09} & 42.34{\scriptsize$\pm$0.40} & 90.86{\scriptsize$\pm$0.04} & 44.78{\scriptsize$\pm$0.35} \\
FedAvg & 34.80{\scriptsize$\pm$3.92} & 13.45{\scriptsize$\pm$0.59} & 78.83{\scriptsize$\pm$0.07} & 24.21{\scriptsize$\pm$0.30} & 79.30{\scriptsize$\pm$0.18} & 26.26{\scriptsize$\pm$0.45} \\ \midrule
FedPUB & 42.65{\scriptsize$\pm$0.40} & 10.09{\scriptsize$\pm$2.14} & 88.97{\scriptsize$\pm$0.18} & 41.01{\scriptsize$\pm$0.18} & 90.72{\scriptsize$\pm$0.07} & 41.66{\scriptsize$\pm$0.30} \\
FedGTA & 35.88{\scriptsize$\pm$6.10} & 13.04{\scriptsize$\pm$3.66} & 88.82{\scriptsize$\pm$0.11} & 42.02{\scriptsize$\pm$0.21} & 90.80{\scriptsize$\pm$0.07} & 44.59{\scriptsize$\pm$0.27} \\
FedTAD & 47.55{\scriptsize$\pm$0.08} & 13.07{\scriptsize$\pm$0.23} & \multicolumn{2}{c|}{OOM} & \multicolumn{2}{c}{OOM} \\
FedSpray & 43.77{\scriptsize$\pm$1.58} & 13.37{\scriptsize$\pm$0.42} & 87.71{\scriptsize$\pm$0.18} & 41.65{\scriptsize$\pm$0.29} & 89.99{\scriptsize$\pm$0.23} & 43.74{\scriptsize$\pm$0.26} \\ 
FedLoG & \multicolumn{2}{c|}{OOM} & \multicolumn{2}{c|}{OOM} & \multicolumn{2}{c}{OOM} \\ 
FedGM & 43.72{\scriptsize$\pm$2.83} & 9.66{\scriptsize$\pm$0.15} & \multicolumn{2}{c|}{OOM} & \multicolumn{2}{c}{OOM} \\ 
GHOST & 41.55{\scriptsize$\pm$0.54} & \textbf{15.63}{\scriptsize$\pm$0.22} & \multicolumn{2}{c|}{OOM} & \multicolumn{2}{c}{OOM} \\ 
\midrule
DENSE & 45.26{\scriptsize$\pm$0.84} & 11.56{\scriptsize$\pm$0.84} & 88.72{\scriptsize$\pm$0.08} & 44.31{\scriptsize$\pm$0.23} & 90.01{\scriptsize$\pm$0.10} & 46.68{\scriptsize$\pm$0.03} \\
Co-Boost & 47.42{\scriptsize$\pm$0.10} & 12.90{\scriptsize$\pm$0.61} & 88.71{\scriptsize$\pm$0.07} & 42.08{\scriptsize$\pm$0.27} & 90.81{\scriptsize$\pm$0.06} & 44.56{\scriptsize$\pm$0.20} \\ \midrule
O-pFGL & \textbf{47.89}{\scriptsize$\pm$0.11} & 13.11{\scriptsize$\pm$0.34} & \textbf{89.59}{\scriptsize$\pm$0.10} & \textbf{45.31}{\scriptsize$\pm$0.36} & \textbf{91.16}{\scriptsize$\pm$0.04} & \textbf{48.56}{\scriptsize$\pm$0.31} \\
\bottomrule[1pt]
\end{tabular}%
}
\caption{Performance on inductive datasets under Metis partition with 10 clients.}
\label{tab:inductive_metis_10_clients}
\end{table}

\subsection{Inductive Performance}
\label{sec:appendix_inductive}
We conduct experiments on three inductive graph datasets: Flickr, Reddit, and Reddit2, and vary the number of clients. Since Flickr has lower homophily, we use the 2-layer SIGN~\cite{frasca2020sign} for better graph learning. With 10 clients, experimental results under Louvain and Metis partitions are shown in Table~\ref{tab:inductive_louvain_10_clients} and Table~\ref{tab:inductive_metis_10_clients}, respectively. With 20 clients, experimental results under Louvain and Metis partitions are shown in Table~\ref{tab:inductive_louvain_20_clients} and Table~\ref{tab:inductive_metis_20_clients}, respectively. Our method is scalable and generally performs best.

\begin{table}[tbp]
\centering
\resizebox{\columnwidth}{!}{%
\begin{tabular}{c|cc|cc|cc}
\toprule[1pt]
\multirow{2}{*}{Method} & \multicolumn{2}{c|}{\textbf{Flickr}} & \multicolumn{2}{c|}{\textbf{Reddit}} & \multicolumn{2}{c}{\textbf{Reddit2}} \\ \cline{2-7} \rule{0pt}{2ex}
  & Acc.           & F1            & Acc.           & F1            & Acc.           & F1           \\ \midrule
Standalone & 47.10{\scriptsize$\pm$0.20} & 13.22{\scriptsize$\pm$0.37} & 87.52{\scriptsize$\pm$0.03} & 28.22{\scriptsize$\pm$0.10} & 89.59{\scriptsize$\pm$0.01} & 25.35{\scriptsize$\pm$0.08} \\
FedAvg & 33.44{\scriptsize$\pm$2.08} & 12.89{\scriptsize$\pm$0.45} & 80.71{\scriptsize$\pm$0.28} & 14.64{\scriptsize$\pm$0.48} & 82.90{\scriptsize$\pm$0.39} & 13.91{\scriptsize$\pm$0.41} \\ \midrule
FedPUB & 42.10{\scriptsize$\pm$0.00} & 8.30{\scriptsize$\pm$0.00} & 87.36{\scriptsize$\pm$0.56} & 26.45{\scriptsize$\pm$0.71} & 89.89{\scriptsize$\pm$0.06} & 22.63{\scriptsize$\pm$0.31} \\
FedGTA & 38.75{\scriptsize$\pm$9.16} & 11.38{\scriptsize$\pm$3.46} & 87.49{\scriptsize$\pm$0.01} & 27.95{\scriptsize$\pm$0.22} & 89.55{\scriptsize$\pm$0.03} & 25.47{\scriptsize$\pm$0.29} \\
FedTAD & \textbf{47.17}{\scriptsize$\pm$0.13} & 13.03{\scriptsize$\pm$0.38} & \multicolumn{2}{c|}{OOM} & \multicolumn{2}{c}{OOM} \\
FedSpray & 43.35{\scriptsize$\pm$0.96} & 14.00{\scriptsize$\pm$0.55} & 86.66{\scriptsize$\pm$0.09} & 27.60{\scriptsize$\pm$0.20} & 88.99{\scriptsize$\pm$0.01} & 24.23{\scriptsize$\pm$0.39} \\ 
FedLoG & \multicolumn{2}{c|}{OOM} & \multicolumn{2}{c|}{OOM} & \multicolumn{2}{c}{OOM} \\ 
FedGM & 45.28{\scriptsize$\pm$0.65} & 10.69{\scriptsize$\pm$0.40} & \multicolumn{2}{c|}{OOM} & \multicolumn{2}{c}{OOM} \\ 
GHOST & 40.09{\scriptsize$\pm$1.72} & \textbf{14.69}{\scriptsize$\pm$0.23} & \multicolumn{2}{c|}{OOM} & \multicolumn{2}{c}{OOM} \\ 
\midrule
DENSE & 44.13{\scriptsize$\pm$0.97} & 11.45{\scriptsize$\pm$0.39} & 87.60{\scriptsize$\pm$0.02} & 30.49{\scriptsize$\pm$0.17} & 89.97{\scriptsize$\pm$0.05} & 27.72{\scriptsize$\pm$0.16} \\
Co-Boost & 47.08{\scriptsize$\pm$0.22} & 13.11{\scriptsize$\pm$0.66} & 87.52{\scriptsize$\pm$0.06} & 28.09{\scriptsize$\pm$0.12} & 89.67{\scriptsize$\pm$0.07} & 25.62{\scriptsize$\pm$0.22} \\ \midrule
O-pFGL & 47.08{\scriptsize$\pm$0.27} & 13.37{\scriptsize$\pm$0.24} & \textbf{88.28}{\scriptsize$\pm$0.01} & \textbf{31.01}{\scriptsize$\pm$0.47} & \textbf{90.25}{\scriptsize$\pm$0.07} & \textbf{30.74}{\scriptsize$\pm$0.25} \\
\bottomrule[1pt]
\end{tabular}%
}
\caption{Performance on inductive datasets under Metis partition with 20 clients.}
\label{tab:inductive_metis_20_clients}
\end{table}

\subsection{More Datasets}
\label{sec:more_datasets}
We further provide more experimental results on Computers, Photo, CS, and Physics datasets. Computers and Photo are Amazon co-purchase networks, and CS and Physics are Co-author networks. The results under Louvain and Metis partitions are shown in Figure~\ref{tab:more_dataset_louvain} and Figure~\ref{tab:more_dataset_metis}, respectively. Our method consistently outperforms others in terms of both accuracy and F1-macro, which demonstrates the general advantages of our method.

\begin{table*}[tbp]
\setlength{\tabcolsep}{1mm}
\small
\centering
\begin{tabular}{c|cc|cc|cc|cc}
\toprule[1pt]
\multirow{2}{*}{Method} & \multicolumn{2}{c|}{\textbf{Computers}}    & \multicolumn{2}{c|}{\textbf{Photo}}       & \multicolumn{2}{c|}{\textbf{CS}}         & \multicolumn{2}{c}{\textbf{Physics}}    \\ \cline{2-9}
 & Acc.           & F1            & Acc.           & F1            & Acc.           & F1            & Acc.           & F1            \\  \midrule
Standalone & 87.34{\scriptsize$\pm$0.06} & 38.27{\scriptsize$\pm$0.91} & 89.16{\scriptsize$\pm$0.20} & 40.46{\scriptsize$\pm$1.31} & 87.18{\scriptsize$\pm$0.02} & 42.97{\scriptsize$\pm$0.07} & 93.65{\scriptsize$\pm$0.06} & 68.40{\scriptsize$\pm$0.53} \\ 
FedAvg & 87.18{\scriptsize$\pm$0.75} & 38.94{\scriptsize$\pm$0.51} & 89.26{\scriptsize$\pm$0.13} & 41.11{\scriptsize$\pm$0.77} & 87.10{\scriptsize$\pm$0.11} & 44.86{\scriptsize$\pm$0.23} & 93.55{\scriptsize$\pm$0.03} & 70.72{\scriptsize$\pm$0.38} \\  \midrule
FedPUB & 83.51{\scriptsize$\pm$2.46} & 31.48{\scriptsize$\pm$2.64} & 89.43{\scriptsize$\pm$0.05} & 38.17{\scriptsize$\pm$1.24} & 87.01{\scriptsize$\pm$0.12} & 41.99{\scriptsize$\pm$0.38} & 93.55{\scriptsize$\pm$1.22} & 64.95{\scriptsize$\pm$1.91} \\ 
FedGTA & 56.73{\scriptsize$\pm$14.13} & 13.41{\scriptsize$\pm$3.08} & 53.68{\scriptsize$\pm$3.75} & 15.09{\scriptsize$\pm$1.37} & 83.26{\scriptsize$\pm$0.11} & 40.34{\scriptsize$\pm$1.10} & 91.81{\scriptsize$\pm$0.22} & 64.34{\scriptsize$\pm$0.77} \\ 
FedTAD & 87.08{\scriptsize$\pm$0.59} & 38.71{\scriptsize$\pm$1.00} & 89.61{\scriptsize$\pm$0.27} & 41.91{\scriptsize$\pm$1.14} & 87.15{\scriptsize$\pm$0.06} & 44.80{\scriptsize$\pm$0.05} & 93.51{\scriptsize$\pm$0.04} & 70.52{\scriptsize$\pm$0.53} \\ 
FedSpray & 86.67{\scriptsize$\pm$2.31} & 28.56{\scriptsize$\pm$4.74} & 87.40{\scriptsize$\pm$0.50} & 35.59{\scriptsize$\pm$0.92} & 86.58{\scriptsize$\pm$0.35} & 38.69{\scriptsize$\pm$1.36} & 92.01{\scriptsize$\pm$0.90} & 48.68{\scriptsize$\pm$1.13} \\
FedLoG & 72.74{\scriptsize$\pm$3.50} & 13.19{\scriptsize$\pm$3.33} & 79.70{\scriptsize$\pm$1.36} & 21.10{\scriptsize$\pm$0.85} & 82.25{\scriptsize$\pm$0.46} & 24.48{\scriptsize$\pm$1.02} & \multicolumn{2}{c}{OOM} \\
FedGM & 87.52{\scriptsize$\pm$0.17} & 39.78{\scriptsize$\pm$0.17} & 89.12{\scriptsize$\pm$0.05} & 40.96{\scriptsize$\pm$0.21} & 88.35{\scriptsize$\pm$0.02} & 45.01{\scriptsize$\pm$0.42} & 94.08{\scriptsize$\pm$0.02} & 69.26{\scriptsize$\pm$0.18} \\
GHOST & 87.44{\scriptsize$\pm$0.12} & 40.15{\scriptsize$\pm$0.20} & 89.04{\scriptsize$\pm$0.28} & 40.89{\scriptsize$\pm$0.64} & 88.28{\scriptsize$\pm$0.07} & 44.87{\scriptsize$\pm$0.26} & 94.01{\scriptsize$\pm$0.01} & 69.09{\scriptsize$\pm$0.22} \\
\midrule
DENSE & 87.19{\scriptsize$\pm$0.44} & 39.18{\scriptsize$\pm$0.36} & 89.36{\scriptsize$\pm$0.22} & 41.87{\scriptsize$\pm$0.55} & 87.15{\scriptsize$\pm$0.07} & 44.92{\scriptsize$\pm$0.25} & 93.52{\scriptsize$\pm$0.04} & 70.02{\scriptsize$\pm$0.42} \\ 
Co-Boost & 87.56{\scriptsize$\pm$0.53} & 38.98{\scriptsize$\pm$0.82} & 89.52{\scriptsize$\pm$0.24} & 41.54{\scriptsize$\pm$0.65} & 87.18{\scriptsize$\pm$0.12} & 44.87{\scriptsize$\pm$0.30} & 93.54{\scriptsize$\pm$0.03} & 70.45{\scriptsize$\pm$0.36} \\  \midrule
O-pFGL & \textbf{88.30}{\scriptsize$\pm$0.06} & \textbf{44.01}{\scriptsize$\pm$1.93} & \textbf{90.24}{\scriptsize$\pm$0.09} & \textbf{48.25}{\scriptsize$\pm$1.79} & \textbf{90.00}{\scriptsize$\pm$0.12} & \textbf{53.88}{\scriptsize$\pm$1.15} & \textbf{94.21}{\scriptsize$\pm$0.22} & \textbf{71.17}{\scriptsize$\pm$1.35} \\ 
\bottomrule[1pt]
\end{tabular}%
\caption{Performance of methods on more datasets under the Louvain partition.}
\label{tab:more_dataset_louvain}
\end{table*}

\begin{table*}[tbp]
\setlength{\tabcolsep}{1mm}
\small
\centering
\begin{tabular}{c|cc|cc|cc|cc}
\toprule[1pt]
\multirow{2}{*}{Method} & \multicolumn{2}{c|}{\textbf{Computers}}    & \multicolumn{2}{c|}{\textbf{Photo}}       & \multicolumn{2}{c|}{\textbf{CS}}         & \multicolumn{2}{c}{\textbf{Physics}}    \\ \cline{2-9}  \rule{0pt}{2ex}
  & Acc.           & F1            & Acc.           & F1            & Acc.           & F1            & Acc.           & F1             \\ \midrule
  
Standalone & 87.50{\scriptsize$\pm$0.34} & 25.43{\scriptsize$\pm$0.69} & 90.62{\scriptsize$\pm$1.41} & 29.70{\scriptsize$\pm$0.16} & 87.39{\scriptsize$\pm$0.13} & 42.30{\scriptsize$\pm$0.35} & 93.89{\scriptsize$\pm$0.04} & 65.11{\scriptsize$\pm$0.20} \\

FedAvg & 87.36{\scriptsize$\pm$0.30} & 24.85{\scriptsize$\pm$1.69} & 90.76{\scriptsize$\pm$0.45} & 30.33{\scriptsize$\pm$1.13} & 87.14{\scriptsize$\pm$0.03} & 42.10{\scriptsize$\pm$0.05} & 93.65{\scriptsize$\pm$0.08} & 64.16{\scriptsize$\pm$0.50} \\ \midrule

FedPUB & 85.20{\scriptsize$\pm$1.11} & 19.76{\scriptsize$\pm$1.52} & 89.66{\scriptsize$\pm$0.30} & 23.64{\scriptsize$\pm$1.85} & 87.30{\scriptsize$\pm$0.14} & 42.00{\scriptsize$\pm$0.47} & 93.99{\scriptsize$\pm$0.06} & 62.92{\scriptsize$\pm$0.96} \\

FedGTA & 74.36{\scriptsize$\pm$4.49} & 15.48{\scriptsize$\pm$1.66} & 53.68{\scriptsize$\pm$3.76} & 15.09{\scriptsize$\pm$1.37} & 87.49{\scriptsize$\pm$0.03} & 42.59{\scriptsize$\pm$0.11} & 84.00{\scriptsize$\pm$3.98} & 23.27{\scriptsize$\pm$0.77} \\

FedTAD & 87.64{\scriptsize$\pm$0.32} & 24.43{\scriptsize$\pm$0.31} & 90.66{\scriptsize$\pm$0.36} & 29.64{\scriptsize$\pm$1.44} & 87.15{\scriptsize$\pm$0.02} & 42.15{\scriptsize$\pm$0.21} & 93.58{\scriptsize$\pm$0.03} & 63.77{\scriptsize$\pm$0.17} \\

FedSpray & 86.34{\scriptsize$\pm$0.49} & 21.72{\scriptsize$\pm$1.82} & 88.77{\scriptsize$\pm$0.49} & 21.34{\scriptsize$\pm$0.98} & 81.50{\scriptsize$\pm$0.29} & 38.04{\scriptsize$\pm$0.26} & 91.17{\scriptsize$\pm$0.12} & 60.79{\scriptsize$\pm$0.78} \\ 

FedLoG & 81.68{\scriptsize$\pm$0.98} & 12.03{\scriptsize$\pm$1.53} & 85.85{\scriptsize$\pm$0.23} & 13.00{\scriptsize$\pm$0.22} & 83.79{\scriptsize$\pm$0.73} & 20.05{\scriptsize$\pm$1.23} & \multicolumn{2}{c}{OOM} \\ 

FedGM & 87.60{\scriptsize$\pm$0.18} & 26.37{\scriptsize$\pm$0.50} & 90.48{\scriptsize$\pm$0.30} & 30.51{\scriptsize$\pm$0.26} & 88.50{\scriptsize$\pm$0.05} & 43.44{\scriptsize$\pm$0.26} & 94.45{\scriptsize$\pm$0.05} & 64.98{\scriptsize$\pm$0.33} \\

GHOST & 87.71{\scriptsize$\pm$0.21} & 26.36{\scriptsize$\pm$0.23} & 90.20{\scriptsize$\pm$1.10} & 30.68{\scriptsize$\pm$0.74} & 88.46{\scriptsize$\pm$0.04} & 43.25{\scriptsize$\pm$0.11} & 94.46{\scriptsize$\pm$0.03} & 65.28{\scriptsize$\pm$0.11} \\
\midrule

DENSE & 87.75{\scriptsize$\pm$0.29} & 23.99{\scriptsize$\pm$0.65} & 91.14{\scriptsize$\pm$0.21} & 31.40{\scriptsize$\pm$1.16} & 87.12{\scriptsize$\pm$0.08} & 42.04{\scriptsize$\pm$0.33} & 93.59{\scriptsize$\pm$0.07} & 64.03{\scriptsize$\pm$0.44} \\

Co-Boost & 87.32{\scriptsize$\pm$0.80} & 22.66{\scriptsize$\pm$1.74} & 90.97{\scriptsize$\pm$0.18} & 30.77{\scriptsize$\pm$0.73} & 87.12{\scriptsize$\pm$0.03} & 42.04{\scriptsize$\pm$0.08} & 93.57{\scriptsize$\pm$0.08} & 63.69{\scriptsize$\pm$0.68} \\ \midrule

O-pFGL & \textbf{88.59}{\scriptsize$\pm$0.13} & \textbf{33.01}{\scriptsize$\pm$3.66} & \textbf{91.79}{\scriptsize$\pm$0.13} & \textbf{42.39}{\scriptsize$\pm$0.81} & \textbf{89.96}{\scriptsize$\pm$0.20} & \textbf{50.33}{\scriptsize$\pm$1.01} & \textbf{94.51}{\scriptsize$\pm$0.09} & \textbf{66.41}{\scriptsize$\pm$1.24} \\ 
\bottomrule[1pt]
\end{tabular}%
\caption{Performance of methods on more datasets under the Metis partition.}
\label{tab:more_dataset_metis}
\end{table*}

\subsection{Different GNN Architectures}
\label{sec:appendix_other_gnn}
To demonstrate the generality of our proposed method under different graph models, we conduct experiments with 3 other prevalent graph models, including 1) GraphSage~\cite{hamilton2017inductive}, 2) GAT~\cite{velivckovic2017graph}, and 3) SGC~\cite{wu2019simplifying}. The experimental results under Louvain and Metis partitions are shown in Table~\ref{tab:diff_arch_louvain} and Table~\ref{tab:diff_arch_metis}, respectively. Our method shows consistent advantages over other methods.

\begin{table*}[tbp]
\centering
\resizebox{\textwidth}{!}{%
\begin{tabular}{c|cccccc|cccccc|cccccc}
\toprule[1pt]
\multirow{3}{*}{Method} & \multicolumn{6}{c|}{\textbf{GraphSage}} & \multicolumn{6}{c|}{\textbf{GAT}} & \multicolumn{6}{c}{\textbf{SGC}} \\ \cline{2-19} \rule{0pt}{2ex}
 & \multicolumn{2}{c}{Cora} & \multicolumn{2}{c}{CiteSeer} & \multicolumn{2}{c|}{ogbn-arxiv} & \multicolumn{2}{c}{Cora} & \multicolumn{2}{c}{CiteSeer} & \multicolumn{2}{c|}{ogbn-arxiv} & \multicolumn{2}{c}{Cora} & \multicolumn{2}{c}{CiteSeer} & \multicolumn{2}{c}{ogbn-arxiv} \\ \cline{2-19} \rule{0pt}{2ex}
 & Acc. & F1 & Acc. & F1 & Acc. & F1 & Acc. & F1 & Acc. & F1 & Acc. & F1 & Acc. & F1 & Acc. & F1 & Acc. & F1 \\ \midrule

Standalone & 66.35 & 38.60 & 59.50 & 39.87 & 67.30 & 20.33 & 68.79 & 43.37 & 60.42 & 40.49 & 67.76 & 20.41 & 65.53 & 39.71 & 60.28 & 41.15 & 59.17 & 9.19 \\

FedAvg & 66.63 & 40.82 & 57.59 & 38.86 & 50.05 & 3.10 & 69.82 & 45.56 & 62.83 & 45.27 & 51.51 & 3.55 & 66.44 & 39.02 & 63.00 & 42.32 & 59.49 & 9.45 \\  \midrule

FedPUB & 43.27 & 18.83 & 44.15 & 28.93 & 41.12 & 7.91 & 66.35 & 42.52 & 66.75 & 49.93 & 56.69 & 12.14 & 57.21 & 42.04 & 65.60 & 49.63 & 35.36 & 9.31 \\  

FedGTA & 47.25 & 26.43 & 65.55 & 49.46 & 58.59 & 14.48 & 47.89 & 30.40 & 65.96 & 51.17 & 59.70 & 15.48 & 48.40 & 32.90 & 64.26 & 46.59 & 53.25 & 8.32 \\ 

FedTAD & 66.82 & 40.87 & 57.14 & 38.31 & 68.78 & 29.83 & 69.63 & 45.58 & 60.22 & 43.63 & 50.04 & 3.55 & 69.59 & 43.68 & 64.36 & 46.04 & 65.38 & 17.76 \\  

FedSpray & 63.75 & 37.49 & 54.08 & 37.84 & 47.85 & 2.26 & 62.91 & 39.32 & 61.05 & 44.21 & 47.52 & 2.16 & 63.40 & 35.54 & 60.63 & 41.19 & 65.38 & 17.76 \\ 

FedLoG & 62.21 & 28.73 & 58.48 & 35.10 & \multicolumn{2}{c|}{OOM} & 65.00 & 36.93 & 60.26 & 39.55 & \multicolumn{2}{c|}{OOM} & \multicolumn{6}{c}{N/A} \\

FedGM & 69.02 & 44.39 & 61.58 & 44.59 & 47.28 & 2.07 & 69.61 & 45.42 & 60.94 & 43.89 & \multicolumn{2}{c|}{OOM} & 63.14 & 32.52 & 60.01 & 38.96 & 46.39 & 1.80 \\ 

GHOST & \multicolumn{6}{c|}{N/A} & 69.80 & 44.72 & 62.13 & 43.02 & 67.75 & 19.55 & \multicolumn{6}{c}{N/A} \\ 
\midrule

DENSE & 66.78 & 40.46 & 57.96 & 39.13 & 50.29 & 3.18 & 70.40 & 47.25 & 58.99 & 42.31 & 69.33 & 28.78 & 69.74 & 43.87 & 64.04 & 45.67 & 62.72 & 14.88 \\  

Co-Boost & 65.34 & 39.98 & 54.80 & 35.54 & 68.74 & 29.98 & 70.37 & 47.34 & 59.28 & 42.40 & 69.32 & 28.93 & 68.81 & 44.39 & 64.02 & 45.68 & 62.72 & 14.89 \\  \midrule

O-pFGL & \textbf{72.80} & \textbf{54.92} & \textbf{69.05} & \textbf{54.86} & \textbf{69.07} & \textbf{32.81} & \textbf{75.03} & \textbf{57.77} & \textbf{71.23} & \textbf{57.36} & \textbf{70.41} & \textbf{34.86} & \textbf{73.65} & \textbf{54.70} & \textbf{69.47} & \textbf{54.04} & \textbf{70.31} & \textbf{35.66} \\ 

\bottomrule[1pt]
\end{tabular}%
}
\caption{Performance of methods with different GNN architectures under Louvain partition.}
\label{tab:diff_arch_louvain}
\end{table*}

\begin{table*}[tbp]
\centering
\resizebox{\textwidth}{!}{%
\begin{tabular}{c|cccccc|cccccc|cccccc}
\toprule[1pt]
\multirow{3}{*}{Method} & \multicolumn{6}{c|}{\textbf{GraphSage}} & \multicolumn{6}{c|}{\textbf{GAT}} & \multicolumn{6}{c}{\textbf{SGC}} \\ \cline{2-19} \rule{0pt}{2ex}
 & \multicolumn{2}{c}{Cora} & \multicolumn{2}{c}{CiteSeer} & \multicolumn{2}{c|}{ogbn-arxiv} & \multicolumn{2}{c}{Cora} & \multicolumn{2}{c}{CiteSeer} & \multicolumn{2}{c|}{ogbn-arxiv} & \multicolumn{2}{c}{Cora} & \multicolumn{2}{c}{CiteSeer} & \multicolumn{2}{c}{ogbn-arxiv} \\ \cline{2-19} \rule{0pt}{2ex}
 & Acc. & F1 & Acc. & F1 & Acc. & F1 & Acc. & F1 & Acc. & F1 & Acc. & F1 & Acc. & F1 & Acc. & F1 & Acc. & F1 \\ \midrule

Standalone & 74.78 & 24.62 & 64.22 & 35.94 & 67.42 & 28.75 & 75.96 & 29.95 & 64.42 & 37.70 & 67.26 & 27.12 & 76.11 & 30.38 & 63.97 & 36.34 & 58.49 & 12.63 \\  

FedAvg & 74.89 & 26.16 & 65.14 & 36.46 & 50.04 & 4.10 & 75.01 & 29.16 & 65.66 & 37.58 & 49.51 & 3.83 & 75.48 & 23.69 & 66.78 & 36.34 & 58.99 & 13.16 \\ \midrule

FedPUB & 49.67 & 16.75 & 50.44 & 30.02 & 43.30 & 9.55 & 72.97 & 28.86 & 65.67 & 38.10 & 58.92 & 16.23 & 51.50 & 27.93 & 60.36 & 40.17 & 35.62 & 8.88 \\  

FedGTA & 53.42 & 13.48 & 64.86 & 31.57 & 55.65 & 12.00 & 53.67 & 13.76 & 65.90 & 34.49 & 54.81 & 12.88 & 54.67 & 16.28 & 64.67 & 29.34 & 54.11 & 10.27 \\ 

FedTAD & 74.98 & 26.80 & 63.63 & 35.32 & 68.56 & 35.21 & 75.17 & 30.61 & 63.06 & 36.35 & 50.04 & 3.55 & 76.47 & 31.17 & 65.40 & 37.54 & 63.74 & 22.61 \\ 

FedSpray & 72.79 & 23.95 & 59.21 & 33.27 & 47.45 & 2.69 & 74.84 & 30.56 & 63.97 & 36.79 & 45.46 & 2.18 & 75.86 & 25.96 & 65.74 & 36.17 & 65.40 & 24.64 \\ 

FedLoG & 72.99 & 15.69 & 63.87 & 28.28 & \multicolumn{2}{c|}{OOM} & 75.34 & 25.11 & 63.94 & 32.37 & \multicolumn{2}{c|}{OOM} &  \multicolumn{6}{c}{N/A} \\

FedGM & 76.00 & 29.41 & 66.67 & 39.22 & 47.70 & 2.69 & 75.92 & 31.58 & 65.50 & 37.76 & \multicolumn{2}{c|}{OOM} & 74.28 & 21.62 & 66.01 & 35.01 & 46.32 & 2.14 \\ 

GHOST & \multicolumn{6}{c|}{N/A} & 75.74 & 29.86 & 65.46 & 37.13 & 67.14 & 26.29 & \multicolumn{6}{c}{N/A} \\ 
\midrule

DENSE & 74.71 & 26.24 & 64.41 & 36.13 & 50.19 & 4.18 & 75.48 & 31.11 & 63.08 & 35.96 & 69.08 & 35.09 & 76.39 & 31.14 & 65.40 & 37.61 & 63.74 & 22.52 \\  

Co-Boost & 74.56 & 26.69 & 62.27 & 33.17 & 68.53 & 35.08 & 75.44 & 31.09 & 63.32 & 36.27 & 69.14 & 35.09 & 76.42 & 31.15 & 65.42 & 37.62 & 63.74 & 22.52 \\ \midrule

O-pFGL & \textbf{78.48} & \textbf{44.59} & \textbf{69.55} & \textbf{48.17} & \textbf{69.20} & \textbf{36.57} & \textbf{80.68} & \textbf{50.02} & \textbf{71.91} & \textbf{49.03} & \textbf{69.98} & \textbf{37.83} & \textbf{79.96} & \textbf{44.48} & \textbf{71.58} & \textbf{46.87} & \textbf{70.26} & \textbf{40.10} \\ 

\bottomrule[1pt]
\end{tabular}%
}
\caption{Performance of methods with different GNN architectures under Metis partition.}
\label{tab:diff_arch_metis}
\end{table*}

\begin{table*}[tbp]
\setlength{\tabcolsep}{1mm}
\small
\centering
\begin{tabular}{l|ccc|ccc}
\toprule[1pt]
\multirow{2}{*}{total \# nodes = 339} & \multicolumn{3}{c|}{\textbf{class1} (\# nodes = 50)} & \multicolumn{3}{c}{\textbf{class2} (\# nodes = 14)} \\
\cmidrule(r){2-4} \cmidrule(l){5-7}
 & \textbf{F1-macro} & \textbf{Recall} & \textbf{Precision} & \textbf{F1-macro} & \textbf{Recall} & \textbf{Precision} \\
\midrule
Standalone            & 51.28 & 50.00 & 52.63 & 0     & 0   & 0   \\
FedAvg                & 45.71 & 40.00 & 53.33 & 0     & 0   & 0   \\
\midrule
FedPUB                & 55.81 & 60.00 & 52.17 & 0     & 0   & 0   \\
FedGTA                & 50.00 & 55.00 & 45.83 & 29.63 & 57.14 & 20.00 \\
FedTAD                & 42.42 & 35.00 & 53.85 & 0     & 0   & 0   \\
FedSpray              & 50.00 & 56.25 & 45.00 & 0     & 0   & 0   \\
FedLoG                & 47.06 & 40.00 & 57.14 & 0     & 0   & 0   \\
FedGM                 & 56.42 & 55.00 & 57.89 & 0     & 0   & 0   \\
GHOST                 & 48.65 & 45.00 & 52.94 & 0     & 0   & 0   \\
\midrule
DENSE                 & 48.65 & 45.00 & 52.94 & 0     & 0   & 0   \\
Co-Boost              & 42.42 & 35.00 & 53.85 & 0     & 0   & 0   \\
\midrule
O-pFGL         & \textbf{63.83} & \textbf{75.00} & 55.56 & \textbf{50.00} & \textbf{42.86} & \textbf{60.00} \\
\bottomrule[1pt]
\end{tabular}
\caption{Fine-grained results regarding F1-macro, Recall, and Precision metrics.}
\label{tab:f1_recall_precision}
\end{table*}

\subsection{Fine-grained Experimental Results}

In Figure~\ref{fig:citeseer_major_minor}, we've separately compared the F1-macro of our and baseline methods on minor classes (class 1 and class 2). The class-specific F1 and F1-macro together support that our method not only improves the performance of major classes, but can also effectively generalize to the minority. We select one client among 10 clients on the CiteSeer dataset under the Louvain partition to provide more comparisons on minor classes regarding F1-macro, Recall, and Precision metrics in Table~\ref{tab:f1_recall_precision}.

\section{Hyper-parameters Analysis}
\label{sec:appendix_hyperparamter_sensitive}

\begin{table*}[tbp]
\setlength{\tabcolsep}{1mm}
\small
\centering
\begin{tabular}{l|cc|cc|cc}
\toprule[1pt]
\multirow{2}{*}{\# nodes per class} & \multicolumn{2}{c|}{\textbf{Cora}} & \multicolumn{2}{c|}{\textbf{CiteSeer}} & \multicolumn{2}{c}{\textbf{PubMed}} \\ \cline{2-7}
& Acc. & F1 & Acc. & F1 & Acc. & F1  \\
\midrule
\textbf{1 (used in our paper)} & 76.43{\scriptsize$\pm$1.24} & 61.58{\scriptsize$\pm$2.16} & 71.61{\scriptsize$\pm$0.40} & 58.24{\scriptsize$\pm$0.96} & 82.71{\scriptsize$\pm$0.03} & 66.40{\scriptsize$\pm$0.34} \\
2                     & 70.40{\scriptsize$\pm$2.62} & 54.29{\scriptsize$\pm$3.45} & 66.86{\scriptsize$\pm$0.56} & 53.63{\scriptsize$\pm$1.48} & 82.57{\scriptsize$\pm$0.04} & 66.01{\scriptsize$\pm$0.28}  \\
3                     & 70.71{\scriptsize$\pm$1.58} & 52.33{\scriptsize$\pm$2.85} & 66.39{\scriptsize$\pm$2.73} & 51.61{\scriptsize$\pm$3.92} & 82.30{\scriptsize$\pm$0.04} & 65.17{\scriptsize$\pm$0.30} \\
\bottomrule[1pt]
\end{tabular}
\caption{Performance with different number of nodes per class}
\label{tab:p_cora_citeseer_pubmed}
\end{table*}

\begin{table}[tbp]
\centering
\resizebox{\columnwidth}{!}{%
\begin{tabular}{l|cc}
\toprule[1pt]
\multirow{2}{*}{\# nodes per class (percent of train set)} & \multicolumn{2}{c}{\textbf{ogbn-arxiv}} \\ \cline{2-3}
& Acc. & F1 \\
\midrule
0.05\%                & 70.51{\scriptsize$\pm$0.10} & 34.86{\scriptsize$\pm$0.27} \\
0.1\%                 & 70.59{\scriptsize$\pm$0.15} & 35.28{\scriptsize$\pm$0.27} \\
\textbf{0.25\% (used in our paper)} & 70.93{\scriptsize$\pm$0.59} & 36.66{\scriptsize$\pm$0.22} \\
0.5\%                 & 70.98{\scriptsize$\pm$0.16} & 36.67{\scriptsize$\pm$0.25} \\
1\%                   & 71.02{\scriptsize$\pm$0.05} & 36.68{\scriptsize$\pm$0.26} \\
\bottomrule[1pt]
\end{tabular}
}
\caption{Results on ogbn-arxiv with different number of nodes per class (percent of train set).}
\label{tab:p_ogbn_arxiv}
\end{table}

\paragraph{Impact of the pre-set number of nodes in synthesized global surrogate graph.}
When increasing the number, the generated global surrogate graph could have more potential to capture more information. But a large number of nodes would incur optimization difficulties and high communication costs when distributing them to clients. Thus, our principles to choose the number are (1) set the smallest number to capture enough information, (2) keep a similar label distribution with the training set. According to Section~\ref{sec:app_hyper-parameter}, we pre-set the number of nodes to synthesize to 1 for three small graphs and pre-set the number of nodes to synthesize to $p$ percent of the total number of nodes in the corresponding class in the training set, while we keep $p$ as small as possible. We conduct experiments on 10 clients under the Louvain partition to show how the number of nodes impacts the performance. The experimental results are shown in Table~\ref{tab:p_cora_citeseer_pubmed} and Table~\ref{tab:p_ogbn_arxiv}. For small-scale graphs: Cora, CiteSeer, and PubMed, we pre-set the smallest number of nodes for each class (i.e., one node for each class) and find that it's already capable. Increasing the number of nodes would not bring benefit since it incurs optimization difficulty. For larger graph: ogbn-arxiv, we pre-set the number of nodes for each class as 0.25\% of the total number of corresponding class's nodes in the train set. Fewer nodes (0.1\%, 0.05\%) could not fully capture the information in the aggregated class-wise statistics, and lead to a performance drop. More nodes (0.5\%, 1\%) are more capable but incur more computation and communication costs, and the performance improvement is marginal. Note that the number of nodes in the global surrogate graph is always small. Even on the largest graph ogbn-product, the total number of nodes of all classes is less than 150. Thus, the memory and communication costs are negligible.

\begin{table*}[tbp]
\setlength{\tabcolsep}{1mm}
\small
\centering
\begin{tabular}{l|cc|cc|cc|cc}
\toprule[1pt]
\multirow{2}{*}{\textbf{ogbn-arxiv}} & \multicolumn{2}{c|}{$d_{th}=10$} & \multicolumn{2}{c|}{$d_{th}=15$} & \multicolumn{2}{c|}{$d_{th}=20$} & \multicolumn{2}{c}{$d_{th}=25$} \\ \cline{2-9}
& Acc. & F1 & Acc. & F1 & Acc. & F1 & Acc. & F1  \\
\midrule
$f_{th}=0.95$ & 70.91{\scriptsize$\pm$0.16} & 36.21{\scriptsize$\pm$0.32} & 70.93{\scriptsize$\pm$0.59} & 36.66{\scriptsize$\pm$0.22} & 70.94{\scriptsize$\pm$0.15} & 36.47{\scriptsize$\pm$0.23} & 70.94{\scriptsize$\pm$0.08} & 36.37{\scriptsize$\pm$0.06}\\
$f_{th}=1$ & 70.82{\scriptsize$\pm$0.07} & 35.99{\scriptsize$\pm$0.11} & 70.79{\scriptsize$\pm$0.12} & 36.14{\scriptsize$\pm$0.51} & 70.72{\scriptsize$\pm$0.02} & 35.78{\scriptsize$\pm$0.11} & 70.84{\scriptsize$\pm$0.22} & 36.30{\scriptsize$\pm$0.61} \\
\bottomrule[1pt]
\end{tabular}
\caption{Performance with different values of $f_{th}$ and $d_{th}$}
\label{tab:f_d_ogbn_arxiv}
\end{table*}

\begin{figure}
    \centering
    \includegraphics[width=0.95\linewidth]{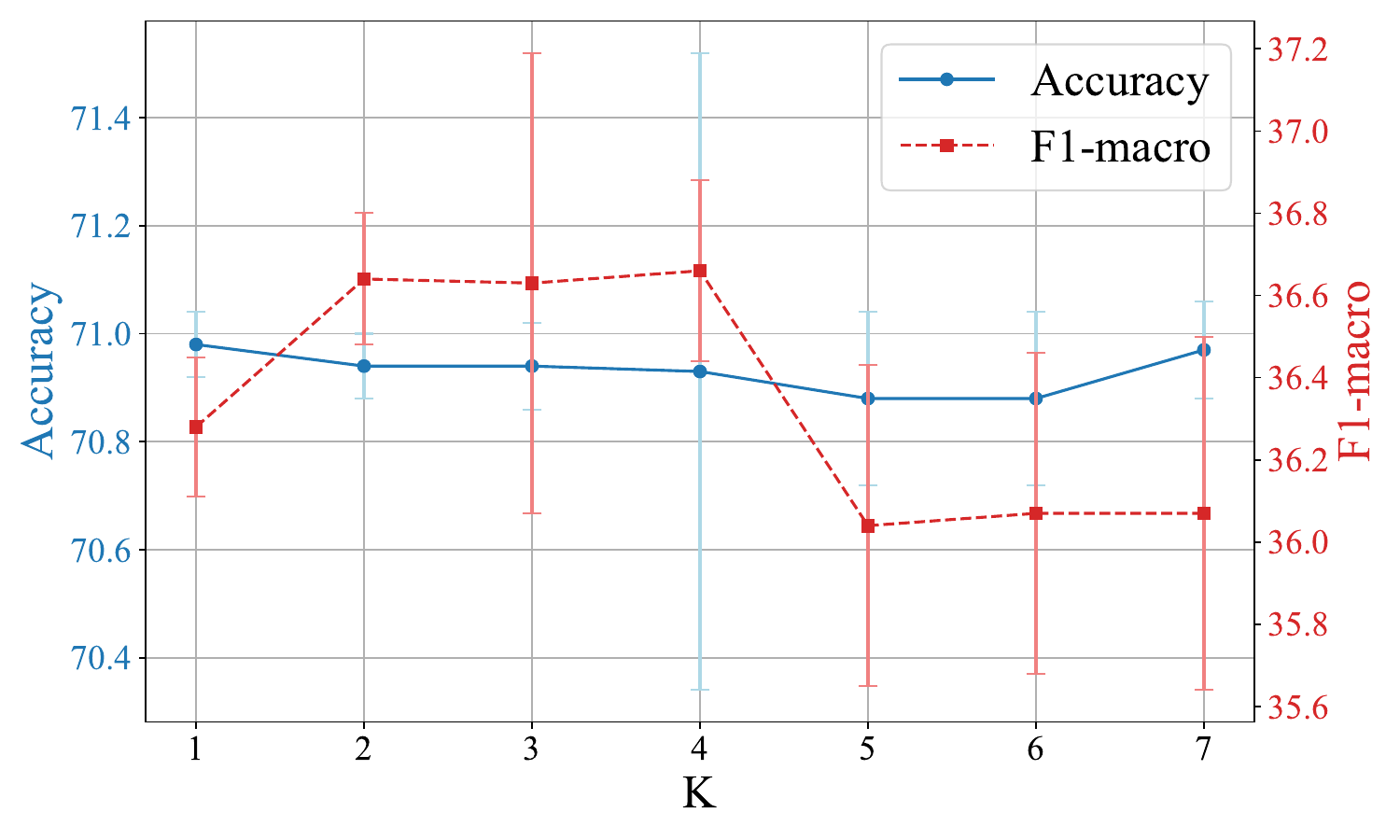}
    \caption{Performance with different $K$ values}
    \label{fig:topk_select}
\end{figure}

\begin{table*}[tbp]
\setlength{\tabcolsep}{1mm}
\small
\centering
\begin{tabular}{l|cc}
\toprule[1pt]
& Acc. & F1  \\
\midrule
Ours & \textbf{76.43}{\scriptsize$\pm$1.24} & \textbf{61.58}{\scriptsize$\pm$2.16} \\
Without $d_{th}$ (i.e., without \textbf{(C1)}) & 75.06{\scriptsize$\pm$0.46} & 58.75{\scriptsize$\pm$0.82} \\
Without $d_{th}$ and $f_{th}$ (i.e., without \textbf{(C1)} and \textbf{(C2)}) & 72.51{\scriptsize$\pm$0.40} & 53.49{\scriptsize$\pm$2.13} \\
Without $d_{th}$, $f_{th}$ and $topK$ (i.e., without \textbf{(C1)}, \textbf{(C2)} and \textbf{(C3)}) & 72.48{\scriptsize$\pm$2.45} & 53.48{\scriptsize$\pm$1.76} \\
\bottomrule[1pt]
\end{tabular}
\caption{Performance without $f_{th}$ and $d_{th}$}
\label{tab:necessity_f_d_topk_cora}
\end{table*}

\paragraph{Impact of $d_{th}$, $f_{th}$ and $topK$ in HRE.}
In our proposed Homophily-guided Reliable node Expansion strategy, we filter out the unreliable soft labels whose node degrees are less than $d_{th}$ (i.e., \textbf{(C1)}) and confidences are less than $f_{th}$ (i.e., \textbf{(C2)}). Also, we only select reliable soft labels that potentially belong to the classes with $topK$ high class homophily (i.e., \textbf{(C3)}). However, our method is not sensitive to the selection of the threshold $f_{th}, d_{th}$, and $K$ values. We conduct experiments on ogbn-arxiv dataset under the Louvain partition with 10 clients. The experimental results under different $f_{th}$ and $d_{th}$ are shown in Table~\ref{tab:f_d_ogbn_arxiv}. We can observe that the performance remains stable under a wide range of $d_{th}$ and both $f_{th}$ values. Regarding $K$ values, each client could select a proper $K$ by statistical analysis of local label distribution, and the performance of our method is not sensitive to $K$ values. The experimental results are in Figure~\ref{fig:topk_select} to show the performance stability under different $K$.

To conclude, our method is not sensitive to these hyper-parameters within a proper range in HRE, and the selection would not incur complexity in deployment.

However, these criteria are still necessary because some soft labels produced by label propagation would be unreliable. We provide an experiment in Table~\ref{tab:necessity_f_d_topk_cora} on the Cora dataset with 10 clients under the Louvain partition to demonstrate the necessity of $d_{th}$, $f_{th}$, and $topK$.

\begin{figure}
    \centering
    \includegraphics[width=0.95\linewidth]{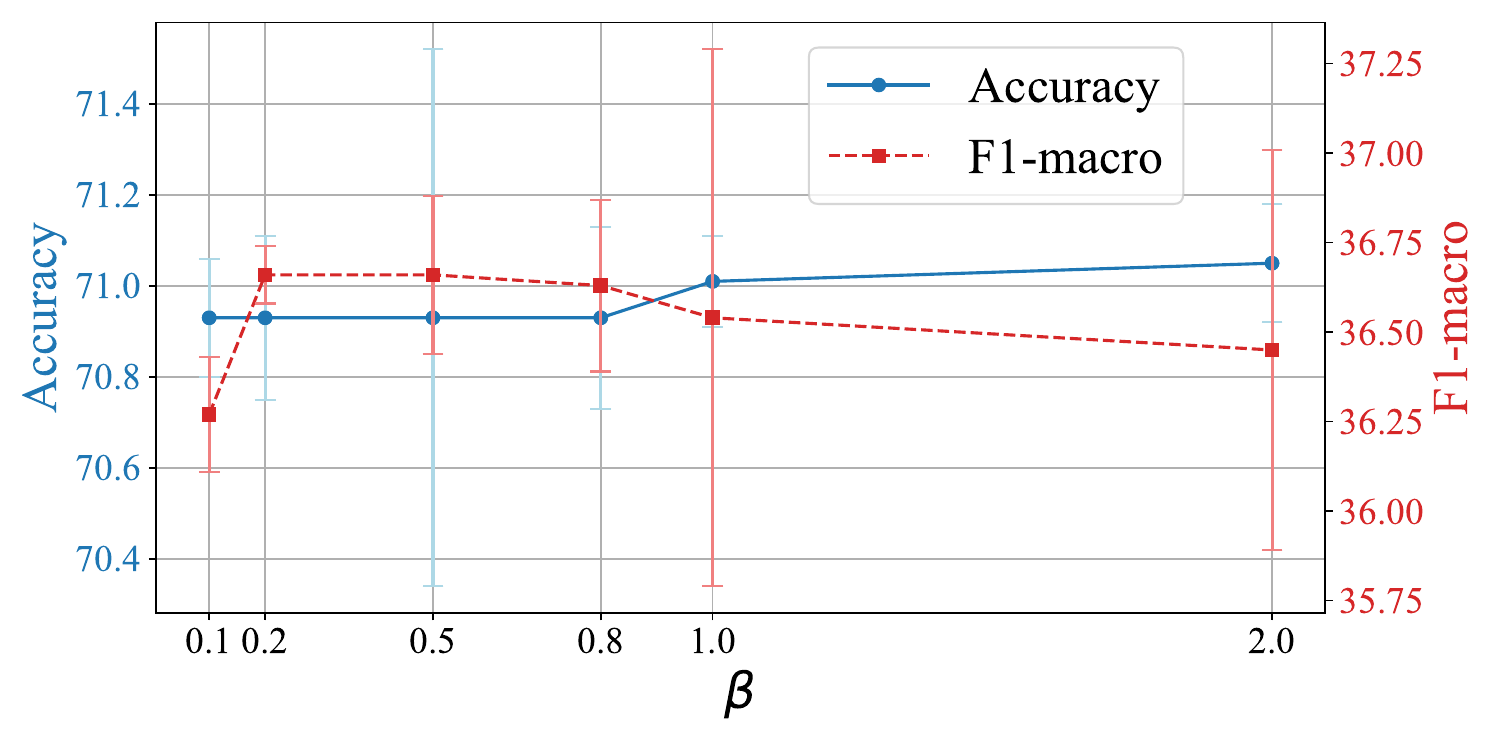}
    \caption{Performance with different $\beta$ values}
    \label{fig:gamma_select}
\end{figure}

\paragraph{Impact of $\beta$ in Node Adaptive Distillation}
In our proposed Node Adaptive Distillation, we use $\beta$ to control the scale of the value range of the $\gamma_i$. We provide experiments on ogbn-arxiv dataset under the Louvain partition with 10 clients to show the performance of our method under different $\beta$ values. The results are in Figure~\ref{fig:gamma_select}. We can observe that although performance would drop under too small or too large $\beta$, our method is not sensitive to $\beta$ as long as its values lie in a proper range.

\section{Complexity Analysis}
\label{sec:appendix_complexity}

We provide the complexity analysis of methods from three perspectives: communication, computation, and memory, with a 2-layer GCN model as an example.

\begin{table*}[h!]
\setlength{\tabcolsep}{1mm}
\small
\centering
\begin{tabular}{c|c|c}
\toprule[1pt]
\textbf{Communication} & \textbf{Upload} & \textbf{Download} \\  \midrule
Standalone & N/A & N/A \\ 
FedAvg & $O(fd + dC)$ & $O(fd + dC)$ \\  \midrule
FedPUB & $O(fd + dC)$ & $O(fd + dC)$ \\ 
FedGTA & $O(fd + dC + kKC)$ & $O(fd + dC)$ \\ 
FedTAD & $O(fd + dC)$ & $O(fd + dC)$ \\ 
FedSpray & $O(fd' + d'C)$ & $O(fd' + d'C)$ \\  
FedLoG & $O(fd + dC + n'^2f + \text{PG} + \text{Clf}_h + \text{Clf}_t)$ & $O(fd + dC + \text{Clf}_h + \text{Clf}_t + n'^2f)$ \\
FedGM & $O(n'f + \text{PGE})$ & $O(n'f + n'^2)$ \\
GHOST & $O(\text{G})$ & $O(fd+dC)$ \\
\midrule
DENSE & $O(fd + dC)$ & $O(fd + dC)$ \\ 
Co-Boost & $O(fd + dC)$ & $O(fd + dC)$ \\  \midrule
O-pFGL & $O(Chf)$ & $O(n'f+n'^2)$ \\ 
\bottomrule[1pt]
\end{tabular}
\caption{Communication complexity of methods.}
\label{tab:comm_complexity}
\end{table*}

\paragraph{Communication Analysis.} We provide the communication analysis of the methods in Table~\ref{tab:comm_complexity}. FedAvg, FedPUB, FedTAD, DENSE, and Co-Boost all need to upload and download the whole model parameters ($O(fd+dC)$, where $d$ is the hidden dimension). FedGTA needs to upload additional mixed moments of neighboring features ($O(kKC)$, where $K$ is the order of the moment and $k$ is the propagation steps of Label Propagation) for personalized aggregation. FedSpray needs to upload and download the global feature-structure encoder and structure proxy ($O(fd'+d'C)$, where $d'$ is the output dimension of the encoder and dimension of the structure proxy). FedLoG needs to upload the pre-trained prompt generator (for simplicity, we use $O(\text{PG})$ to denote the size of the prompt generator), head-branch classifier and tail-branch classifier (we use $O(\text{Clf}_h)$ and $O(\text{Clf}_t)$ to denote the size of the two branch classifiers), synthetic graph ($O(n'^2f)$, where $n'$ is the pre-set number of nodes), along with the model, to the server. Each client in FedLoG needs to download the aggregated head-branch and tail-branch classifier and the model. The PG, Clf$_h$, and Clf$_t$ are implemented as multi-layer perceptrons with relatively large hidden dimensions. Thus, FedLoG would incur high upload and download communication costs. FedGM needs to upload the synthesized graphs and a link predictor ($\text{PGE}$) on each client. The link predictor is usually larger than the model. And the server in FedGM would distribute the aggregated synthesized graphs to clients. For GHOST, each client needs to upload the proxy model, which is inherently a generator $\text{G}$. And the server would distribute the generalized model. In contrast, our method only requires uploading the class-wise statistics ($O(chf)$) and downloading the generated global surrogate graph ($O(n'f+n'^2)$). $C$ is the number of classes. $h$ is the propagation step in Eq.~\ref{eq:client_feat_prop} and $h \leq 2$ in our experiments. Although in some rare cases where $C$ is extremely large, the upload costs would still not become a huge burden compared to the model size. $n'$ is the number of nodes in the global surrogate graph, which is usually a pre-set small value, as indicated in Sec~\ref{sec:app_hyper-parameter}. The communication costs of our method are independent of the model size. The larger the model, the more communication-efficient our method is compared to other methods.

\begin{table*}[htbp]
\setlength{\tabcolsep}{1mm}
\small
\centering
\begin{tabular}{c|c|c}
\toprule[1pt]
\textbf{Computation} & \textbf{Client} & \textbf{Server} \\  \midrule
Standalone & $O(m(f + d) + nd(f + C))$ & N/A \\ 
FedAvg & $O(m(f + d) + nd(f + C))$ & $O(N(fd + dC))$ \\  \midrule
FedPUB & $O(m(f + d) + nd(f + C))$ & $O(N^2(fd + dC) + K n'^2)$ \\ 
FedGTA & $O(m(f + d + kC) + nd(f + C))$ & $O(N^2(fd + dC + kKC))$ \\ 
FedTAD & $O(pn^2 + m(f + d) + nd(f + C))$ & $O(N(fd + dC) + n'^2 f + N(m'(f + d) + n'(d(f + C))) + L'n'd'^2)$ \\ 
FedSpray & $O(m(f + d) + nd(f + C) + nfd' + nd'C)$ & $O(Nfd')$ \\  
FedLoG & $O(m(f + d) + nd(f + C) + (n'+n)d'^2)$ & $O(N(fd + dC + \text{PG} + \text{Clf}_h + \text{Clf}_t + n'^2f))$ \\
FedGM & $O(m(f + d) + nd(f + C))$ & $O(N \cdot \text{PGE})$ \\
GHOST & $O(\text{G} + n^2d' + m(f+d)+nd(f+C))$ & $O(N(\text{G}+n^2d'+m(fd+d^2) +dC))$ \\
\midrule
DENSE & $O(m(f + d) + nd(f + C))$ & $O(N(fd + dC) + n'^2 f + N(m'(f + d) + n'(d(f + C))) + L'n'd'^2)$ \\ 
Co-Boost & $O(m(f + d) + nd(f + C))$ & $O(N(fd + dC) + n'^2 f + N(m'(f + d) + n'(d(f + C))) + L'n'd'^2)$ \\ \midrule
O-pFGL & $O(m(f + d) + nd(f + C))$ & $O(NChf + m'f+ \text{PGE} + n'^2f)$ \\ 
\bottomrule[1pt]
\end{tabular}
\caption{Computational complexity of methods.}
\label{tab:comp_complexity}

\end{table*}

\paragraph{Computational Analysis.}
We provide computational analysis of methods in Table~\ref{tab:comp_complexity}. We slightly abuse the notation by redefining the number of clients as $N$, the number of edges in local graph data as $m$, and the number of nodes in local graph data as $n$. $K'$ is the number of proxy graphs generated in FedPUB, and $n'$ is the number of nodes in each proxy graph. $p$ is the diffusion step in FedTAD, and $O(pn^2)$ is the complexity to calculate the topology embedding, which could be highly computationally expensive. For FedTAD, DENSE, and Co-Boost, $O(n'^2f)$ is the complexity of constructing the generated graph. $L'$ is the number of layers of the generator, $f'$ is the dimension of the latent features in the generator, and $n'$ is the number of generated nodes. $O(L'n'd'^2)$ is the complexity of forwarding the generator in the server. For FedLoG, the additional computation costs come from the prompt generator pretraining and graph synthesis. Although the prompt generator and classifiers are multilayer perceptrons, the pre-training process involves bilevel optimization, which would incur high computation costs in practice. For FedGM, each client needs to conduct gradient matching to synthesize a graph. The server needs to aggregate and infer the synthesized graphs with uploaded $\text{PGE}$ from $N$ clients. For GHOST, the computational complexity is very high. Each client needs to construct a pseudo graph with the same size as the local graph by feature matrix multiply. This operation is $O(n^2d)$, which would be impractical on large-scale datasets. On the server side, to integrate the knowledge and avoid catastrophic forgetting, the server needs to traverse every edge in the pseudo graphs. The number of edges is approximately the same order as the number of edges in the latent global graphs. Thus, this process is with $O(m(fd+d^2))$ complexity, also impractical under large-scale datasets. For our method, $O(NChf)$ is the complexity of aggregating the statistics. $m'$ is the estimated number of edges in the generated global surrogate graph, usually a small value. $O(m'f+Chf)$ is the complexity of calculating alignment loss. $O(n'^2f)$ is the complexity of calculating the smoothness loss. The computational complexity on clients in our method is kept as least as possible, the same as the vanilla FedAvg and Standalone.

\begin{table*}[htbp]
\setlength{\tabcolsep}{1mm}
\small
\centering
\begin{tabular}{c|c|c}
\toprule[1pt]
\textbf{Memory} & \textbf{Client} & \textbf{Server} \\  \midrule
Standalone & $O(fd + dC)$ & N/A \\ 
FedAvg & $O(fd + dC)$ & $O(N(fd + dC))$ \\  \midrule
FedPUB & $O(fd + dC)$ & $O(N(fd + dC) + K'n'f)$ \\ 
FedGTA & $O(fd + dC + kKc)$ & $O(N(fd + dC + kKc))$ \\ 
FedTAD & $O(fd + dC + pn^2)$ & $O(N(fd + dC) + n'f + n'^2)$ \\ 
FedSpray & $O(fd + dC + fd' + d'C + nd')$ & $O(Nnd' + f'd' + d'C)$ \\  
FedLoG & $O(fd + dC + n'f + \text{PG} + \text{Clf}_h + \text{Clf}_t + n\overline{d}^kf )$ & $O(N(fd + dC + \text{PG} + \text{Clf}_h + \text{Clf}_t + n'^2f))$ \\  
FedGM & $O(n'f+fd+dC)$ & $O(n'f + N\cdot\text{PGE})$ \\
GHOST & $O(n^2 + \text{G} + fd+dC)$ & $O(N(\text{G} + n^2 + nf) + fd+dC)$ \\
\midrule
DENSE & $O(fd + dC)$ & $O(N(fd + dC)+ n'f + n'^2)$ \\ 
Co-Boost & $O(fd + dC)$ & $O(N(fd + dC)+ n'f + n'^2)$ \\  \midrule
O-pFGL & $O(fd + dC)$ & $O(Nchf + n'f + n'^2)$ \\
\bottomrule[1pt]
\end{tabular}
\caption{Memory complexity of methods.}
\label{tab:mem_complexity}
\end{table*}

\paragraph{Memory Analysis.} We provide memory analysis of methods in Table~\ref{tab:mem_complexity}. Standalone, FedAvg, FedPUB, DENSE, and Co-Boost only need to store the model parameters ($O(fd+dC)$). FedTAD needs to additionally store the topology embedding calculated by graph diffusion ($O(pn^2)$), thus it faces OOM on larger graphs. FedSpray needs to store additional encoders and the structure proxy. FedLoG needs to store the additional prompt generator, two branch classifiers, and generated graphs. Notably, in prompt generator pretraining, FedLoG additionally needs to obtain the k-hop subgraph of each node in the local graph, which would incur huge memory ($O(n\overline{d}^kf)$, where $\overline{d}$ is the average degree and $k$ is the pre-set number of hops). This is the key factor leading to the out-of-memory issue in dense graphs or large-scale graphs. FedGM needs to store the gradient of synthesized graphs additionally, and the server needs to store the $\text{PGE}$ from $N$ clients. GHOST needs to store the pseudo graph on both clients and the server, thus incurs high memory consumption with $O(n^2)$, which is impractical on large-scale datasets. For FedTAD, DENSE, Co-Boost, and our method, the server needs to additionally store the generated graph, which costs approximately $O(n'f+n'^2)$ memory.  Note that $n'$ is always a small number. Even on the largest graph ogbn-product, the total number of nodes of all classes is less than 150 ($ n' < 150$). Thus, it would not incur much memory overhead on the server.

\section{An Equivalent Server-Efficient Variant}
\label{sec:appendix_sever_efficient_variant}
An equivalent variant of our method is to move the global surrogate graph generation process to the clients. Specifically, the server only securely aggregates the uploaded statistics and then distributes $\{N^c, \boldsymbol{\mu}^c, \boldsymbol{s^2}^c \}$ to clients (with Homomorphic Encryption~\cite{acar2018survey}). Then the clients generate the global surrogate graph with aggregated statistics locally. It still supports model heterogeneity and Secure Aggregation protocols. This variant enables extreme efficiency on the server since the server only needs to perform a series of weighted average operations, moving the computation overhead of global surrogate graph generation to the client side.

\section{Related Works}
\label{sec:appendix_related_works}
\subsection{Federated Graph Learning}
With the rapid development of federated learning methods, recent works introduce federated graph learning to collaboratively train graph models~\cite{kipf2016semi} in a privacy-preserving manner and apply it to many applications~\cite{wu2021fedgnn, zhang2024gpfedrec, yan2024federated, tang2024personalized, chen2021fede}. From the graph level, each client possesses multiple completely disjoint graphs (e.g., molecular graphs). Recent works~\cite{xie2021federated, tan2023federated, tan2024fedssp, fu2025virtual} mainly focus on the intrinsic heterogeneity among graphs from different clients. From the subgraph level, the graph possessed by each client can be regarded as a part of a larger global graph. To cope with heterogeneous graph data, GraphFL~\cite{wang2022graphfl} adopts meta-learning~\cite{finn2017model} for better generalization. FedGL~\cite{chen2024fedgl} uploads the node embedding and prediction for global supervision but faces a heavy communication burden and potential privacy concerns. FGSSL~\cite{huang2024federated} augments the local graph to mitigate the heterogeneity. To enhance the model utility on each client, personalized federated graph learning methods are proposed. FedPUB~\cite{baek2023personalized} generates random graphs~\cite{holland1983stochastic} to measure the similarity in model aggregation and conducts adaptive weight masks for better personalization. FedGTA~\cite{li2024fedgta} proposes topology-aware personalized optimization. AdaFGL~\cite{li2024adafgl} studies the structure non-IID problem. FedTAD~\cite{zhu2024fedtad} utilized ensemble local models to perform data-free distillation on the server. FedGM~\cite{zhang2025rethinking} adopts gradient matching to synthesize graphs and uploads the synthesized graphs or gradients to the server for aggregation. GHOST~\cite{qianghost} trains a proxy model to align with local graph data, and the server integrates the knowledge from the uploaded proxy model. To complete missing connections between graphs of clients, FedSage~\cite{zhang2021subgraph} and FedDEP~\cite{zhang2024deep} additionally train a neighborhood generator. FedGCN~\cite{yao2024fedgcn} additionally uploads and downloads the encrypted neighbor features to supplement the features of the neighborhood. FedStruct~\cite{aliakbari2024decoupled} decouples the structure learning and node representation learning. To better represent the minority in local graph data, FedSpray~\cite{fu2024federated} learns local class-wise structure proxies to mitigate biased neighboring information. But it needs numerous communication rounds for optimization. FedLoG~\cite{kim2025subgraph} synthesizes node features in each client, and the server aggregates these node features to generate global synthetic data. Each client then generalizes its local training via the synthetic graph.

\subsection{One-shot Federated Learning}
One-shot federated learning largely reduces communication costs and circumvents potential man-in-the-middle attacks. Mainstream OFL methods can be classified into 3 categories. (1) Ensemble-based: The original OFL study~\cite{guha2019one} ensembles local models and conducts knowledge distillation with public data. DENSE~\cite{zhang2022dense} employs model inversion~\cite{yin2020dreaming} from the ensemble model to generate images for distillation. FedOV~\cite{diao2023towards} introduces placeholders in the model prediction layer. IntactOFL~\cite{zeng2024one} trains a MoE~\cite{jacobs1991adaptive} network by the generated images. Co-Boost~\cite{dai2024enhancing} further optimizes the generated images and ensemble weights iteratively. (2) Distillation-based: DOSFL~\cite{zhou2020distilled} and FedD3~\cite{song2023federated} conduct dataset distillation locally and upload distilled data for server-side training. (3) Generative-based: FedCVAE~\cite{heinbaugh2023data} trains VAEs for each client to generate similar images, addressing data heterogeneity. FedDISC~\cite{yang2024exploring} and FedDEO~\cite{yang2024feddeo} leverage the pre-trained Stable Diffusion~\cite{rombach2022high} to generate images and mitigate data heterogeneity. However, existing OFL methods primarily focus on image data and are either incompatible or ineffective for graph learning.

\subsection{Critical Analysis of Current Related Work}

\paragraph{Most FGL methods are ineffective within one-shot communication under data heterogeneity.}
Existing FGL methods typically follow the conventional paradigm in which the server adaptively aggregates client models. Under practical non-IID scenarios, numerous studies have shown that this paradigm can lead to issues such as weight divergence~\cite{zhao2018federated}, client drift~\cite{karimireddy2020scaffold}, and biases and conflicts among client models~\cite{zhu2021data}. These challenges necessitate extensive communication rounds to achieve acceptable performance.

Specialized FGL methods (e.g., FedPUB, FedTAD, FedGTA) necessitate unbiased client models with strong predictive abilities for various tasks, including proxy embedding computation, knowledge distillation, and label propagation. These methods require significant communication rounds to resolve these challenges. FedSpray additionally optimizes the structure proxy, which demands increased communication rounds for optimization. FedLoG needs to optimize the global synthetic data, which also needs iterative communications.

FedGM and GHOST offer methods for training a generalized model in one-shot communication. However, they are not robust to non-IID scenarios. FedGM adopts gradient matching on each client, which is unstable under biased and heterogeneous local graph data. GHOST constructs the pseudo graph with the aligned proxy model, but the proxy model is still biased and heterogeneous under non-IID client graph data. And the performance degrades easily in our experiments. Moreover, neither of them is scalable with respect to the time complexity and memory consumption. The high complexity hinders them to adapt to larger graphs effectively.

\paragraph{Most OFL methods are not designed for graph data.}
Existing OFL methods are mostly designed for image data. When dealing with graph data, these methods are incompatible or impractical. To be specific, ensemble-based methods such as DENSE train a generator to generate pseudo images via model inversion. However, it's not perfectly compatible with graph data generation since it can hardly capture the fine-grained structural information of nodes. Distillation-based methods generate distilled data on clients, and the server assembles the distilled data. However, the nodes in a graph are not independent. It's tricky to assemble generated graphs from clients on the server side. Generative-based methods use VAE or pretrained Stable Diffusion, which cannot generate graph data. Thus, there are no OFL methods that fit perfectly for graph data.

\paragraph{Most FGL and OFL methods are not compatible with Secure Aggregation.}
Secure Aggregation protocols are widely used in federated learning, which enable the server to compute the sum of large, user-held data vectors securely, without accessing the individual client contributions. Thus, the client's uploaded models or other information could be protected from third-party and the curious server. The application of Secure Aggregation is based on the fact that only weighted average operations are conducted in vanilla federated learning methods like FedAvg. However, the recent federated graph learning methods have become complicated. The server needs to conduct other operations, including ensemble (e.g., FedTAD, DENSE, and Co-Boost), data-free distillation (e.g., FedTAD, DENSE, and Co-Boost), functional embedding calculation (e.g., FedPUB), complex similarity calculation (e.g., FedGTA), and others (e.g., FedLoG). The server must access every client uploading to conduct these operations. Thus, they are not compatible with Secure Aggregation and third-party, and the curious server could easily access each client's uploaded information.

\begin{figure}[htbp]
\centerline{\includegraphics[width=0.9\columnwidth]{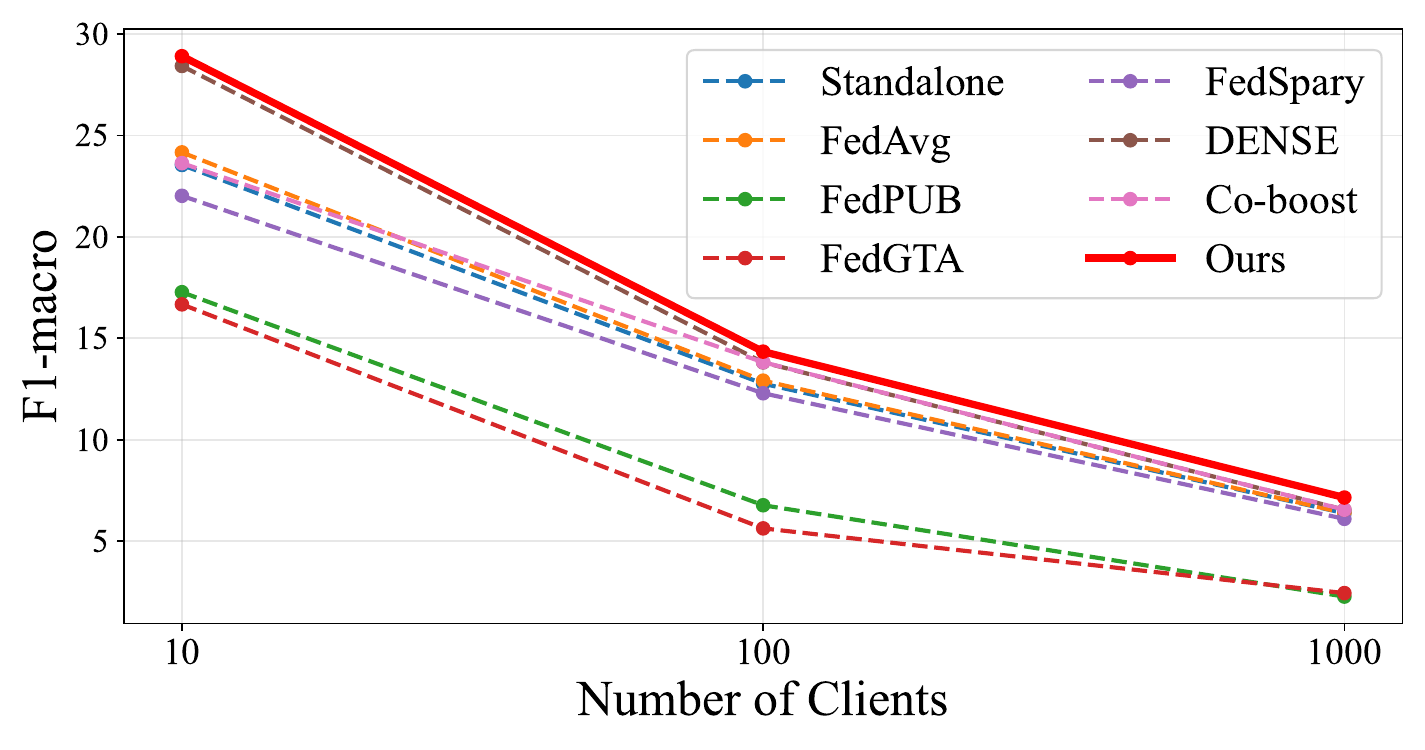}}
\caption{Performance of method with a large number of clients on ogbn-products dataset under the Metis partition.}
\label{fig:large_client}
\end{figure}

\begin{figure*}[htbp]
\centering
\subfigure[Cora]{
\includegraphics[width=0.32\textwidth]{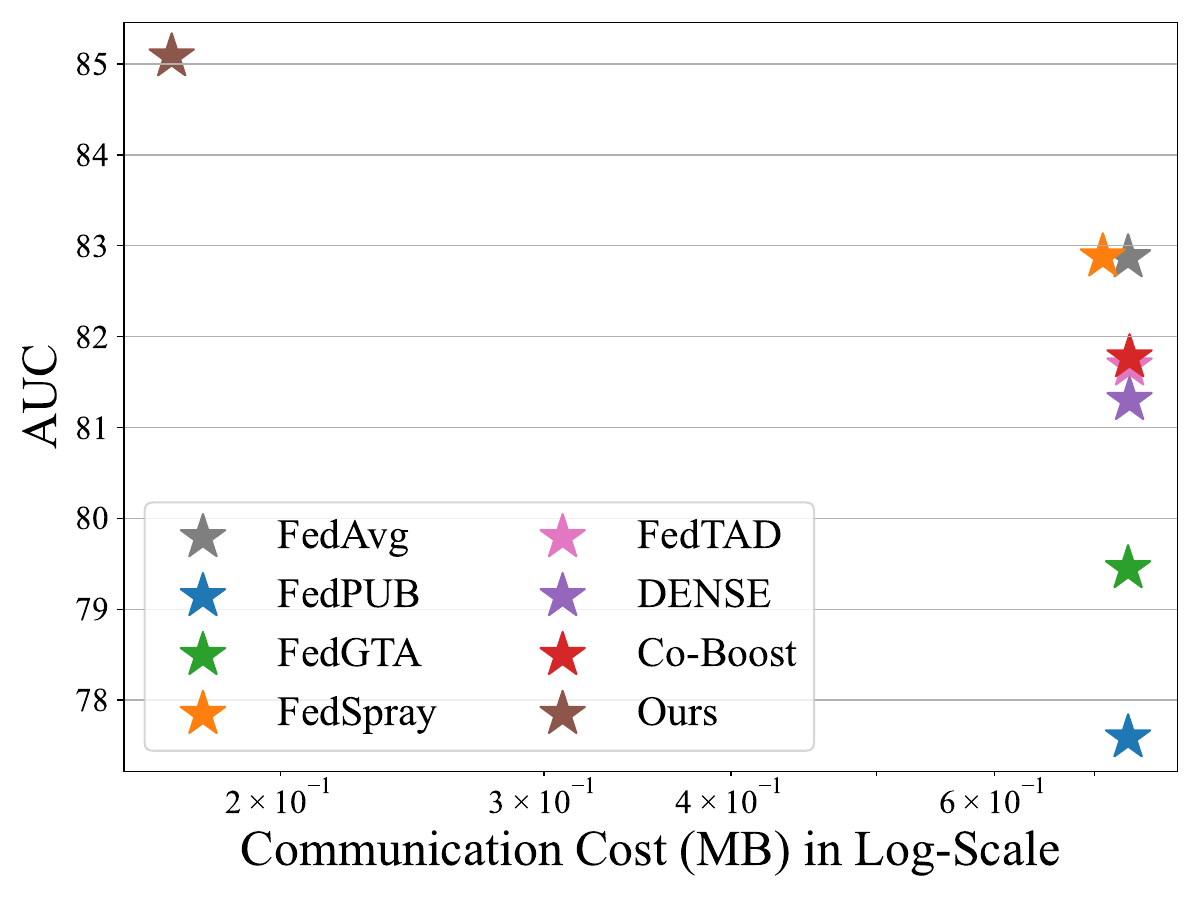}
}
\subfigure[CiteSeer]{
\includegraphics[width=0.32\textwidth]{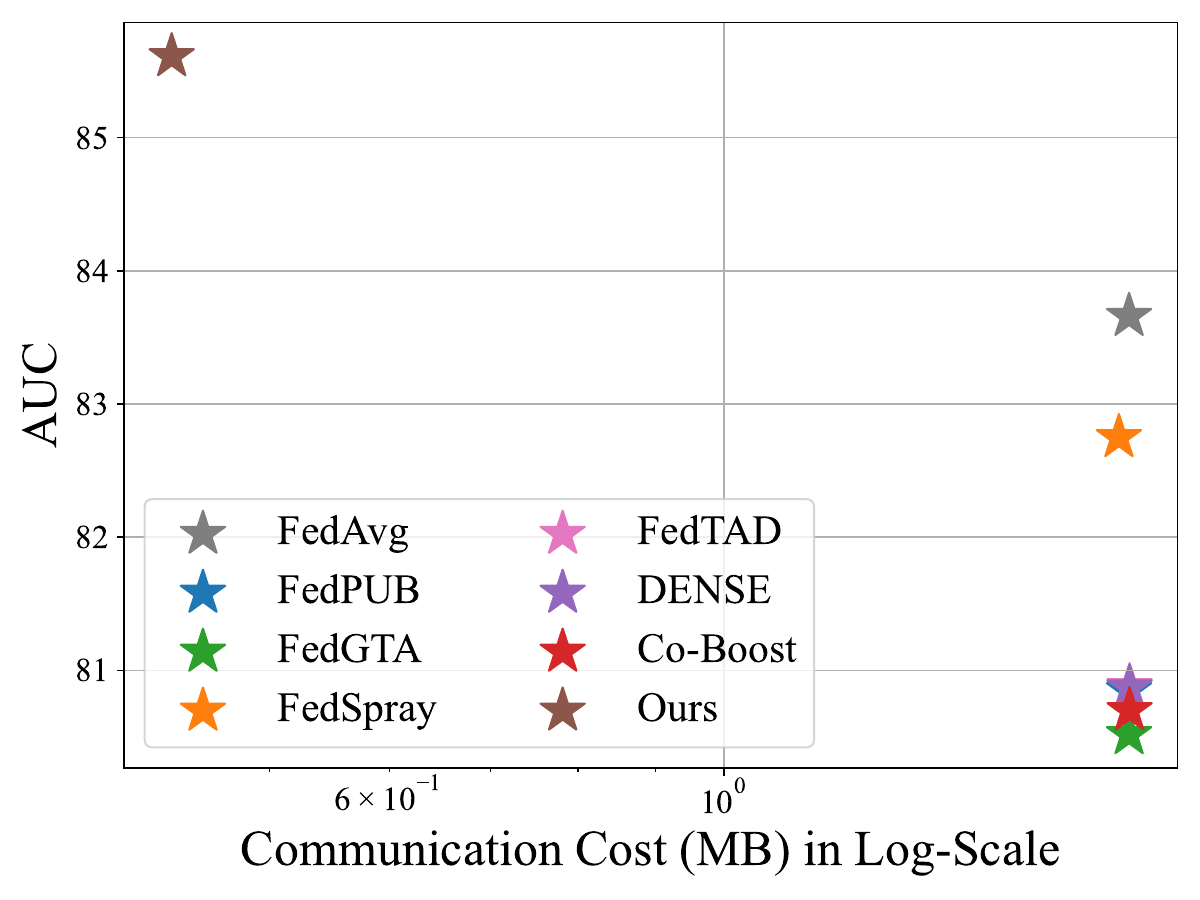}
}
\subfigure[PubMed]{
\includegraphics[width=0.32\textwidth]{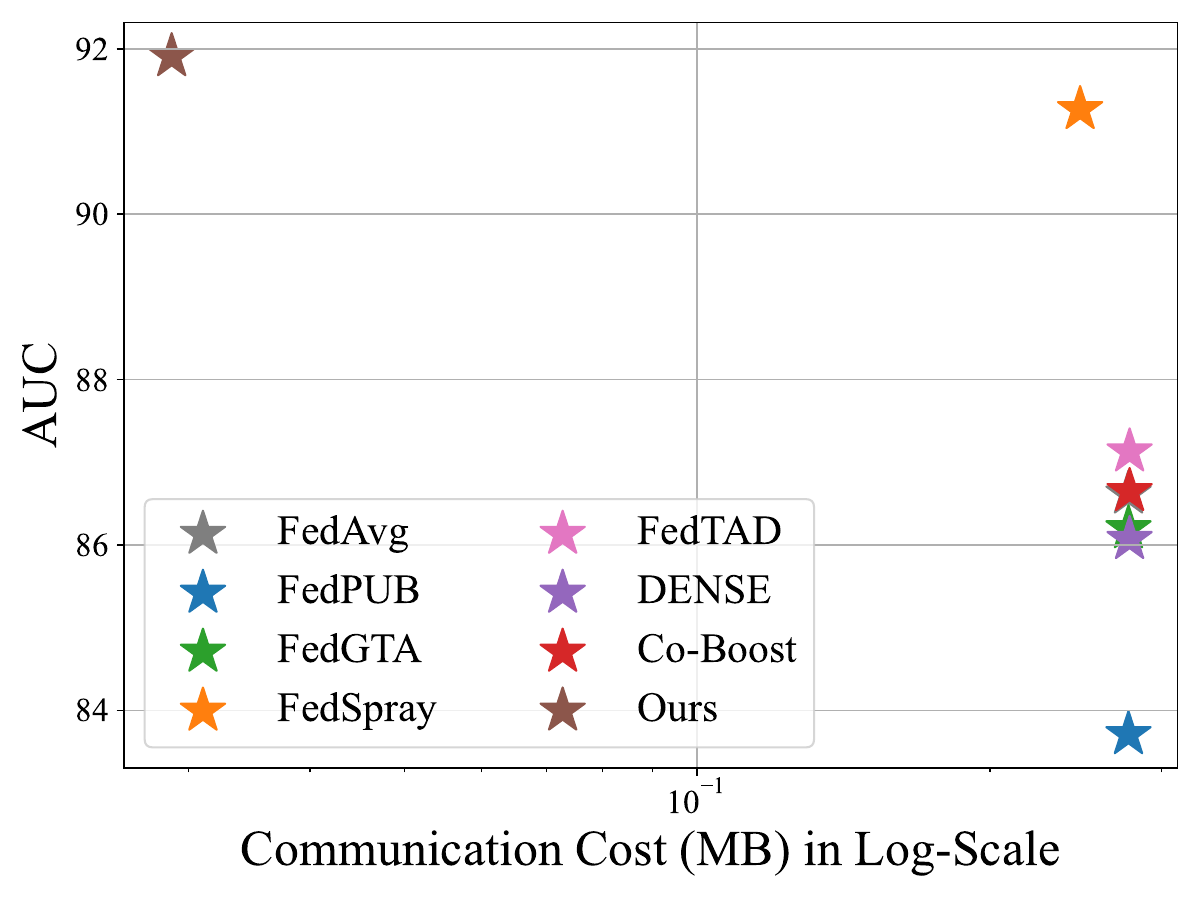}
}
\caption{Performance of method on the subsequent link prediction task on Cora, CiteSeer, PubMed datasets. The x-axis is set to a logarithmic scale.}
\label{fig:link_pred}
\end{figure*}

\section{Baselines.}
\label{sec:appendix_baselines}
We provide a detailed description of the baseline methods in the experiments.

\textbf{Standalone} is a non-federated learning method where each client trains its model only using its local graph data, without communication.

\textbf{FedAvg} is the fundamental federated learning method where each client trains its local model and uploads it to the server. The server weights averages the client models' parameters to obtain the global model and then distributes the global model to clients. In our experiments, we limit the upload-download communication round to one round. For better performance, we perform additional fine-tuning on local graph data after downloading the global model. 

\textbf{FedPUB} is a personalized federated graph learning method. FedPUB computes similarities between local models via functional embeddings derived from random graphs generated on the server. The similarities are used for weighted average in personalized aggregation. Additionally, FedPUB employs a personalized mask for each client model to selectively upload parameters. In our experiments, we limit the upload-download communication round to one round.

\textbf{FedGTA} is a personalized federated graph learning method. FedGTA computes and uploads the local smoothing confidence and mixed moments of neighbor features to the server to facilitate the weighted average in personalized aggregation. In our experiments, we limit the upload-download communication round to one round.

\textbf{FedTAD} computes the class-wise knowledge reliability scores on each client. The scores, along with local models, are uploaded to the server for data-free knowledge distillation from the ensemble of local models to the global model or personalized model. In our experiments, we limit the upload-download communication round to one round. For better performance, we perform additional fine-tuning on local graph data after downloading the model. 

\textbf{FedSpray} focuses on the minor classes. It optimizes the structure proxy along with the local model on each client. FedSpray uploads and aggregates the structure proxy, which is used in local model training to prevent overfitting and ensure the generalization to the minor classes. In our experiments, we limit the upload-download communication round to one round.

\textbf{FedLoG} focuses on the local generalization of model training. Each client pretrains a prompt generator and synthesizes nodes. The server aggregates the model and synthesized nodes and generates global synthetic data. Then, each client trains its model on local graph data and global synthetic data to derive a generalized model. In our experiments, we limit the upload-download communication round to one round.

\textbf{FedGM} adopts gradient matching to condense the synthesized graph on each client. Each client uploads the synthesized graphs or gradients to the server, and the server aggregates them. In our experiment, we follow the setting from the original paper to conduct one-round communication and use aggregated synthesized graphs to aid the model training. For better performance, we perform additional fine-tuning on local graph data after downloading the global model.

\textbf{GHOST} aims to train a generalized model in one-shot communication. Each client constructs a pseudo graph and trains a proxy model by aligning the pseudo graph and the real local graph. Each client then uploads the proxy model. The server generates the pseudo-graph with each uploaded proxy model and trains a global model with crucial parameters stabilized to alleviate catastrophic forgetting. For better performance, we perform additional fine-tuning on local graph data after downloading the global model.

\textbf{DENSE} is a one-shot federated learning method. Each client uploads its trained local model to the server. The server ensembles the client models to generate pseudo data via model inversion, which is used to distill knowledge from client models to the global model or personalized models. For better performance, we perform additional fine-tuning on local graph data after downloading the data. Note that it's designed for the image classification task. In our node classification task on graphs, the generators in model inversion on the server-side are trained to generate node features, and the topology structure is constructed using the $K$-Nearest Neighbors strategy as outlined in~\cite{zhu2024fedtad}.

\textbf{Co-Boost} is a one-shot federated learning method. It advances the DENSE. The server ensembles the client models with learnable weights. The ensemble weights and generated pseudo data are optimized alternately. For better performance, we perform additional fine-tuning on local graph data after downloading the data. Note that it's designed for the image classification task. In our node classification task on graphs, the generators in model inversion on the server-side are trained to generate node features, and the topology structure is constructed using the $K$-Nearest Neighbors strategy as outlined in~\cite{zhu2024fedtad}.

\section{Our method's Advantages in One Communication Round}
Most existing methods meet issues like weight divergence, client drift, and discrepancies and conflicts between client models raised from non-IID scenarios and fewer communication rounds. However, our method is robust to non-IID and can even achieve optimal performance in one round on average compared with multi-round federated learning methods. In our method, we do not aggregate the model parameters (which is a major source of information loss in other methods). Rather, the aggregation process in our method is lossless and unbiased regardless of the data distributions. Thus, we do not need multi-round communications to tackle the issues met by the existing methods above. Also, our method enhances the local generalization of the client models by considering both local and global information, which further boosts our method's performance.

\section{Experimental Results on a Large Number of Clients}
We conduct experiments on a large number of clients to evaluate the robustness of our method. We simulate 10, 100, and 1000 clients on ogbn-products with Metis partition. The F1-macro metrics of methods are shown in Figure~\ref{fig:large_client}. Our method consistently outperforms other methods on a large number of clients.

\section{Experimental Results on Link Prediction Task}
To evaluate the effectiveness of the methods of node representation learning, we perform the link prediction task. We first perform the node classification task to pre-train a GCN as the feature encoder with a classification head to generate distinguishable node representations, and then fine-tune the feature encoder for link prediction. Note that FedLoG is not included since its inference needs the participation of synthetic head and tail branch nodes, which cannot apply to the link prediction task directly. The experimental results (take AUC as our metrics) are shown in Figure~\ref{fig:link_pred}, which shows that our method learns better node representation and achieves the best on the subsequent link prediction task while having less communication cost.

\section{Limitations}
\label{sec:limitations}
Our method is based on the assumption that the clients and the server are honest and benign in the federated learning process, which is a common assumption in this research area. We leave the discussion about border scenarios with fewer assumptions for future work.



\end{document}